\documentclass{article}

    \usepackage[final]{neurips_2020}

\usepackage[utf8]{inputenc} % allow utf-8 input
\usepackage[T1]{fontenc}    % use 8-bit T1 fonts
\usepackage{hyperref}       % hyperlinks
\usepackage{url}            % simple URL typesetting
\usepackage{booktabs}       % professional-quality tables
\usepackage{amsfonts}       % blackboard math symbols
\usepackage{nicefrac}       % compact symbols for 1/2, etc.
\usepackage{microtype}      % microtypography
\usepackage{mathtools}

\usepackage{amsmath,amssymb, amsthm}

\usepackage{graphicx}
\usepackage{subcaption}
\usepackage{adjustbox}

\usepackage{booktabs}
\usepackage{multirow}
\usepackage{algorithm}
\usepackage[noend]{algpseudocode}
\usepackage{natbib}
\usepackage[colorinlistoftodos]{todonotes}

\title{Self-Distillation as Instance-Specific Label Smoothing}

\author{
 Zhilu Zhang \\
 Cornell University \\
 \texttt{zz452@cornell.edu} \\
  \And
  Mert R.~Sabuncu  \\
  Cornell Univerisity \\
  \texttt{msabuncu@cornell.edu}
}

\begin{document}

\maketitle

\begin{abstract}
    It has been recently demonstrated that multi-generational self-distillation can improve generalization~\cite{furlanello2018born}. 
    Despite this intriguing observation, reasons for the enhancement remain poorly understood. 
    In this paper, we first demonstrate experimentally that the improved performance of multi-generational self-distillation is in part associated with the increasing diversity in teacher predictions. 
    With this in mind, we offer a new interpretation for teacher-student training as amortized MAP estimation, such that teacher predictions enable instance-specific regularization.
    % Our framework allows us to theoretically understand why self-distillation can outperform label smoothing, a commonly used simple technique that regularizes predictive uncertainty.
    Our framework allows us to theoretically relate self-distillation to label smoothing, a commonly used technique that regularizes predictive uncertainty, and suggests the importance of predictive diversity in addition to predictive uncertainty.
    % We present experimental results using multiple datasets and neural network architectures that, overall, demonstrate that self-distillation leads to better performance than label smoothing.
    We present experimental results using multiple datasets and neural network architectures that, overall, demonstrate the utility of predictive diversity.
    Finally, we propose a novel instance-specific label smoothing technique that promotes predictive diversity without the need for a separately trained teacher model. 
    We provide an empirical evaluation of the proposed method, which, we find, often outperforms classical label smoothing.
    %thus further confirming the importance of predictive diversity.
    %Lastly, we demonstrate that self-distillation can also lead to improved calibration performance.
\end{abstract}

\section{Introduction}
First introduced as a simple method to compress high-capacity neural networks into a low-capacity counterpart for computational efficiency, knowledge distillation~\cite{hinton2015distilling} has since gained much popularity across various application domains ranging from computer vision to natural language processing~\cite{ kim2016sequence, li2017learning,papernot2016distillation,yim2017gift, yu2017visual} as an effective method to transfer knowledge or features learned from a teacher network to a student network. This empirical success is often justified with the intuition that deeper teacher networks learn better representation with greater model complexity, and the "dark knowledge" that teacher networks provide facilitates student networks to learn better representations and hence enhanced generalization performance. Nevertheless, it still remains an open question as to how exactly student networks benefit from this dark knowledge. 
The problem is made further puzzling by the recent observation that even self-distillation, a special case of the teacher-student training framework in which the teacher and student networks have identical architectures, can lead to better generalization performance~\cite{furlanello2018born}. It was also demonstrated that repeated self-distillation process with multiple generations can further improve classification accuracy. 
% Indeed, the intuition of dark knowledge does not seem to provide a convincing explanation for improvements seen with self-distillation, since both teacher and student models have the same capacity, and the dark knowledge learned by the teacher should not contain much more information than its student. 

In this work, we aim to shed some light on self-distillation. 
We start off by revisiting the multi-generational self-distillation strategy, and experimentally demonstrate that the performance improvement observed in multi-generational self-distillation is correlated with increasing diversity in teacher predictions. 
Inspired by this, we view self-distillation as instance-specific regularization on the neural network softmax outputs, and cast the teacher-student training procedure as performing amortized maximum a posteriori (MAP) estimation of the softmax probability outputs. The proposed framework provides us with a new interpretation of the teacher predictions as instance-specific priors conditioned on the inputs. 
%Interestingly, various other algorithms involving training with soft labels like label smoothing and mixup regularization can be unified under this perspective with different ways of obtaining the instance-specific priors. 
This interpretation allows us to theoretically relate distillation to label smoothing, a commonly used technique to regularize predictive uncertainty of NNs, and suggests that regularization on the softmax probability simplex space in addition to the regularization on predictive uncertainty can be the key to better generalization. 
To verify the claim, we systematically design experiments to compare teacher-student training against label smoothing. Lastly, to further demonstrate the potential gain from regularization on the probability simplex space, we also design a new regularization procedure based on label smoothing that we term ``Beta smoothing.'' 

Our contributions can be summarized as follows:
\begin{enumerate}
    \item We provide a plausible explanation for recent findings on multi-generational self-distillation.
    \item We offer an amortized MAP interpretation of the teacher-student training strategy.
    % \item We suggest a simple enhancement to the current distillation process.
    \item We attribute the success of distillation to regularization on both the label space and the softmax probability simplex space, and verify the importance of the latter with systematically designed experiments on several benchmark datasets. 
    \item We propose a new regularization technique termed ``Beta smoothing'' that improves upon classical label smoothing at little extra cost.
    \item We demonstrate self-distillation can improve calibration.
\end{enumerate}

\section{Related Works}
Knowledge distillation was first proposed as a way for model compression~\cite{ba2014deep, bucilu2006model, hinton2015distilling}. In addition to the standard approach in which the student model is trained to match the teacher predictions, numerous other objectives have been explored for enhanced distillation performance. For instance, distilling knowledge from intermediate hidden layers were found to be beneficial~\cite{heo2019comprehensive, huang2017like, kim2018paraphrasing, romero2014fitnets, srinivas2018knowledge,yim2017gift}. Recently, data-free distillation, a novel scenario in which the original data for the teacher is unavailable to students, has also been extensively studied~\cite{cai2020zeroq, chen2019data, micaelli2019zero, yoo2019knowledge}. 

The original knowledge distillation technique for neural networks~\cite{hinton2015distilling} has stimulated a flurry of interest in the topic, with a large number of published improvements and applications. 
For instance, prior works~\cite{balan2015bayesian, shen2020learning} have proposed Bayesian techniques in which distributions are distilled with Monte Carlo samples into more compact models like a neural network.
More recently, there has also been work on the importance of distillation from an ensemble of model~\cite{malinin2019ensemble}, which provides a complementary view on the role of predictive diversity. 
Lopez-Paz et al.~\cite{lopez2015unifying} combined distillation with the theory of privileged information, and offered a generalized framework for distillation. 
To simplify distillation, Zhu et al.~\cite{zhu2018knowledge} proposed a method for one-stage online distillation. 
There have also been successful applications of distillation for adversarial robustness~\cite{papernot2016distillation}.

Several papers have attempted to study the effect of distillation training on student models. Furlanello et al.~\cite{furlanello2018born} examined the effect of distillation by comparing the gradients of the distillation loss against that of the standard cross-entropy loss with ground truth labels. Phuong et al.~\cite{phuong2019towards} considered a special case of distillation using linear and deep linear classifiers, and theoretically analyzed the effect of distillation on student models. Cho and Hariharan~\cite{cho2019efficacy} conducted a thorough experimental analysis of knowledge distillation, and observed that larger models may not be better teachers. Another experimentally driven work to understand the effect of distillation was also done in the context of natural language processing~\cite{zhou2019understanding}. Most similar to our work is~\cite{yuan2019revisit}, in which the authors also established a connection between label smoothing and distillation. However, our argument comes from a different theoretical perspective and offers complementary insights. Specifically, \cite{yuan2019revisit} does not highlight the importance of instance-specific regularization. We also provide a general MAP framework and a careful empirical comparison of label smoothing and self-distillation.

\section{Preliminaries}
We consider the problem of $k$-class classification. Let $\mathcal{X} \subseteq \mathbb{R}^d$ be the feature space and $\mathcal{Y} = \{ 1, .., k \}$ be the  label space. Given a dataset $\mathcal{D} = \{ \boldsymbol{x}_i, y_i \}_{n=1}^n$ where each feature-label pair $(\boldsymbol{x}_i, y_i) \in \mathcal{X} \times \mathcal{Y}$, and we are interested in finding a function that maps input features to corresponding labels $f : \mathcal{X} \rightarrow \mathbb{R}^c$. In this work, we restrict the function class to the set of neural networks $f_{\boldsymbol{w}}(\boldsymbol{x})$ where $\boldsymbol{w} = \{ W_i \}^L_{i=1}$ are the parameters of a neural network with $L$ layers. We define a likelihood model $p(y | \boldsymbol{x}; \boldsymbol{w}) = \text{Cat}\left( \text{softmax}\left(f_{\boldsymbol{w}}(\boldsymbol{x})\right) \right)$, a categorical distribution with parameters $\text{softmax}\left(f_{\boldsymbol{w}}(\boldsymbol{x})\right)  \in \Delta(L)$. Here $\Delta(L)$ denotes the $L$-dimensional probability simplex. Typically, maximum likelihood estimation (MLE) is performed. This leads to the cross-entropy loss 
\begin{align}
    \mathcal{L}_{cce}(\boldsymbol{w}) = - \sum^n_{i=1} \sum^k_{j=1} \boldsymbol{y}_{ij} \log p(y = j | \boldsymbol{x}_i; \boldsymbol{w}), 
\end{align}
where $\boldsymbol{y}_{ij}$ corresponds to the $j$-th element of the one-hot encoded label $y_i$. 

\subsection{Teacher-Student Training Objective}
Given a pre-trained model (teacher) $f_{\boldsymbol{w}_t}$, distillation loss can be defined as:
\begin{align}
    \mathcal{L}_{dist}(\boldsymbol{w}) = - \sum^n_{i=1} \sum^k_{j=1} [\text{softmax}\big(f_{\boldsymbol{w}_t}(\boldsymbol{x}) / T \big)]_j \log p(y = j | \boldsymbol{x}_i; \boldsymbol{w}),
\end{align}
where $[\cdot]_j$ denotes the $j$'th element of a vector.
A second network (student) $f_{\boldsymbol{w}}$ can then be trained with the following total loss:
\begin{align}
\label{eqn:distillation}
    \mathcal{L}(\boldsymbol{w}) = \alpha \mathcal{L}_{cce}(\boldsymbol{w}) + (1 - \alpha) \mathcal{L}_{dist}(\boldsymbol{w}), 
\end{align}
where $\alpha \in [0,1]$ is a hyper-parameter, and $T$ corresponds to the temperature scaling hyper-parameter that flattens teacher predictions. In self-distillation, both teacher and student models have the same network architecture. In the original self-distillation experiments conducted by Furlanello et al.~\cite{furlanello2018born}, $\alpha$ and $T$ are set to $0$ and $1$, respectively throughout the entire training process.

% We can view the distillation loss in  Eq.~\ref{eqn:distillation} as instance-specific regularization, which will be discussed in detail in the following sections.
Note that, temperature scaling has been applied differently compared to previous literature on distillation~\cite{hinton2015distilling}.
As addressed in Section~\ref{section:map}, we only apply temperature scaling to teacher predictions in computing distillation loss. 
We empirically observe that this yields results consistent with previous reports. 
Moreover, as we show in the Appendix~\ref{appendix:student_temp}, performing temperature scaling only on the teacher but not the student models can lead to significantly more calibrated predictions. 
  
\section{Multi-Generation Self-Distillation: A Close Look}
\label{section:BAN}

Self-distillation can be repeated iteratively such that during training of the $i$-th generation, the model obtained at $(i-1)$-th generation is used as the teacher model. 
This approach is referred to as multi-generational self-distillation, or ``Born-Again Networks'' (BAN).
Empirically it has been observed that student predictions can consistently improve with each generation.
However, the mechanism behind this improvement has remained elusive.
% An immediate question to consider is: what is the most significant difference between each subsequent teacher predictions in such a sequential teacher-student training process that leads to the improvements in performance of the student models? 
%One possible argument is that the representations learned by each subsequent model become increasingly refined due to "dark knowledge" of previous generations, thereby leading to better generalization performance. 
In this work, we argue that the main attribute that leads to better performance is the increasing uncertainty and diversity in teacher predictions. 
Similar observations that more ``tolerant'' teacher predictions lead to better students were also made by Yang et al.~\cite{yang2019training}. Indeed, due the monotonicity and convexity of the negative log likelihood function, since the element that corresponds to the true label class of the softmax output $p(y = y_i | \boldsymbol{x}_i; \boldsymbol{w})$ is often much greater than that of the other classes, together with early stopping, each subsequent model will likely have increasingly unconfident softmax outputs corresponding to the true label class.

\subsection{Predictive Uncertainty} 
\label{section:pred_uncertainty}
We use Shannon Entropy to quantify the uncertainty in instance-specific teacher predictions $p(y | \boldsymbol{x}; \boldsymbol{w}_i)$, averaged over the training set, which we call ``Average Predictive Uncertainty,'' and define as: 
\begin{align}
    \mathbb{E}_{\boldsymbol{x}} \left[ H\left(p(\cdot | \boldsymbol{x}; \boldsymbol{w}_i) \right) \right] \approx \frac{1}{n} \sum_{j=1}^n H\left(p( \cdot  | \boldsymbol{x}_j; \boldsymbol{w}_i)\right) = \frac{1}{n} \sum_{j=1}^n \sum_{c=1}^k -p(y_c | \boldsymbol{x}_j; \boldsymbol{w}_i) \log p(y_c | \boldsymbol{x}_j; \boldsymbol{w}_i).
\end{align}
Note that previous literature~\cite{dubey2018maximum, pereyra2017regularizing} has also proposed to use the above measure as a regularizer to prevent over-confident predictions. Label smoothing~\cite{pereyra2017regularizing, szegedy2016rethinking} is a closely related technique that also penalizes over-confident predictions by explicitly smoothing out ground-truth labels. A detailed discussion on the relationship between the two can be found in Appendix~\ref{appendix:entropy_reg}.

\subsection{Confidence Diversity} 
\label{section:diversity}
Average Predictive Uncertainty is insufficient to fully capture the variability associated with teacher predictions. 
In this paper, we argue it is also important to consider the amount of spreading of teacher predictions over the probability simplex among different (training) samples. 
For instance, two teachers can have very similar Average Predictive Uncertainty values, but drastically different amounts of spread on the probability simplex if the softmax predictions of one teacher are much more diverse among different samples than the other. 
We coin this population spread in predictive probabilities ``Confidence Diversity." 
As we show below, characterizing the Confidence Diversity can be important for understanding teacher-student training. 

The differential entropy\footnote{This is distinct from the average predictive uncertainty discussed in the previous section, which measures the average Shannon entropy of probability vectors.} over the entire probability simplex is a natural measure to quantify the confidence diversity. However, accurate entropy estimation can be challenging, and its computation is severely hampered by the curse of dimensionality, particularly in applications with a large number of classes.
To alleviate the problem, in this paper, we propose to measure only the entropy of the softmax element corresponding to the true label class, thereby simplifying the measure to a one-dimensional entropy estimation task. 
Mathematically, if we denote $c = \phi(\boldsymbol{x}, y) \coloneqq [\text{softmax}\big(f_{\boldsymbol{w}}(\boldsymbol{x})\big)]_y$, and let $p_{C}$ be the probability density function of the random variable $C \coloneqq \phi(\boldsymbol{X}, Y)$ where $(\boldsymbol{X}, Y) \sim p(\boldsymbol{x}, y)$, then, we quantify Confidence Diversity via the differential entropy of $C$:
\begin{align}
    h(C) = -\int p_{C}(c) \log p_{C}(c) \, dc.
\end{align}
% \begin{align}
%     h(Z|Y) &= -\sum_{y \in \mathcal{Y}} p\left(y\right) \int p_{Z}(z|y) \log p_{Z}(z|y) \,dz \\
%     &= -\sum_{y \in \mathcal{Y}} p\left(y\right) \int p_{Z}\left(\phi(\boldsymbol{x}, y)|y \right) \log p_{Z}\left(\phi(\boldsymbol{x}, y)|y \right) \,d \boldsymbol{x}
% \end{align}
We use the KNN-based entropy estimator to compute $h(C)$ over the training set~\cite{beirlant1997nonparametric}. 
In essence, the above measure quantifies the amount of spread associated with the teacher predictions on the true label class. The smaller the value, the more similar the softmax values are across different samples. 
\begin{figure}
\centering
\includegraphics[width=1\linewidth]{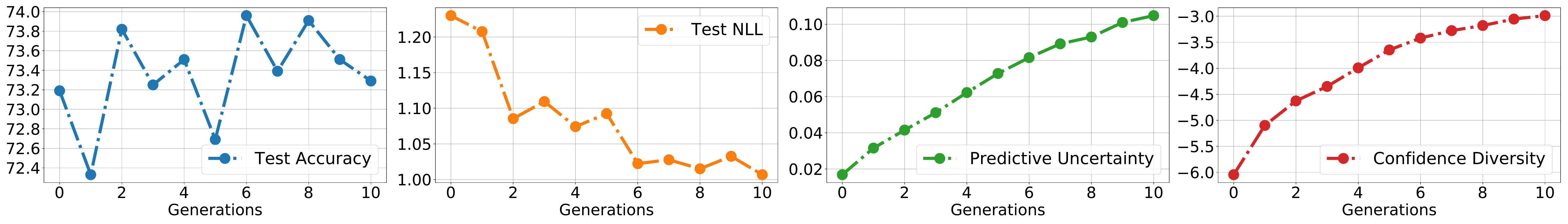}
\caption{Results for sequential self-distillation over 10 generations are shown above. Model obtained at the $(i-1)$-th generation is used as the teacher model for training at the $i$-th generation. Accuracy and NLL are obtained on the test set using the student model, whereas the predictive uncertainty and confidence diversity are evaluated on the training set with teacher predictions.}
\label{fig:BAN_results}
\end{figure}
\subsection{Sequential Self-Distillation Experiment} 

We perform sequential self-distillation with ResNet-34 on the CIFAR-100 dataset for 10 generations. At each generation, we train the neural networks for 150 epochs using the identical optimization procedure as in the original ResNet paper~\cite{he2016deep}. Following Furlanello et al.~\cite{furlanello2018born}, $\alpha$ and $T$ are set to $0$ and $1$ respectively throughout the entire training process. Additional experiments with different values of $T$ can be found in Appendix~\ref{appendix:sequential_temp}. Fig.~\ref{fig:BAN_results} summarizes the results. As indicated by the general increasing trend in test accuracy, sequential distillation indeed leads to improvements. 
The entropy plots also support the hypothesis that subsequent generations exhibit increasing diversity and uncertainty in predictions. 
Despite the same increasing trend, the two entropy metrics quantify different things. 
The increase in average predictive uncertainty suggests overall a drop in the confidence of the categorical distribution, while the growth in confidence diversity suggests an increasing variability in teacher predictions.  Interestingly, we also see obvious improvements in terms of NLL, suggesting in addition that BAN can improve calibration of predictions~\cite{guo2017calibration}.

\begin{figure}
\centering
\includegraphics[width=1\linewidth]{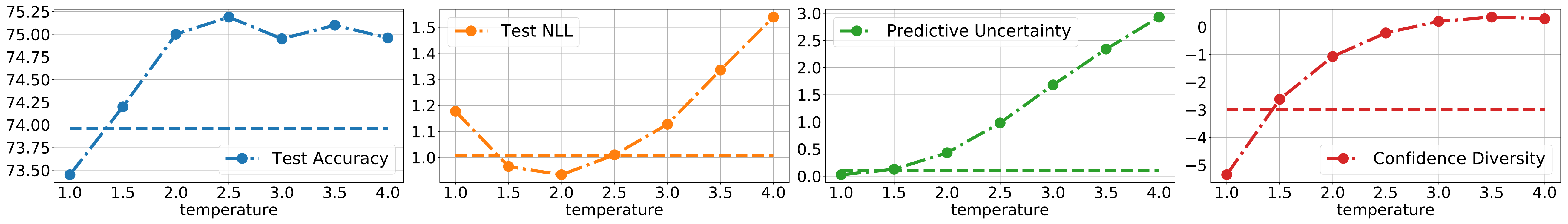}
\caption{Results with teacher predictions scaled by varying temperature $T$. The flat lines in the plots correspond to the largest/smallest values achieved over 10 generations of sequential distillation with $T=1$ in the previous experiments for accuracy, predictive uncertainty and confidence diversity/NLL.}
\label{fig:temp_scaling_results}
\end{figure}

To further study the apparent correlation between student performance and entropy of teacher predictions over generations, we conduct a new experiment, where we instead train a single teacher.
This teacher is then used to train a single generation of students while varying the temperature hyper-parameter $T$ in Eq.~\ref{eqn:distillation}, which explicitly adjusts the uncertainty and diversity of teacher predictions. 
For consistency, we keep $\alpha = 0$. 
Results are illustrated in Fig.~\ref{fig:temp_scaling_results}. As expected, increasing $T$ leads to greater predictive uncertainty and diversity in teacher predictions. 
Importantly, we see this increase leads to drastic improvements in the test accuracy of students. 
In fact, the gain is much greater than the best achieved with 10 generations of BAN with $T=1$ (indicated with the flat line in the plot).
%which did not use likely due to a considerably smaller entropy obtained with BAN. 
The identified correlation is consistent with the recent finding that early-stopped models, which typically have much larger entropy than fully trained ones, serve as better teachers~\cite{cho2019efficacy}.
Lastly, we also see improvements in NLL with increasing entropy of teacher predictions. However, too high $T$ leads to a subsequent increase in NLL, likely due to teacher predictions that lack in confidence.

A closer look at the entropy metrics of the above experiment reveals an important insight. While the average predictive uncertainty is strictly increasing with $T$, the confidence diversity plateaus after $T = 2.5$. 
The plateau of confidence diversity coincides closely with the stagnation of student test accuracy, hinting at the importance of confidence diversity in teacher predictions. 
The apparent correlation between accuracy and confidence diversity can be also seen from the additional sequential self-distillation experiments found in Appendix~\ref{appendix:sequential_temp}.
This makes intuitive sense.
Given a training set, we would expect that some of the samples be much more typically representative of the label class than others. 
Ideally, we would hope to classify the typical examples with much greater confidence than an ambiguous example of the same class. 
Previous results show that training with such instance-specific uncertainty can indeed lead to better performance~\cite{peterson2019human}.
Our view is that in self-distillation, the teacher provides the means for instance-specific regularization.
%We believe that this can be especially important for over-parametrized models like NNs which can overfit arbitrarily well to the training set. 

% Hence, diversity in probabilistic labels to regularize the relative confidence of samples can be crucial for better generalization performance. 
% This is especially important for over-parametrized models like NNs which can overfit arbitrarily well to the training set. Previous results show that training with this additional uncertainty information can indeed lead to better performance~\cite{peterson2019human}. As a result, the teacher-student training procedure, which provides uncertainty information through a pre-trained teacher model, can be in fact seen as instance-specific regularization that promotes the diversity of the probability softmax output. The amount of regularization for each sample is determined by the teacher predictions. 

\section{An Amortized MAP Perspective of Self-Distillation}
\label{section:map}
The instance-specific regularization perspective on self-distillation motivates us to recast the training procedure as performing Maximum a posteriori (MAP) estimation on the softmax probability vector. Specifically, suppose now that the likelihood $p(y | \boldsymbol{x}, \boldsymbol{z}) = \text{Cat}(\boldsymbol{z})$ be a categorical distribution with parameter $\boldsymbol{z} \in \Delta(L)$ and the conditional prior $p(\boldsymbol{z} | \boldsymbol{x}) = \text{Dir}(\boldsymbol{\alpha}_{\boldsymbol{x}})$ be a Dirichlet distribution with instance-specific parameter $\boldsymbol{\alpha}_{\boldsymbol{x}}$. 
% As we describe below, this prior can be computed by a (teacher) neural network.
% % By conjugacy of the Dirichlet prior, the posterior distribution $p(\boldsymbol{z} | \boldsymbol{x}, y) = \text{Dir}(\boldsymbol{\alpha}_{\boldsymbol{x}} + \boldsymbol{c})$ is also a Dirichlet distribution, where $\boldsymbol{c}_i$ corresponds to number of occurrences of the $i$-th category.
% In most of the modern machine learning problems, typically only one observation of label $y$ is available for a sample $\boldsymbol{x}$, making the above model unuseful. Instead, we can amortize the conditional posterior distribution with a neural network and
% Then, performing MAP estimate of $\boldsymbol{z}$ reduces to solving following optimization problem for each sample $\boldsymbol{x}_i$:
Due to conjugacy of the Dirichlet prior, a closed-form solution of $\hat{\boldsymbol{z}}_i = \frac{\boldsymbol{c}_i + \boldsymbol{\alpha_x}_i - 1}{ \sum_j \boldsymbol{c}_j + \boldsymbol{\alpha_x}_j - 1}$, where $\boldsymbol{c}_i$ corresponds to number of occurrences of the $i$-th category, can be easily obtained.
% \begin{align}
%     \hat{\boldsymbol{z}_i} = \max_{\boldsymbol{z} \in \Delta(L)} \log p(\boldsymbol{z}|\boldsymbol{x}_i,y_i;\boldsymbol{w}, \boldsymbol{\alpha_x}) &= \max_{\boldsymbol{z} \in \Delta(L)} \log p(y = y_i|\boldsymbol{z},\boldsymbol{x}_i; \boldsymbol{w}) + \log p(\boldsymbol{z}|\boldsymbol{x}_i;\boldsymbol{w}, \boldsymbol{\alpha_x}) \nonumber  \\
%     &=\max_{\boldsymbol{z} \in \Delta(L)}  \log \boldsymbol{z}_{y_i} + 
%   \sum^k_{c=1} ([\boldsymbol{\alpha}_{\boldsymbol{x}_i}]_{c} - 1) \log [\boldsymbol{z}]_{c}.
% \end{align}

The above framework is not useful for classification when given a new sample $\boldsymbol{x}$ without any observations $y$. Moreover, in the common supervised learning setup, only one observation of label $y$ is available for each sample $\boldsymbol{x}$. The MAP solution shown above merely relies on the provided label $y$ for each sample $\boldsymbol{x}$, without exploiting the potential similarities among different samples $(\boldsymbol{x}_i)$'s in the entire dataset for more accurate estimation.
For example, we could have different samples that are almost duplicates (cf.~\cite{barz2019we}), but have different $y_i$'s, which could inform us about other labels that could be drawn from $\boldsymbol{z}_i$. 
Thus, instead of relying on the instance-level closed-form solution, we can train a (student) network to amortize the MAP estimation $\hat{\boldsymbol{z}}_i \approx \text{softmax}\big(f_{\boldsymbol{w}}(\boldsymbol{x}_i)\big)$ with a given training set, resulting in an optimization problem of:
\begin{align}
    \max_{\boldsymbol{w}} \sum_{i=1}^n \log p(\boldsymbol{z}|\boldsymbol{x}_i,y_i;\boldsymbol{w}, \boldsymbol{\alpha_x}) 
    &= \max_{\boldsymbol{w}} \sum_{i=1}^n \log p(y = y_i|\boldsymbol{z},\boldsymbol{x}_i; \boldsymbol{w}) + \log p(\boldsymbol{z}|\boldsymbol{x}_i;\boldsymbol{w}, \boldsymbol{\alpha_x}) \nonumber \\
    % &= \max_{\boldsymbol{w}} \frac{1}{n}\sum_{i=1}^n \left( \log \boldsymbol{z}^{(y_i)} +  \log \Gamma \left( \sum^k_{c=1} \boldsymbol{\alpha}_{\boldsymbol{x}_i}^{(c)} \right) - \sum^k_{c=1} \log \Gamma \left( \boldsymbol{\alpha}_{\boldsymbol{x}_i}^{(c)} \right) + \sum^k_{c=1} (\boldsymbol{\alpha}_{\boldsymbol{x}_i}^{(c)} - 1) \log \boldsymbol{z}^{(c)} \right)\\
    &= \max_{\boldsymbol{w}} \underbrace{\sum_{i=1}^n \log [\text{softmax}\left(f_{\boldsymbol{w}}(\boldsymbol{x}_i)\right)]_{y_i}}_\text{Cross entropy} + 
    \underbrace{\sum_{i=1}^n \sum^k_{c=1} ([\boldsymbol{\alpha}_{\boldsymbol{x}_i}]_{c} - 1) \log [\boldsymbol{z}]_{c}}_\text{Instance-specific regularization}.
    \label{eq:MAP1}
\end{align}
Eq.~\ref{eq:MAP1} is an objective that provides us with a function to obtain a MAP solution of $\boldsymbol{z}$ given an input sample $\boldsymbol{x}$. 
Note that, we do not make any assumptions about the availability or number of label observations of $y$ for each sample $\boldsymbol{x}$. 
%As such, $y$'s are not used as inputs to the student network.
This enables us to find an approximate MAP solution to $\boldsymbol{x}$ at test-time when $\boldsymbol{\alpha_x}$ and $y$ are unavailable.
The resulting framework can be generally applicable to various scenarios like semi-supervised learning or learning from multiple labels per sample. 
Nevertheless, in the following, we restrict our attention to supervised learning with a single label per training sample. 

\subsection{Label Smoothing as MAP}
The difficulty now lies in obtaining the instance-specific prior $\text{Dir}(\boldsymbol{\alpha}_{\boldsymbol{x}})$. A naive independence assumption that $p(\boldsymbol{z} | \boldsymbol{x}) = p(\boldsymbol{z})$ can be made. Under such an assumption, a sensible choice of prior would be a uniform distribution across all possible labels. Choosing $[\boldsymbol{\alpha}_{\boldsymbol{x}}]_{c} = [\boldsymbol{\alpha}]_{c} = \frac{\beta}{k} + 1$ for all $c \in \{ 1, ..., k \}$ for some hyper-parameter $\beta$, the MAP objective becomes
\begin{align}
    \mathcal{L}_{LS} = \sum_{i=1}^n -\log [\boldsymbol{z}]_{y_i} + \beta \, \sum_{i=1}^n \sum^k_{c=1} -\frac{1}{k} \log [\boldsymbol{z}]_{c}.
    \label{eq:LS}
\end{align}
As noted in prior work, this loss function is equivalent to the commonly used label smoothing (LS) regularization~\cite{pereyra2017regularizing, szegedy2016rethinking} (derivations can be found in Appendix~\ref{appendix:entropy_reg}). Observe also that the training objective in essence promotes predictions with larger predictive uncertainty, but not confidence diversity.

\subsection{Self-Distillation as MAP}
A better instance-specific prior distribution can be obtained using a pre-trained (teacher) neural network. Let us consider a network $f_{\boldsymbol{w}_t}$ trained with the regular MLE objective, by maximizing $p(y | \boldsymbol{x}; \boldsymbol{w}_t) = \text{Cat}(
\text{softmax}\big(f_{\boldsymbol{w}_t}(\boldsymbol{x})\big)$, where $[\text{softmax}\left(f_{\boldsymbol{w}_t}(\boldsymbol{x})\right)]_i = \frac{[\text{exp}( f_{\boldsymbol{w}_t}(\boldsymbol{x}) )]_i}{\sum_j [\text{exp}( f_{\boldsymbol{w}_t}(\boldsymbol{x}))]_j}$.
Now, due to conjugacy of the Dirichlet prior, the marginal likelihood $p(y| \boldsymbol{x}; \boldsymbol{\alpha_x})$ is a Dirichlet-multinomial distribution~\cite{minka2000estimating}. In the case of single label observation considered, the marginal likelihood reduces to a categorical distribution. As such, we have:
$p(y| \boldsymbol{x}; \boldsymbol{\alpha_x}) = \text{Cat}(\overline{\boldsymbol{\alpha_x}})$, where $\overline{\boldsymbol{\alpha_x}}$ is normalized such that $[\overline{\boldsymbol{\alpha_x}}]_i  = \frac{[\boldsymbol{\alpha_x}]_i}{\sum_j [\boldsymbol{\alpha_x}]_j}$.
%\boldsymbol{\mu}_{\boldsymbol{x}})$ where  $\boldsymbol{\mu}_{\boldsymbol{x}} = \text{softmax}\big(f_{\boldsymbol{w}_t}(\boldsymbol{x})\big)$. 
We can thus interpret $\text{exp}\left(f_{\boldsymbol{w}_t}(\boldsymbol{x})\right)$ as the parameters of the Dirichlet distribution to obtain a useful instance-specific prior on $\boldsymbol{z}$. 
However, we observe that there is a scale ambiguity that needs resolving, since any of the following will yield the same $\overline{\boldsymbol{\alpha_x}}$:
% $\text{Dir}\left(\text{exp}\left(f_{\boldsymbol{w}_t}(\boldsymbol{x})\right) \right)$ is problematic, as these distributions are typically over-concentrated at the label class, rendering the regularizer futile. Instead, we can flatten the distribution with some temperature scaling term $T > 1$ and an additive smoothing term $\gamma > 0$, yielding an instance-specific prior 
\begin{align}
    \boldsymbol{\alpha}_{\boldsymbol{x}} =  \beta \, \text{exp}(f_{\boldsymbol{w}_t}(\boldsymbol{x}) / T) + \gamma,
\end{align}
where $T=1$ and $\gamma=0$, and $\beta$ corresponds to some hyper-parameter.
Using $T>1$ and $\gamma>0$ corresponds to flattening the prior distribution, which we found to be useful in practice - an observation consistent with prior work. Note that in the limit of $T \rightarrow \infty$, the instance-specific prior reduces to a uniform prior corresponding to classical label smoothing. 
%The larger the $T$ and $\gamma$, the closer the mean of Dirichlet is to the uniform distribution. 
Setting $\gamma = 1$ (we also experimentally explore the effect of varying $\gamma$. See Appendix~\ref{appendix:gamma} for details), we obtain 
\begin{align}
    \boldsymbol{\alpha}_{\boldsymbol{x}} &=   \beta \, \text{exp}(f_{\boldsymbol{w}_t}(\boldsymbol{x}) / T) + 1 =  \beta \, \sum_j [\text{exp}(f_{\boldsymbol{w}_t}(\boldsymbol{x}) / T)]_j \, \text{softmax}(f_{\boldsymbol{w}_t}(\boldsymbol{x}) / T) + 1.
    \label{eq:alpha}
\end{align}
Plugging this into Eq.~\ref{eq:MAP1} yields 
\begin{align}
    \mathcal{L}_{SD} = \sum_{i=1}^n -\log [\boldsymbol{z}]_{y_i} + \beta \, \sum_{i=1}^n \omega_{\boldsymbol{x}_i} \sum^k_{c=1} - [\text{softmax}(f_{\boldsymbol{w}_t}(\boldsymbol{x}_i) / T)]_c \log [\boldsymbol{z}]_{c},
\end{align}
very similar to the distillation loss of Eq.~\ref{eqn:distillation}, with an additional sample-specific weighting term $\omega_{\boldsymbol{x}_i} = \sum_j [\text{exp}(f_{\boldsymbol{w}_t}(\boldsymbol{x}_i) / T)]_j$! 

Despite the interesting result, we empirically observe that, with temperature values $T$ found to be useful in practice, the relative weightings of samples are too close to yield a significant difference from regular distillation loss. 
%For smaller temperatures used, on the other hand, there will be a handful of samples with orders of magnitude bigger weighting than the rest, resulting in ineffective regularization. 
Hence, for all of our experiments, we still adopt the distillation loss of Eq.~\ref{eqn:distillation}. However, we believe that, with teacher models trained with an objective more appropriate than MLE, the difference might be bigger. 
We hope to explore alternative ways of obtaining teacher models to effectively utilize the sample re-weighted distillation objective as future work. 

The MAP interpretation, together with empirical experiments conducted in Section~\ref{section:BAN}, suggests that multi-generational self-distillation can in fact be seen as an inefficient approach to implicitly flatten and diversify the instance-specific prior distribution. 
Our experiments suggest that instead, we can more effectively tune for hyper-parameters $T$ and $\gamma$ to achieve similar, if not better, results.
Moreover, from this perspective, distillation in general can be understood as a regularization strategy. Some empirical evidence for this can be found in Appendix~\ref{appendix:train_size} and \ref{appendix:weight_decay}.

\subsection{On the Relationship between Label Smoothing and Self-Distillation}
The MAP perspective reveals an intimate relationship between self-distillation and label smoothing. 
Label smoothing increases the uncertainty of predictive probabilities. 
However, as discussed in Section~\ref{section:BAN}, this might not be enough to prevent overfitting, as evidenced by the stagnant test accuracy despite increasing uncertainty in Fig.~\ref{fig:temp_scaling_results}. 
Indeed, the MAP perspective suggests that, ideally, each sample should have a distinct probabilistic label.
Instance-specific regularization can encourage confidence diversity, in addition to predictive uncertainty. 

While the predictive uncertainty can be explicitly used for regularization as previously discussed in Section~\ref{section:pred_uncertainty}, we observe empirically that promoting confidence diversity directly through the proposed measure in Section~\ref{section:diversity} can be hard in practice, yielding unsatisfactory results. This could have been caused by difficulty in estimating confidence diversity accurately using mini-batch samples. Naively promoting confidence diversity during the early stage of training could also have harmed learning. As such, we can view distillation as an indirect way of achieving this objective. We leave it as future work to further explore alternative techniques to enable direct regularization of confidence diversity. 

\section{Beta Smoothing Labels}
Self-distillation requires training a separate teacher model. In this paper, we propose an efficient enhancement to label smoothing strategy where the amount of smoothing will be proportional to the uncertainty of predictions. Specifically, we make use of the exponential moving average (EMA) predictions as implemented by Tarvainen and Valpola~\cite{tarvainen2017mean} of the model at training, and obtain a ranking based on the confidence (the magnitude of the largest element of the softmax) of predictions at each mini-batch, on the fly, from smallest to largest. Instead of assigning uniform distributions $[\boldsymbol{\alpha}_{\boldsymbol{x}}]_{c} =\frac{\beta}{k} + 1$ for all $c \in \{ 1, ..., k \}$ to all samples as priors, during each iteration, we sample and sort a set of i.i.d. random variables $\{ b_1 \leq ... \leq b_m \}$ from $\textit{Beta}(a, 1)$ where $m$ corresponds to the mini-batch size and $a$ corresponds to the hyper-parameter associated with the Beta distribution. Then, we assign $[\boldsymbol{\alpha}_{\boldsymbol{x}_i}]_{y_i} = \beta b_i + 1$ and $[\boldsymbol{\alpha}_{\boldsymbol{x}_i}]_{c} = \beta \frac{1 - b_i}{k-1} + 1$ for all $c \neq y_i$ as the prior to each sample $\boldsymbol{x}_i$, based on the ranking obtained. In this way, samples with larger confidence obtained through the EMA predictions will receive less amount of label smoothing and vice versa.
Thus, the amount of label smoothing applied to a sample will be proportional to the amount of confidence the model has about that sample's prediction. Those instances that are more challenging to classify will, therefore, have more smoothing applied to their labels.

In practice, for consistency with distillation, Eq.~\ref{eqn:distillation} is used for training. Beta-smoothed labels of $b_i$ on the ground truth class and $\frac{1 - b_i}{k-1}$ on all other classes are used in lieu of teacher predictions for each $\boldsymbol{x}_i$.
Lastly, note that EMA predictions are used in order to stabilize the ranking obtained at each iteration of training. We empirically observe a significant performance boost with the EMA predictions.
We term this method \textit{Beta smoothing}. 

To better examine the role of EMA predictions has on Beta smoothing, we conduct two ablation studies. Firstly, since the EMA predictions are used for Beta smoothed labels, we compare the effectiveness of Beta smoothing against self-training explicitly using the EMA predictions (see Appendix~\ref{appendix:ema} for details). 
Moreover, to test the importance of ranking obtained from EMA predictions, we include in the Appendix~\ref{appendix:beta} an additional experiment for which random Beta smoothing is applied to each sample.
% Beta distribution is chosen due to its support on $[0,1]$ and similarity to the distribution of softmax output of the ground-truth class of teacher predictions. 

Beta smoothing regularization implements an instance-specific prior that encourages confidence diversity, and yet does not require the expensive step of training a separate teacher model. We note that, due to the constantly changing prior used at every iteration of training, Beta smoothing does not, strictly speaking, correspond to the MAP estimation in Eq.~\ref{eq:MAP1}. Nevertheless, it is a simple and effective way to implement the instance-specific prior strategy. As we demonstrate in the following section, it can lead to much better performance than label smoothing. Moreover, unlike teacher predictions which have unique softmax values for all classes, the difference between Beta and label smoothing only comes from the ground-truth softmax element. This enables us to conduct more systematic experiments to illustrate the additional gain from promoting confidence diversity.

\section{Empirical Comparison of Distillation and Label Smoothing}
\label{section:experiment}
To further demonstrate the benefits of the additional regularization on the softmax probability vector space, we design a systematic experiment to compare self-distillation against label smoothing. In addition, experiments on Beta smoothing are also conducted to further verify the importance of confidence diversity, and to promote Beta smoothing as a simple alternative that can lead to better performance than label smoothing at little extra cost. We note that, while previous works have highlighted the similarity between distillation and label smoothing from another perspective~\cite{yuan2019revisit}, we provide a detailed empirical analysis that uncovers additional benefits of instance-specific regularization. 

\subsection{Experimental Setup}
We conduct experiments on CIFAR-100~\cite{krizhevsky2009learning}, CUB-200~\cite{WelinderEtal2010} and Tiny-imagenet~\cite{deng2009imagenet} using ResNet~\cite{he2016deep} and DenseNet~\cite{huang2017densely}. We follow the original optimization configurations, and train the ResNet models for $150$ epochs and DeseNet models for $200$ epochs. $10\%$ of the training data is split as the validation set. All experiments are repeated 5 times with random initialization. For simplicity, label smoothing is implemented with explicit soft labels instead of the objective in Eq.~\ref{eq:LS}. We fix $\epsilon = 0.15$ in label smoothing for all our experiments (additional experiments with $\epsilon =0.1, 0.3$ can be found in the Appendix~\ref{appendix:epsilon_exp}). The hyper-parameter $\alpha$ of Eq.~\ref{eqn:distillation} is taken to be $0.6$ for self-distillation. Only one generation of distillation is performed for all experiments. To systematically decompose the effect of the two regularizations in self-distillation, given a pre-trained teacher and $\alpha$, we manually search for temperature $T$ such that the average effective label of the ground-truth class, $\alpha + (1 - \alpha)[\text{softmax}\left(f_{\boldsymbol{w}_t}(\boldsymbol{x}_i) / T \right)]_{y_i}$, is approximately equal to $0.85$ to match the hyper-parameter $\epsilon$ chosen for label smoothing. 
Eq.~\ref{eqn:distillation} is also used for Beta smoothing with $\alpha = 0.4$. 
The parameter $a$ of the Beta distribution is set such that $\mathbb{E}[\alpha + (1 - \alpha)b_i] = \epsilon$, to make the average probability of ground truth class the same as $\epsilon-$label smoothing. 

We emphasize that the goal of the experiment is to methodically decompose the gain from the two aforementioned regularizations of distillation. 
Note that, both $\alpha$ and $T$ can influence the amount of predictive uncertainty and confidence diversity in teacher predictions at the same time. This coupled effect can make hyper-parameter tuning hard. 
Due to limited computational resources, hyper-parameter tuning is not performed, and the results for all methods can be potentially enhanced. 
Lastly, we also incorporate an additional distillation experiment in which the deeper DenseNet model is used as the teacher model for comparison against self-distillation. Results can be found in Appendix~\ref{appendix:cross-distillation}. 

\begin{figure}
\centering
\begin{subfigure}{.32\textwidth}
\centering
\includegraphics[width=1.05\linewidth]{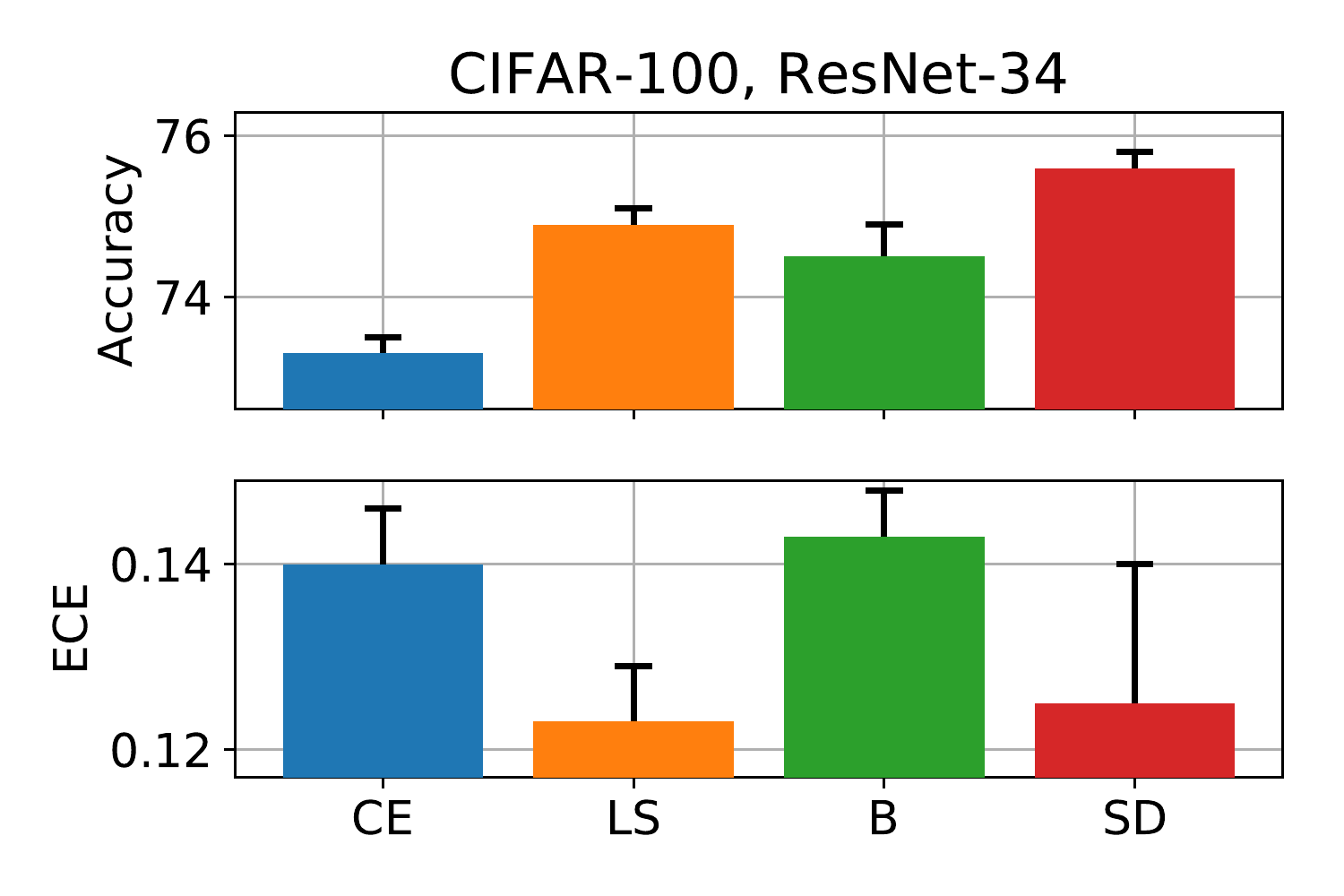}
\end{subfigure}
\begin{subfigure}{.32\textwidth}
\centering
\includegraphics[width=1.05\linewidth]{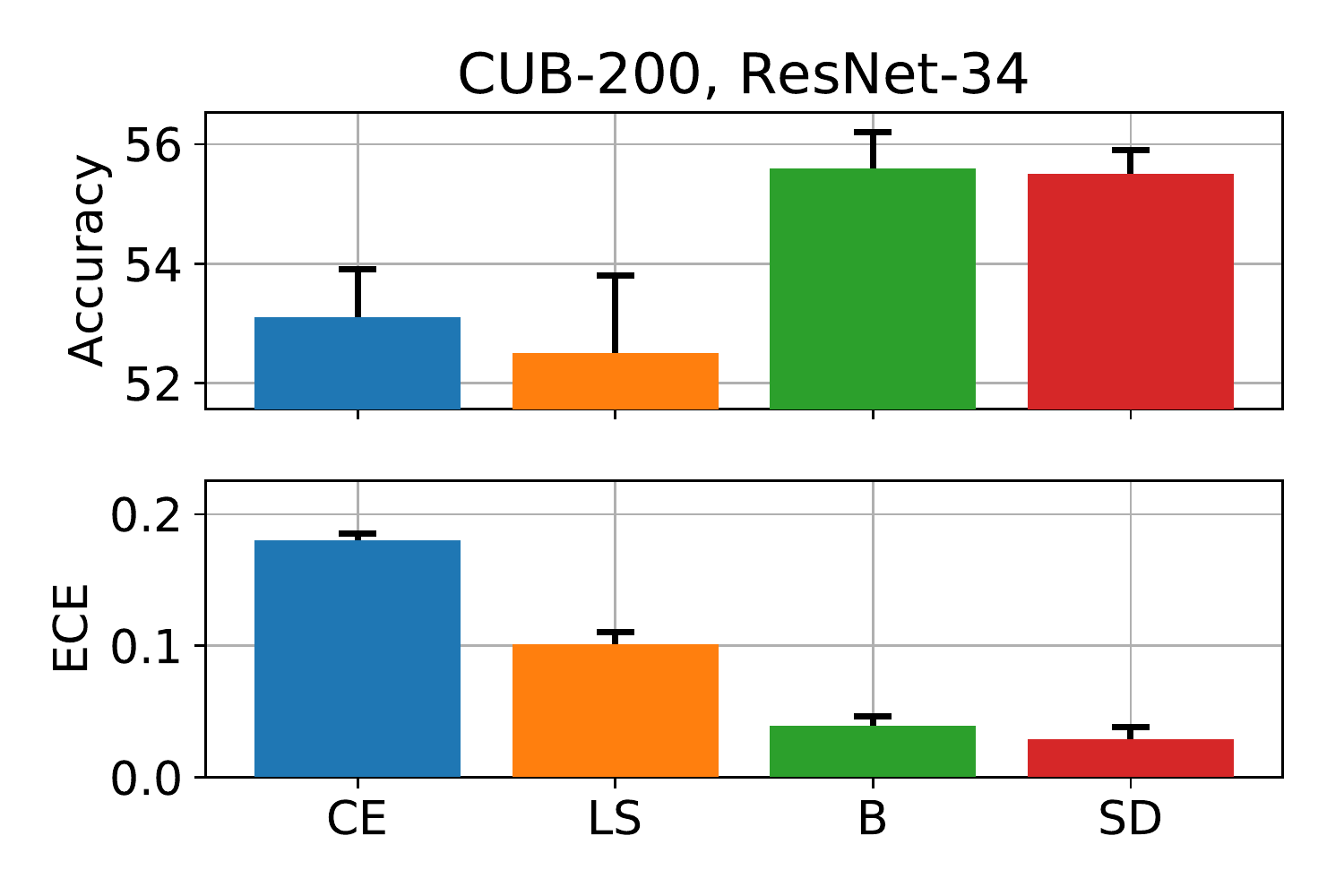}
\end{subfigure}
\begin{subfigure}{.32\textwidth}
\centering
\includegraphics[width=1.05\linewidth]{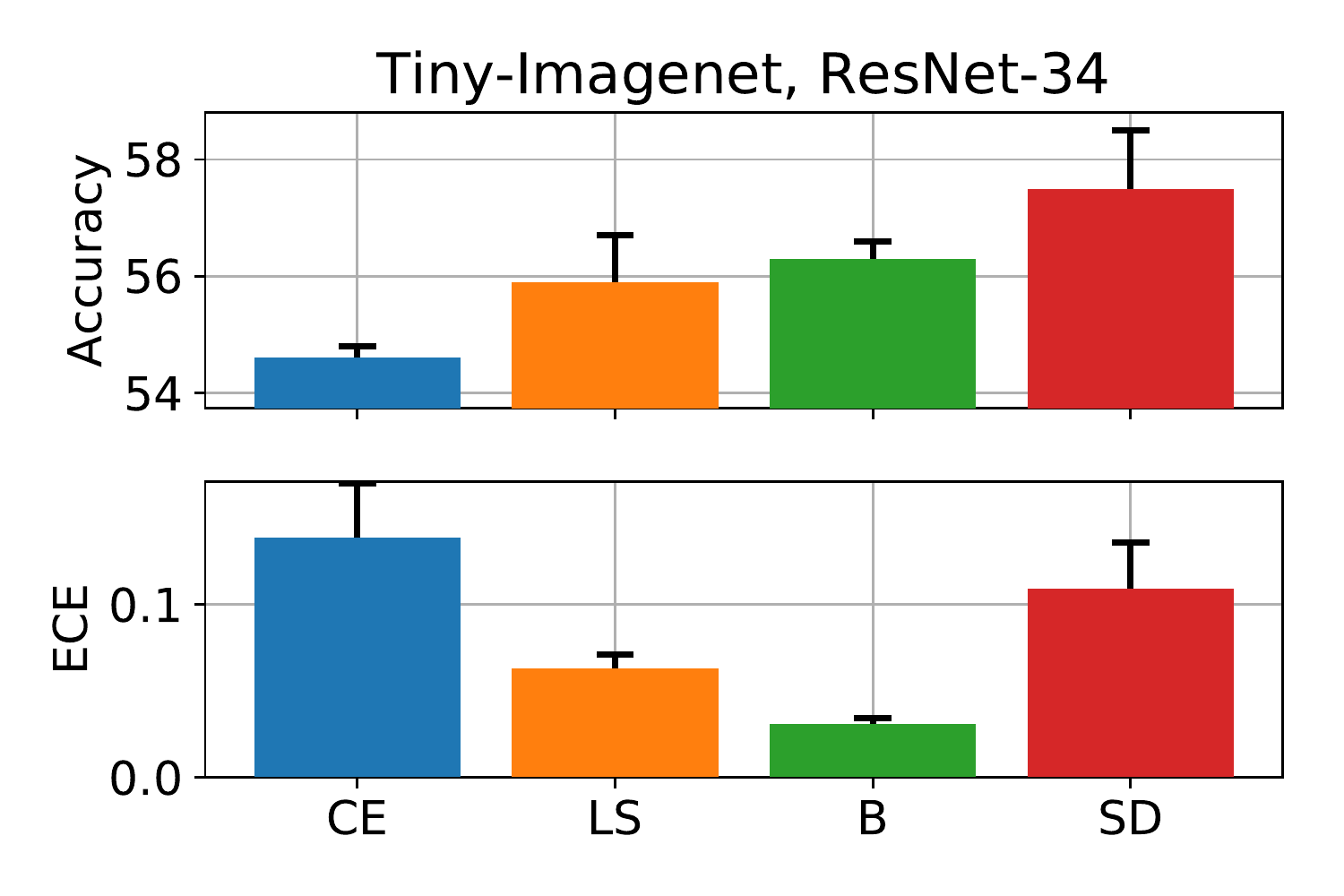}
\end{subfigure}
\begin{subfigure}{.32\textwidth}
\centering
\includegraphics[width=1.05\linewidth]{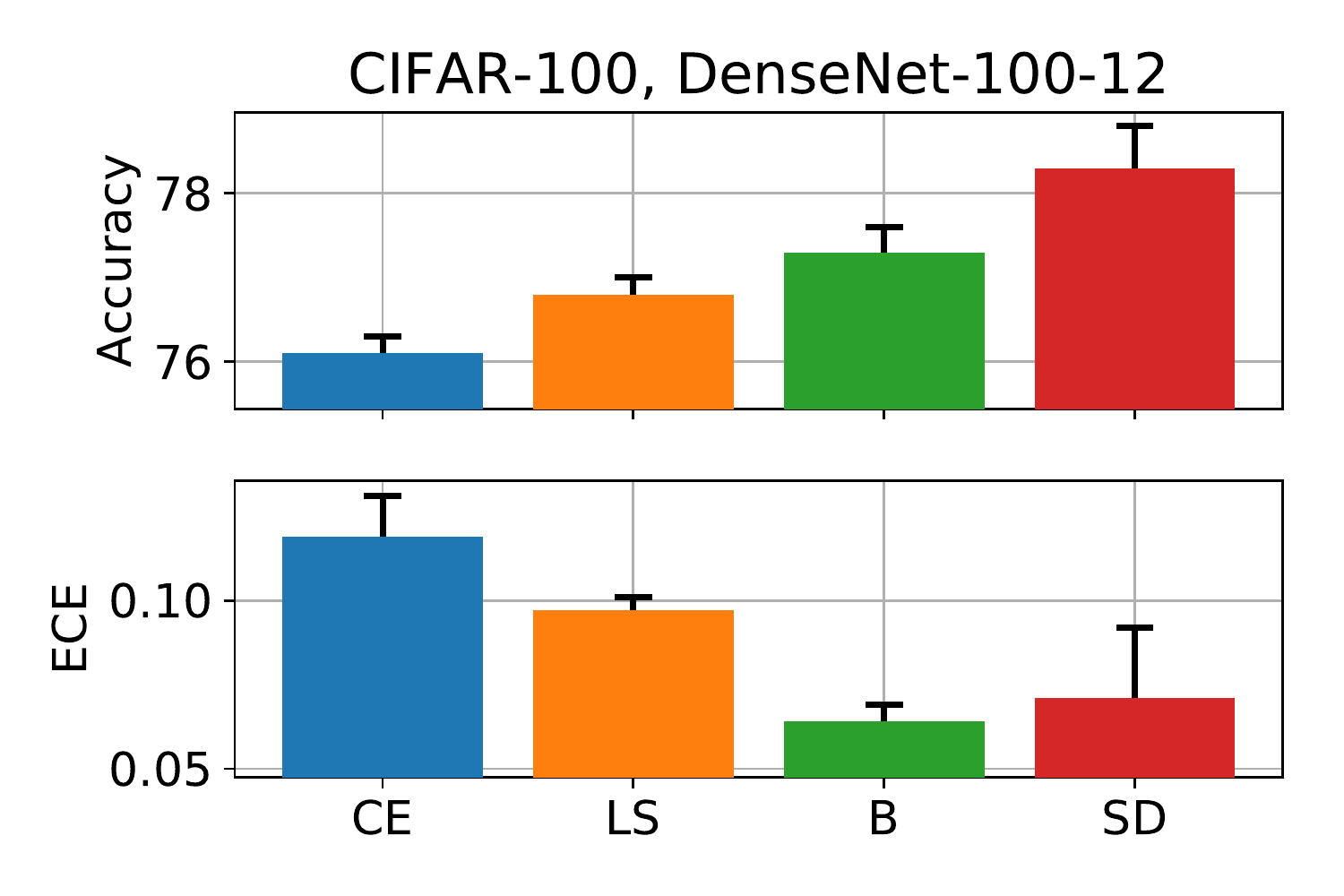}
\end{subfigure}
\begin{subfigure}{.32\textwidth}
\centering
\includegraphics[width=1.05\linewidth]{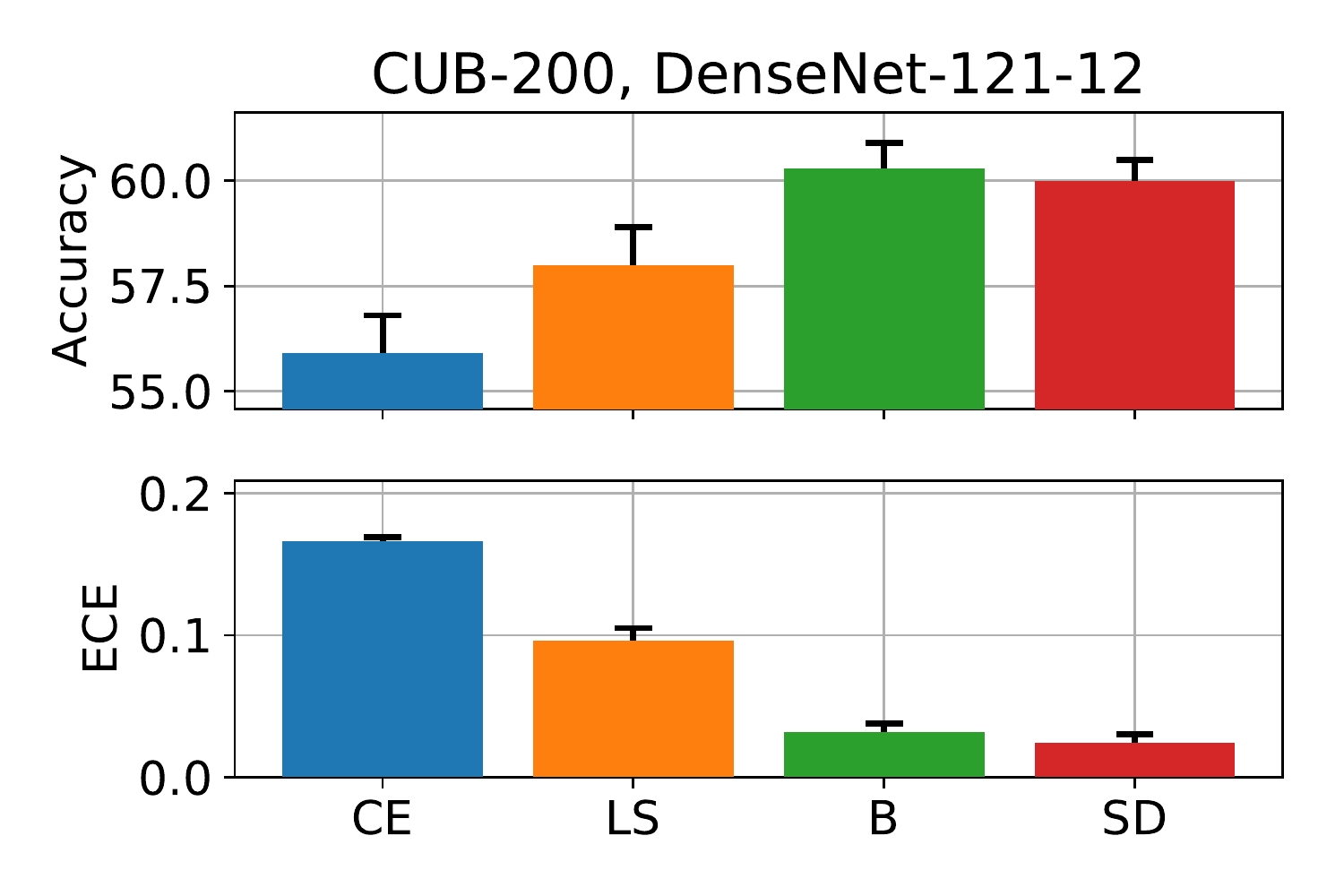}
\end{subfigure}
\begin{subfigure}{.32\textwidth}
\centering
\includegraphics[width=1.05\linewidth]{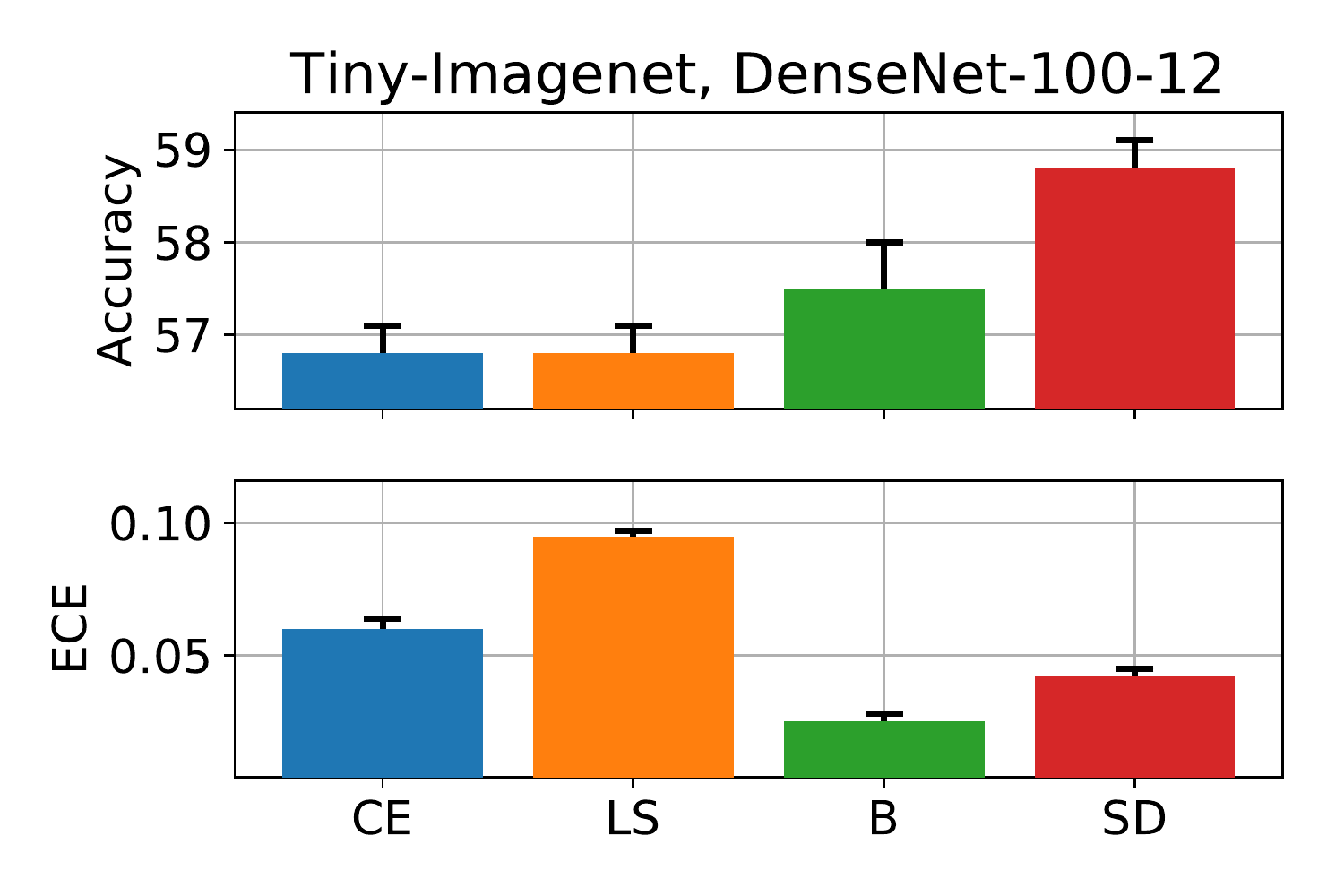}
\end{subfigure}
\caption{Experimental Results performed on CIFAR-100, CUB-200 and the Tiny-Imagenet dataset. "CE", "LS", "B" and "SD" refers to "Cross Entropy", "Label Smoothing", "Beta Smoothing" and "Self-Distillation" respectively. The top rows of each experiment show bar charts of accuracy on test set for each experiment conducted, while the bottom rows are bar charts of expected calibration error.}
\label{fig:results}
\end{figure}

\subsection{Results}
Test accuracies are summarized in the top row for each experiment in Fig.~\ref{fig:results}. Firstly, all regularization techniques lead to improved accuracy compared to the baseline model trained with cross-entropy loss. In agreement with previous results, self-distillation performs better than label smoothing in all of the experiments with our setup, in which the effective degree of label smoothing in distillation  is, on average, the same as that of regular label smoothing. The results suggest the importance of confidence diversity in addition to predictive uncertainty. 
It is worth noting that we obtain encouraging results with Beta smoothing. 
Outperforming label smoothing in all but the CIFAR-100 ResNet experiment, it can even achieve comparable performance to that of self-distillation for the CUB-200 dataset with no separate teacher model required. The improvements of Beta smoothing over label smoothing also serve direct evidence on the importance of confidence diversity, as the only difference between the two is the additional spreading of the ground truth classes.
We hypothesize that the gap in accuracy between Beta smoothing and self-distillation is mainly due to better instance-specific priors set by a pre-trained teacher network. The differences in the non-ground-truth classes between the two methods could also account for the small gap in accuracy performance.

Results on calibration are shown in the bottom rows of Fig.~\ref{fig:results}, where we report the expected calibration error (ECE)~\cite{guo2017calibration}. As anticipated, all regularization techniques lead to enhanced calibration. Nevertheless, we see that the errors obtained with self-distillation are much smaller in general compared to label smoothing. As such, instance-specific priors can also lead to more calibrated models. 
Beta smoothing again not only produces models with much more calibrated predictions compared to label smoothing but compares favorably to self-distillation in a majority of the experiments.

\section{Discussion and Future Directions}
In this paper, we provide empirical evidence that diversity in teacher predictions is correlated with student performance in self-distillation.
Inspired by this observation, we offer an amortized MAP interpretation of the popular teacher-student training strategy. The novel viewpoint provides us with insights on self-distillation and suggests ways to improve it. 
%There are several directions for future research. 
%Firstly, as discussed in Section~\ref{section:map}, the current approach of using a pre-trained model for instance-specific priors can be improved. 
For example, encouraged by the results obtained with Beta smoothing, there are possibly better and/or more efficient ways to obtain priors for instance-specific regularization.
%With the viewpoint of instance-specific prior as a proxy for regularization on the softmax simplex space in mind, we are also interested in investigating alternative ways for regularizing the entropy of the probability simplex.

Recent literature shows that label smoothing leads to better calibration performance~\cite{muller2019does}. In this paper, we demonstrate that distillation can also yield more calibrated models. We believe this is a direct consequence of not performing temperature scaling on student models during training. Indeed, with temperature scaling also on the student models, the student logits are likely pushed larger during training, leading to over-confident predictions. 
%We use the widely adopted expected calibration error (ECE) as the metric to evaluate the calibration of model predictions~\cite{guo2017calibration}.

More generally, we have only discussed the teacher-student training strategy as MAP estimation. There have been other recently proposed techniques involving training with soft labels, which we can interpret as encouraging confidence diversity or implementing instance-specific regularization. 
For instance, the mixup regularization~\cite{zhang2017mixup} technique creates label diversity by taking random convex combinations of the training data, including the labels. Recently proposed consistency-based semi-supervised learning methods such as~\cite{laine2016temporal,tarvainen2017mean}, on the other hand, utilize predictions on unlabeled training samples as an instance-specific prior. We believe this unifying view of regularization with soft labels can stimulate further ideas on instance-specific regularization.

\subsubsection*{Acknowledgments}
This work was supported by NIH R01 grants (R01LM012719 and R01AG053949), the NSF NeuroNex grant 1707312, and NSF CAREER grant (1748377).

\subsubsection*{Statement of the Potential Broader Impact}
In this paper, we offer a new interpretation of the self-distillation training framework, a commonly used technique for improved accuracy used among practitioners in the deep learning community, which allows us to gain some deeper understanding of the reasons for its success. With the ubiquity of deep learning in our society today and countless potential future applications of it, we believe our work can potentially bring positive impacts in several ways.

Firstly, despite the empirical utility of distillation and numerous successful applications in many tasks and applications ranging from computer vision to natural language processing problems, we still lack a thorough understanding of why it works. In our opinion, blindly applying methods and algorithms without a good grasp on the underlying mechanisms can be dangerous. Our perspective offers a theoretically grounded explanation for its success that allows us to apply the techniques to real-world applications broadly with greater confidence.

In addition, the proposed interpretation of distillation as a regularization to neural networks can potentially allow us to obtain models that are more generalizable and reliable. This is an extremely important aspect of applying deep learning to sensitive domains like healthcare and autonomous driving, in which wrong predictions made by machines can lead to catastrophic consequences. Moreover, our new experimental demonstration that models trained with the distillation process can potentially lead to better-calibrated models that can facilitate safer and more interpretable applications of neural networks. Indeed, for real-world classification tasks like disease diagnosis, in addition to accurate predictions, we need reliable estimates of the level of confidence of the predictions made, which is something that neural networks are lacking currently as pointed out by recent research. More calibrated models, in our opinion, enhances the explainability and transparency of neural network models.

Lastly, we believe the introduced framework can stimulate further research on the regularization of deep learning models for better generalization and thus safer applications. It was recently demonstrated that deep neural networks do not seem to suffer from overfitting. Our finding suggests that overfitting can still occur, though in a different way than conventional wisdom, and deep learning can still benefit from regularization. As such, we encourage research into more efficient and principled forms of regularization to improve upon the distillation strategy.

We acknowledge the risks associated with our work. To be more specific, our finding advocates for the use of priors for the regularization of neural networks. Despite the potentially better generalization performance of trained models, depending on the choice of priors used for training, unwanted bias can be inevitably introduced into the deep learning system, potentially causing issues of fairness and privacy.  

\bibliography{reference}
\bibliographystyle{plain}

\newpage

\appendix
\section{Appendix}
\subsection{On Label Smoothing and Predictive Uncertainty Regularization}
\label{appendix:entropy_reg}
We first give a derivation on the equivalence of label smoothing regularization and Eq.~\ref{eq:LS}. With some simple rearrangement of the terms, 
\begin{align*}
    \mathcal{L}_{LS} &= \sum_{i=1}^n -\log [\boldsymbol{z}]_{y_i} + \beta \, \sum_{i=1}^n \sum^k_{c=1} -\frac{1}{k} \log [\boldsymbol{z}]_{c} \\
    % &= -\sum_{i=1}^n \left( \left(1 + \frac{\beta}{k}\right)\log [\boldsymbol{z}]_{y_i} + \sum_{c\neq y_i} \frac{\beta}{k} \log [\boldsymbol{z}]_{c} \right) \\
    &= -\left( 1+\beta \right)\sum_{i=1}^n \left( \frac{k+\beta}{k(1+\beta)} \log [\boldsymbol{z}]_{y_i} + \sum_{c\neq y_i} \frac{\beta}{k(1+\beta)} \log [\boldsymbol{z}]_{c} \right).
\end{align*}
The above objective is clearly equivalent to the label smoothing regularization with $1 - \epsilon = \frac{k+\beta}{k(1+\beta)}$, up to a constant factor of $( 1+\beta)$.

Label smoothing regularizes predictive uncertainty. The amount of regularization is controlled by the amount of smoothing applied. Evidently, the objective does not regularize confidence diversity. Indeed, assuming a NN with capacity capable of fitting the entire training data, predictions on training data will be pushed arbitrarily close to the smoothed soft label. Empirical evidence for this form of overfitting can be seen from experiments done by Müller et al.~\cite{muller2019does}, in which the authors demonstrated that applying label smoothing leads to hampered distillation performance. The authors hypothesize that this is likely due to erasure of "relative information between logits" when label smoothing is applied, hinting at the overfitting of predictions to the smoothed labels. 

A closely related regularization technique is to explicitly regularize on predictive uncertainty:
\begin{align*}
    \mathcal{L}_{PU} &= \sum_{i=1}^n -\log [\boldsymbol{z}]_{y_i} + \beta \, \frac{1}{n} \sum_{j=1}^n \sum_{c=1}^k [\boldsymbol{z}]_{c} \log [\boldsymbol{z}]_{c}.
\end{align*}
Prior papers~\cite{dubey2018maximum, pereyra2017regularizing} have demonstrated that directly regularizing predictive uncertainty can lead to better performance than label smoothing.
However, we note that the above objective does not regularize confidence diversity either. In fact, it can be easily solved, with the method of Lagrange multiplier, that the optima for the objective above is achieved when $[\boldsymbol{z}]_{y_i} = \frac{1}{\beta W\left( \text{exp}(-1/\beta)(k-1) /\beta \right) + 1}$ where $W$ corresponds to the Lambert W function, and $[\boldsymbol{z}]_{c}  =\frac{1 - [\boldsymbol{z}]_{y_i}}{k-1}$ for all $c \neq y_i$, for all sample pairs $(\boldsymbol{x}_i, y_i)$. As such, the global optima obtained by directly regularizing predictive uncertainty is identical to that of label smoothing. 
In practice, differences between the two can arise due to the details of the optimization procedure (like early stopping), and/or due to model capacity.

\newpage
\subsection{Additional Experiments with Temperature Scaling on Student Models}
\label{appendix:student_temp}
\begin{figure}[!htb]
\centering
\begin{subfigure}{.45\textwidth}
\centering
\includegraphics[width=1.0\linewidth]{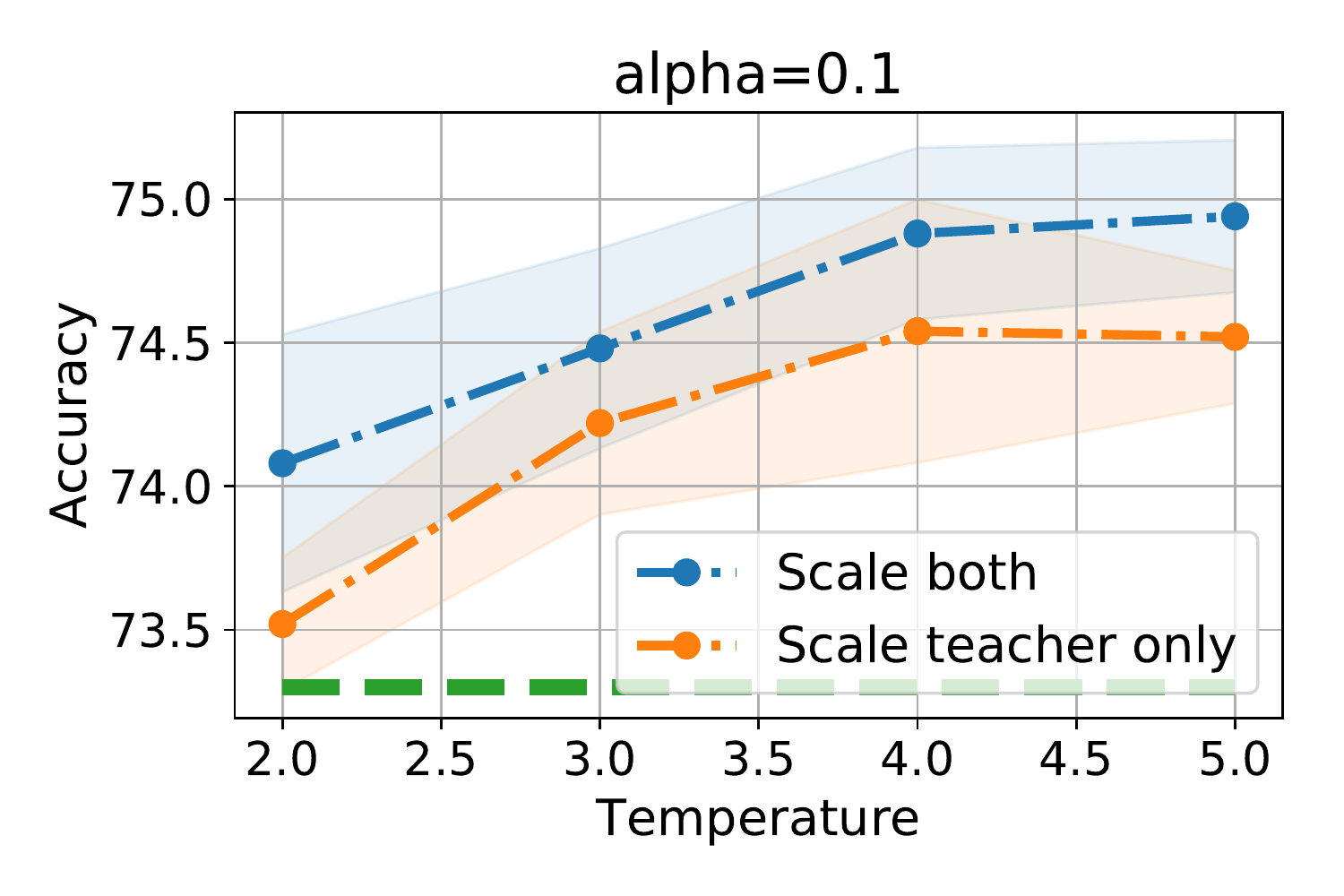}
\end{subfigure}
\begin{subfigure}{.45\textwidth}
\centering
\includegraphics[width=1.0\linewidth]{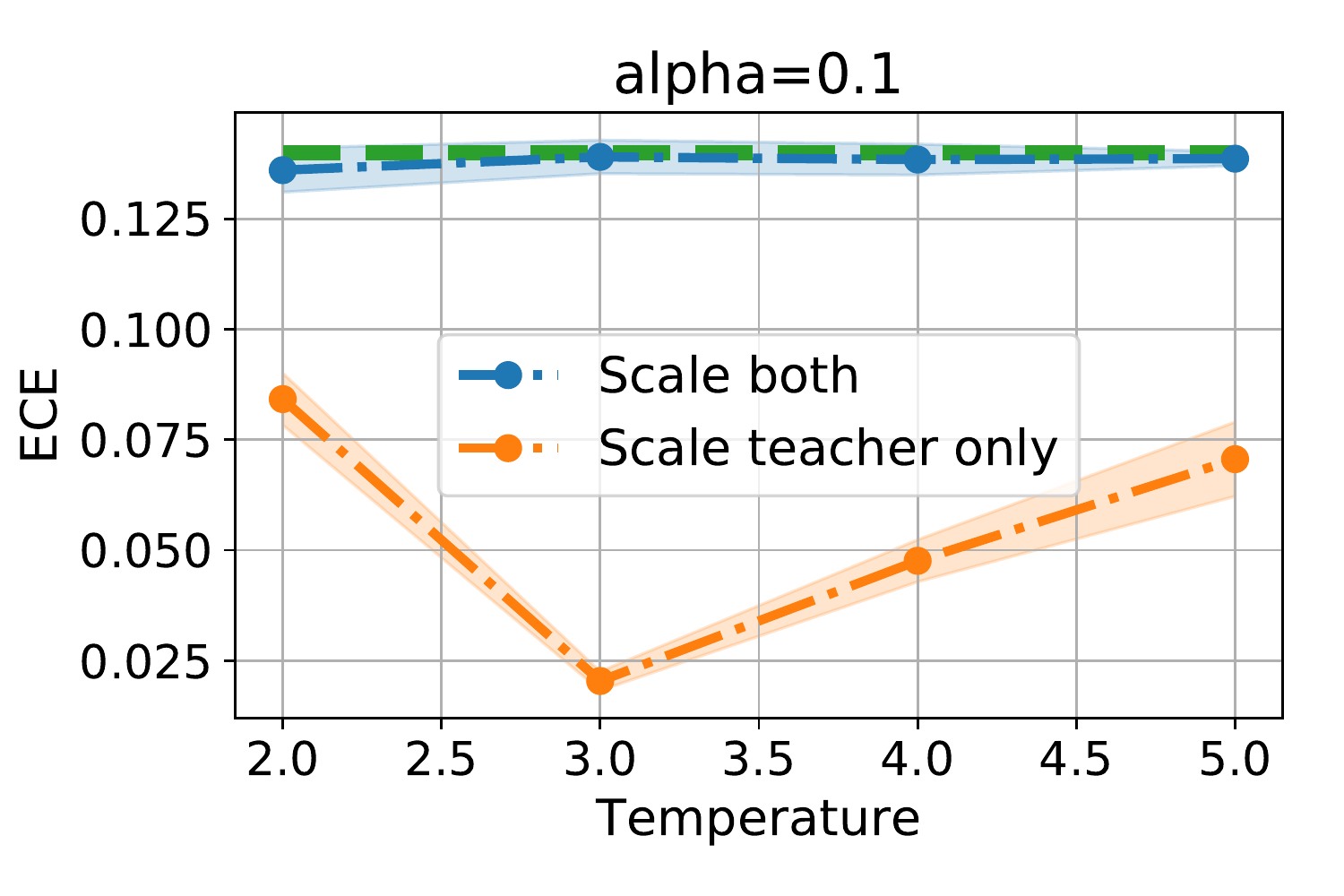}
\end{subfigure}
\begin{subfigure}{.45\textwidth}
\centering
\includegraphics[width=1.0\linewidth]{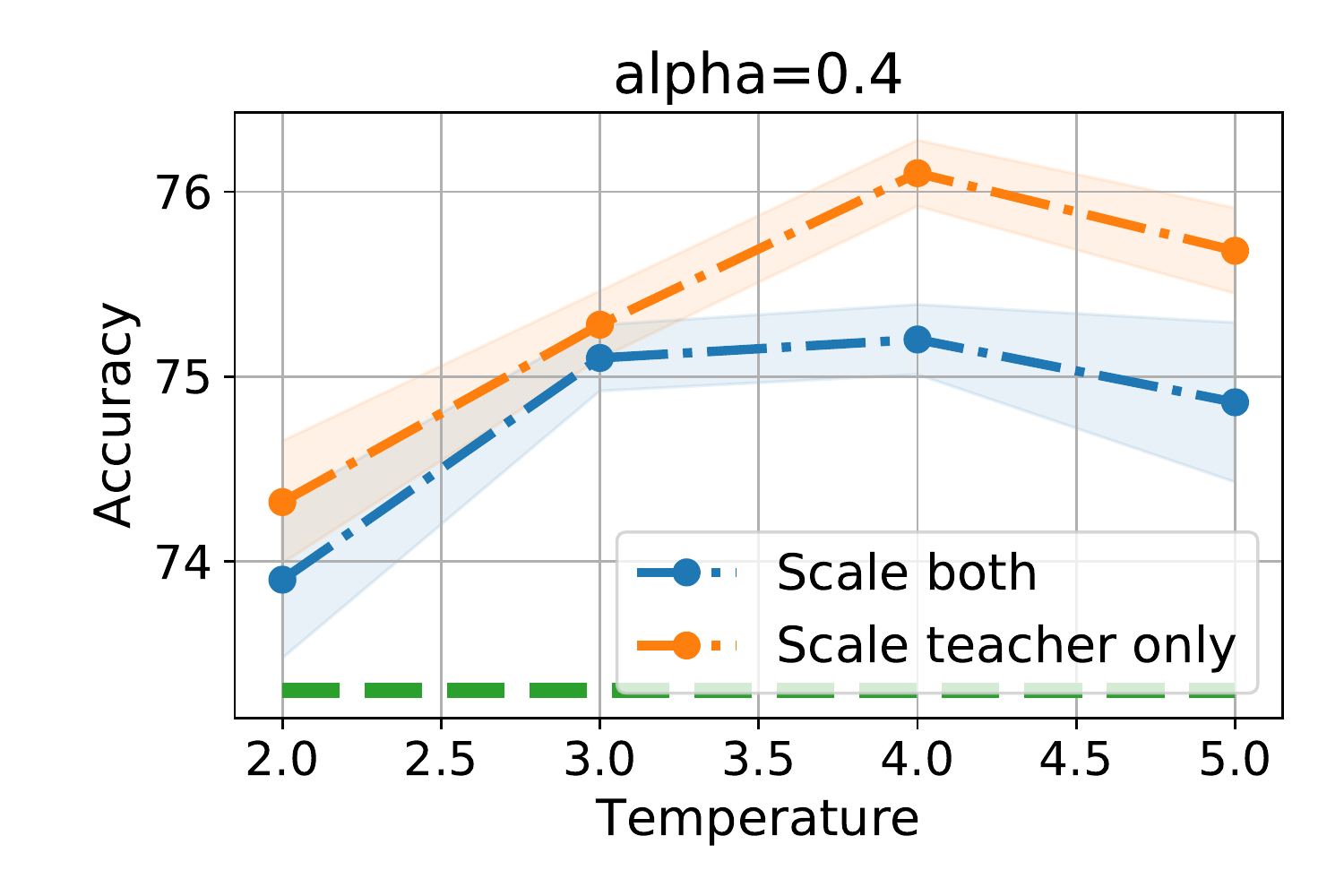}
\end{subfigure}
\begin{subfigure}{.45\textwidth}
\centering
\includegraphics[width=1.0\linewidth]{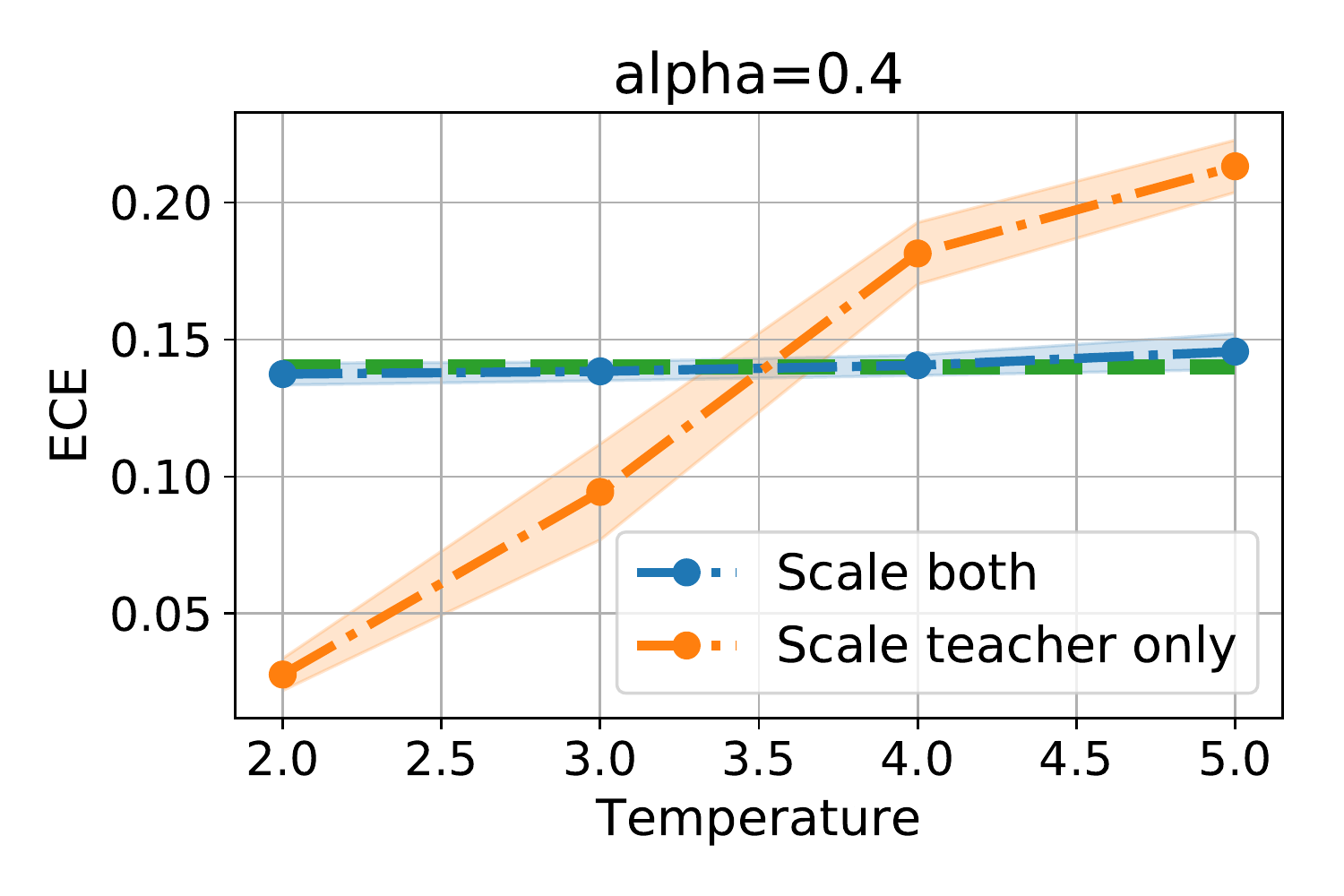}
\end{subfigure}
\caption{\textit{Left:} Test accuracies of ResNet-34 models on the CIFAR-100 dataset when varying temperature. \textit{Right:} ECE of ResNet-34 models on the CIFAR-100 dataset when varying temperature. "Scale both" corresponds to the originally proposed distillation objective in which both teacher and student models are temperature-scaled during training. "Scale teacher only" corresponds to only temperature scaling teacher models during distillation. The \textit{green flat line} represents the performance achieved by the teacher model trained with cross-entropy loss.}
\label{fig:student_scale_results}
\end{figure}

To examine the effect of not applying temperature scaling on student models, we conduct an experiment to compare models trained with and without temperature scaling on student models for distillation loss with the ResNet-34 on the CIFAR-100 dataset, using the training objective of Eq.~\ref{eqn:distillation}. On top of the hyper-parameter $\alpha  = 0.4$ used for experiments in Section~\ref{section:experiment}, we also include results with $\alpha  = 0.1$, a widely used value for knowledge distillation in prior work~\cite{cho2019efficacy}. We vary the amount of temperature scaling applied to illustrate the effect of different temperatures have on student models. 

Plots of test accuracy and ECE against amount of temperature scaling applied are shown in Fig.~\ref{fig:student_scale_results}. Firstly, we observe that models trained with student scaling have ECE almost identical to that of the teacher models. As a direct contrast, we see that the student models trained without student scaling perform much better in terms of calibration error in general over its teacher. Note that the relatively large ECE when $\alpha = 0.4$ and $T > 3$ is likely due to overly unconfident teacher predictions. In addition, we highlight that, with the optimal hyper-parameters of $\alpha$ and $T$ used, student models trained without student scaling can also outperform significantly in terms of test accuracy. We acknowledge that there can be conflicts between the performance of ECE and accuracy, as seen from superior test accuracy but poor ECE achieved for $\alpha=0.4$ and $T=4.0$. In practice, we can use the negative log likelihood, a metric influenced by both ECE and accuracy, to find the optimal $\alpha$ and $T$.  
Lastly, we note that, both $\alpha$ and $T$ alter the amount of predictive uncertainty and confidence diversity in teacher predictions at the same time. This coupled effect could be the reason for the observed conflict between ECE and accuracy. We leave it as a future work to explore alternative ways to decouple the two measures for more efficient and effective parameter search. We believe a decoupled set of parameters can lead to models with better calibration and accuracy at the same time.

\newpage
\subsection{Additional Experiments on Sequential Self-Distillation with Different Temperatures}
\label{appendix:sequential_temp}
\begin{figure}[!htb]
\centering
\includegraphics[width=1\linewidth]{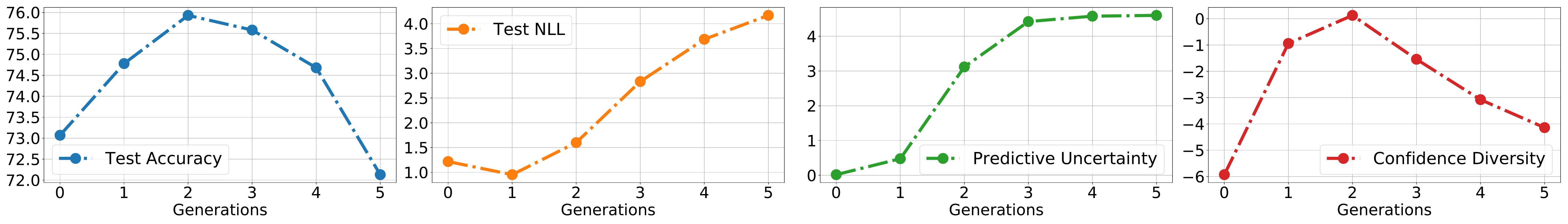}
\includegraphics[width=1\linewidth]{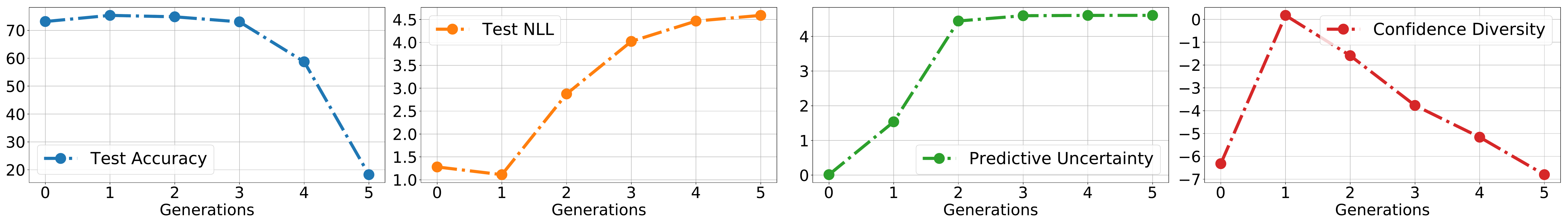}
\caption{Results for sequential self-distillation over 5 generations are shown above for different temperatures. \textit{Top:} temperature $T = 2.0$; \textit{Bottom:} temperature $T = 3.0$. The same temperatures are used throughout the entire sequential distillation process. Model obtained at the $(i-1)$-th generation is used as the teacher model for training at the $i$-th generation. Accuracy and NLL are obtained on the test set using the student model, whereas the predictive uncertainty and confidence diversity are evaluated on the training set with teacher predictions.}
\label{fig:BAN_results_temp_2}
\end{figure}

To further verify the observation on predictive uncertainty and confidence diversity made empirically in Section 4, we conduct additional sequential self-distillation experiments with different values of temperature. Figure \ref{fig:BAN_results_temp_2} summarizes the results when temperature is 2 (\textit{top}) and 3 (\textit{bottom}) respectively. As seen clearly, test accuracy and NLL performance correlate strongly with that of confidence diversity, further demonstrating the importance of confidence diversity for greater generalizability in neural networks.

\newpage
\subsection{Additional Experiments with Different Amount of Label Smoothing $\epsilon$}
\label{appendix:epsilon_exp}
\begin{figure}[!htb]
\centering
\begin{subfigure}{.32\textwidth}
\centering
\includegraphics[width=1.05\linewidth]{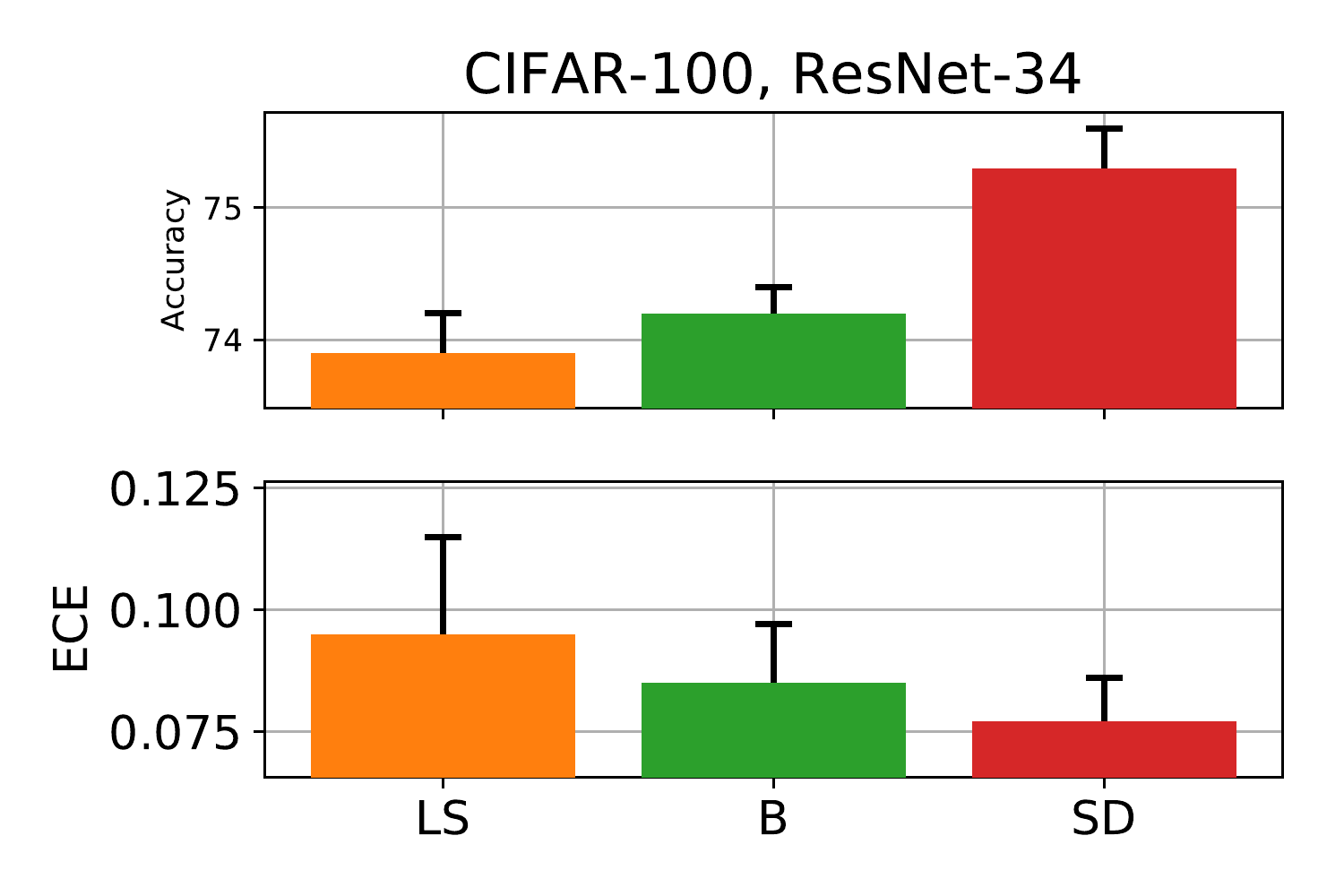}
\end{subfigure}
\begin{subfigure}{.32\textwidth}
\centering
\includegraphics[width=1.05\linewidth]{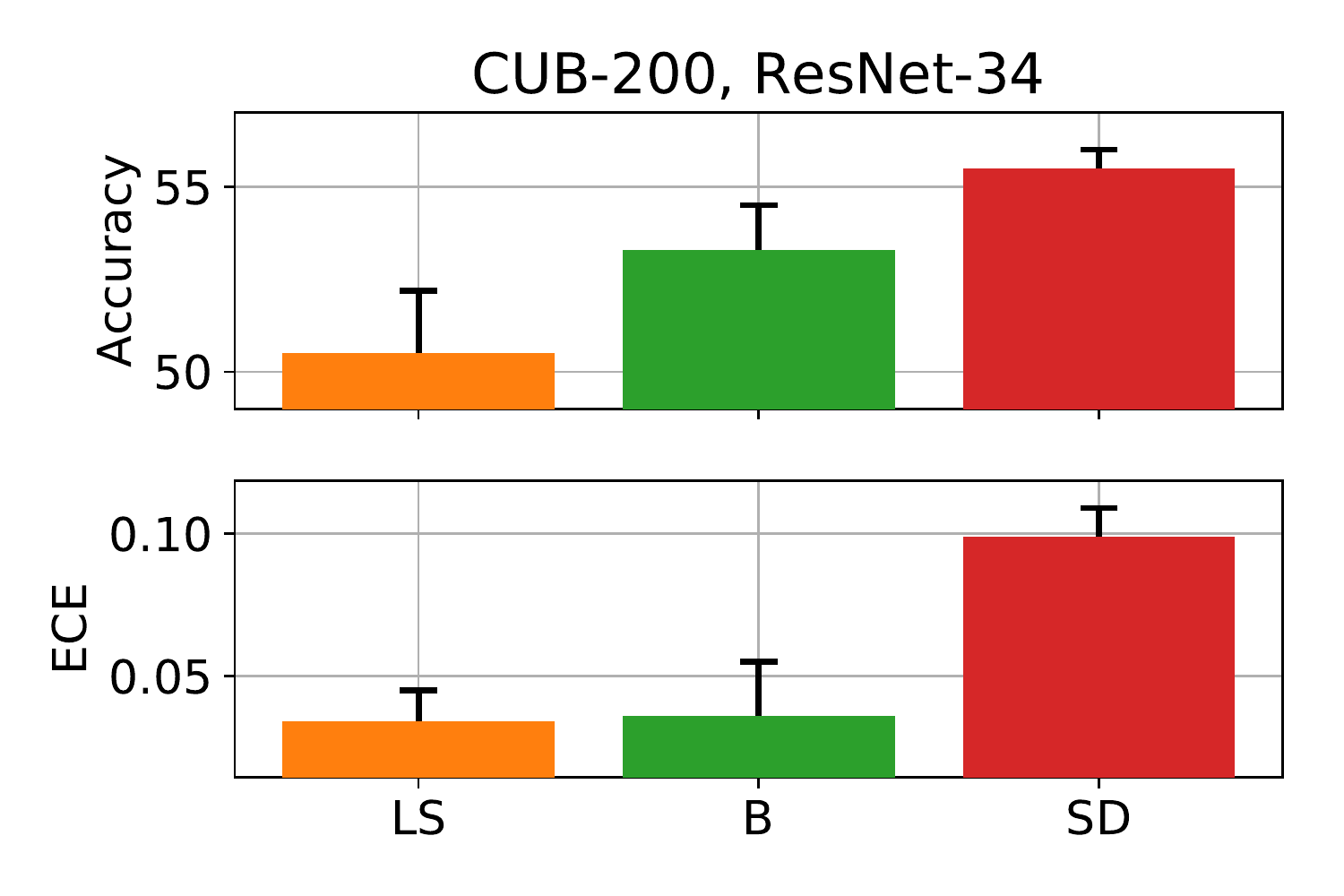}
\end{subfigure}
\begin{subfigure}{.32\textwidth}
\centering
\includegraphics[width=1.05\linewidth]{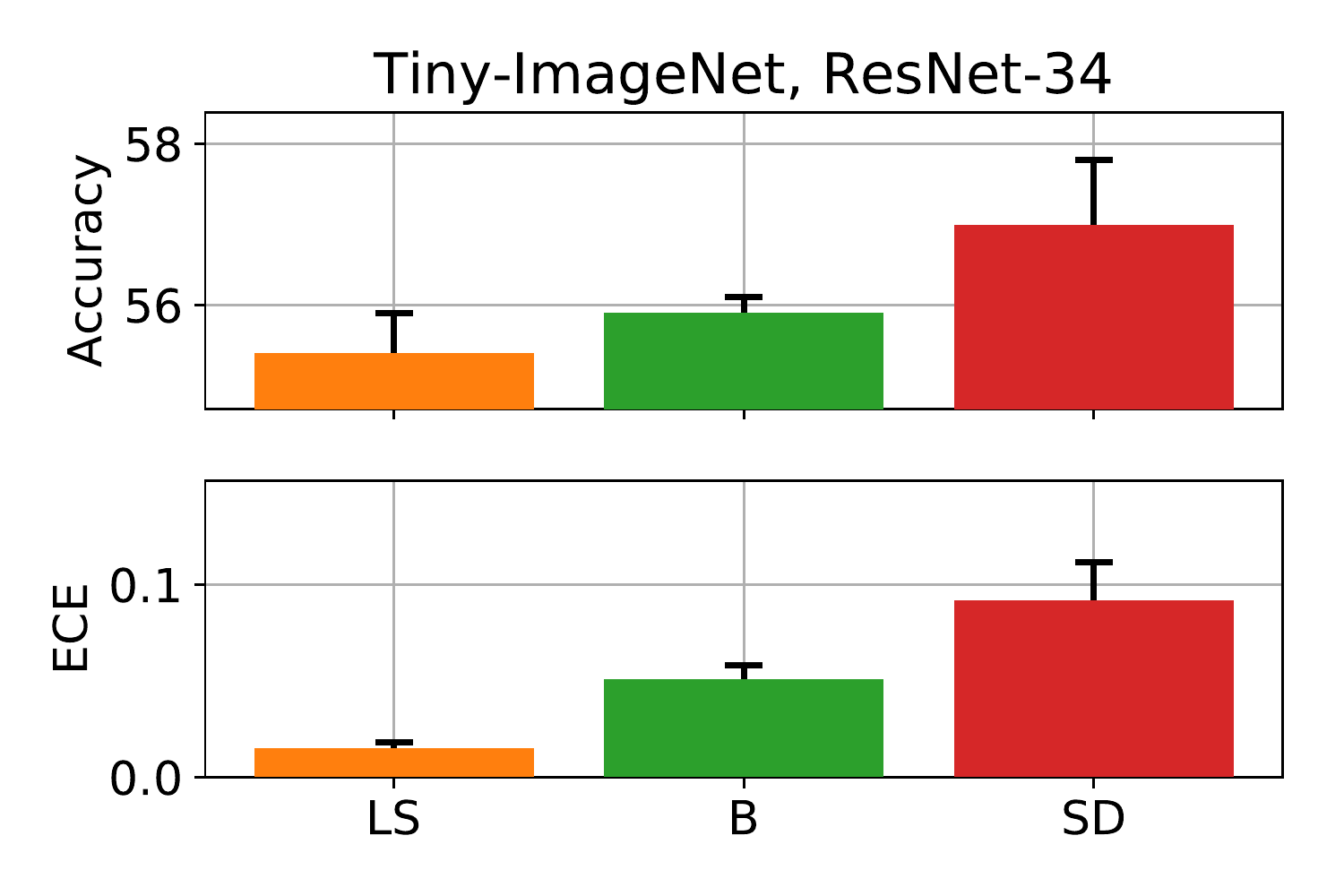}
\end{subfigure}
\begin{subfigure}{.32\textwidth}
\centering
\includegraphics[width=1.05\linewidth]{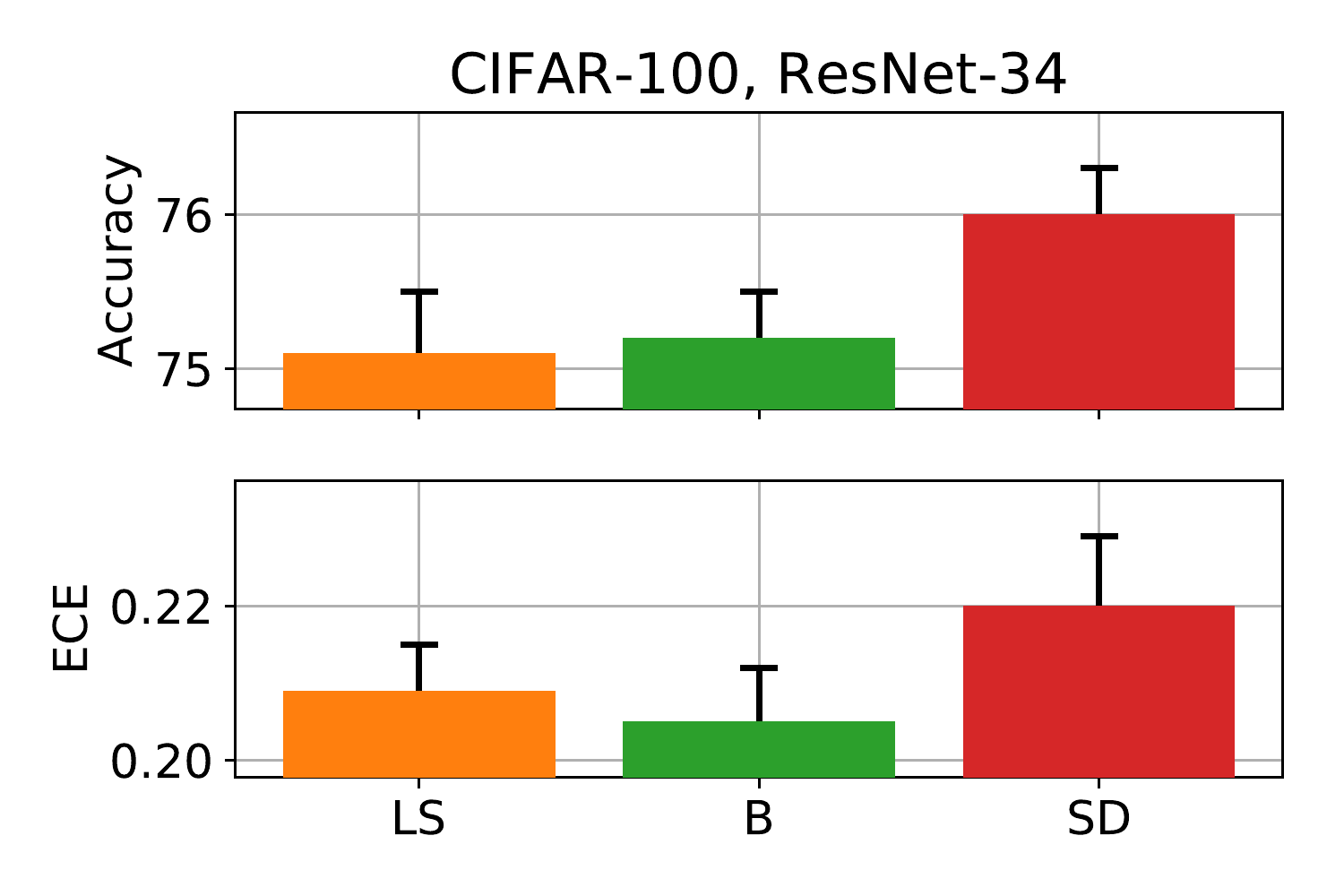}
\end{subfigure}
\begin{subfigure}{.32\textwidth}
\centering
\includegraphics[width=1.05\linewidth]{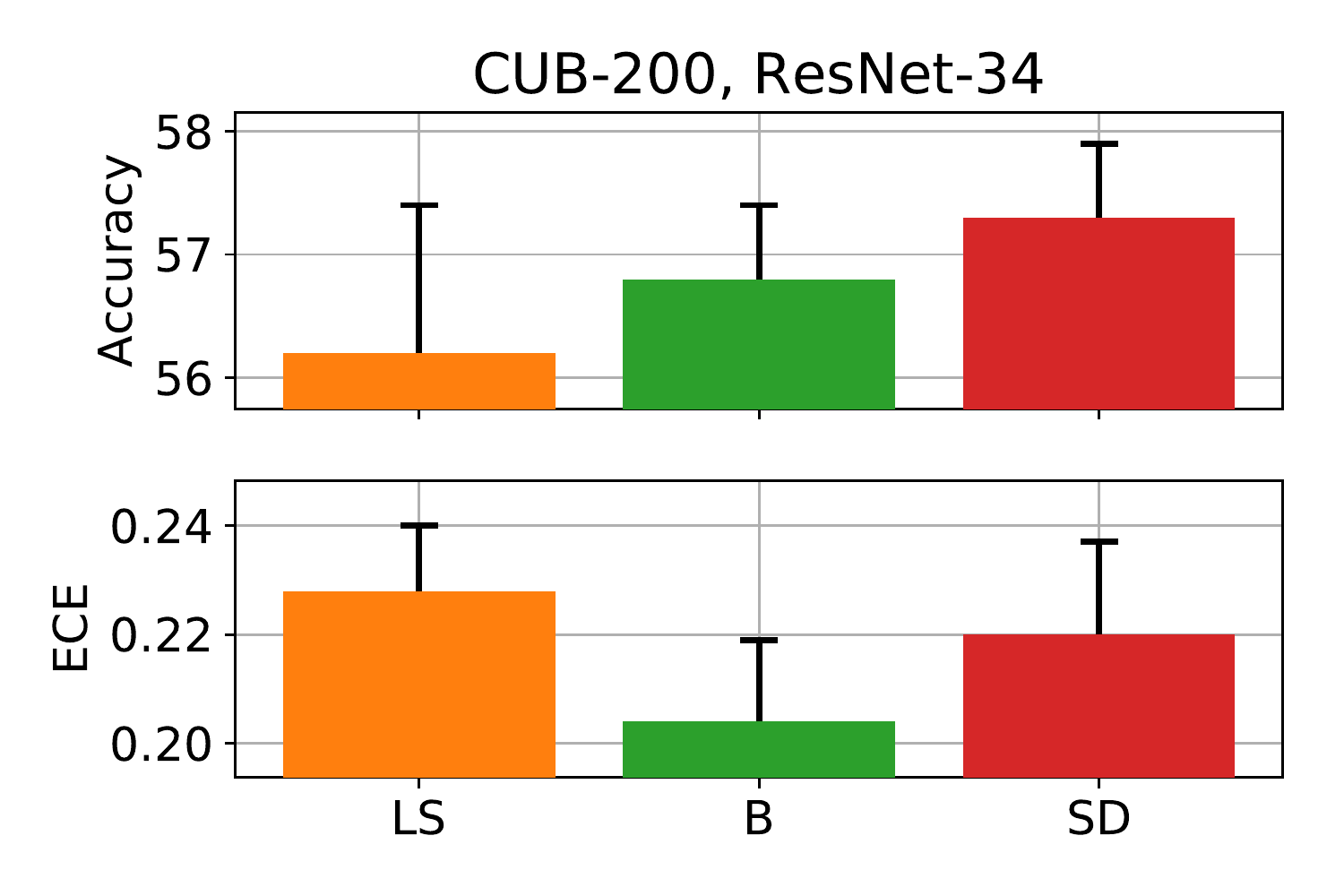}
\end{subfigure}
\begin{subfigure}{.32\textwidth}
\centering
\includegraphics[width=1.05\linewidth]{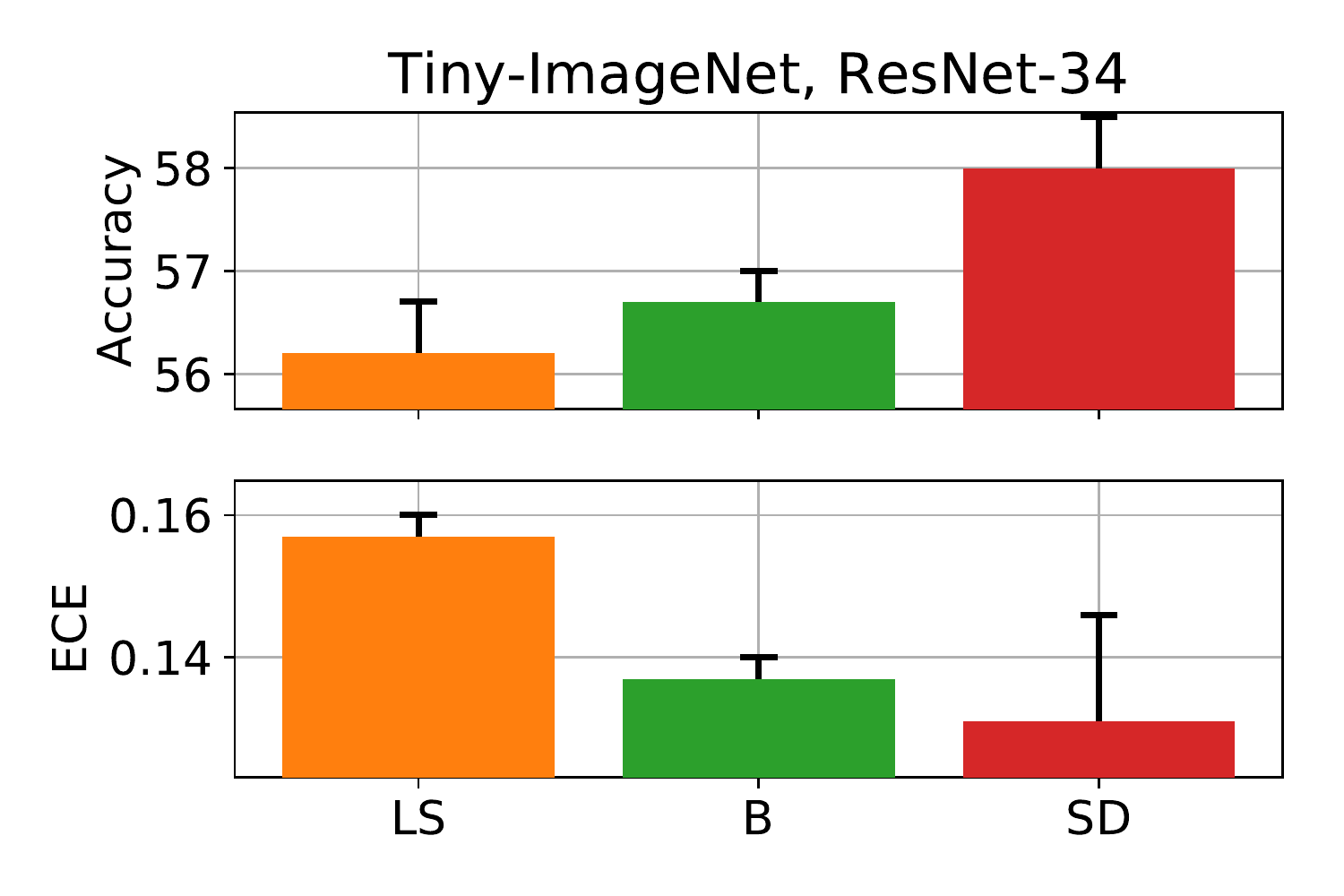}
\end{subfigure}
\caption{Experimental Results performed on CIFAR-100, CUB-200 and the Tiny-Imagenet dataset with different amount of label smoothing. \textit{Top:} $\epsilon = 0.1$, \textit{Bottom:} $\epsilon = 0.3$. "CE", "LS", "B" and "SD" refers to "Cross Entropy", "Label Smoothing", "Beta Smoothing" and "Self-Distillation" respectively. The top rows of each experiment show bar charts of accuracy on test set for each experiment conducted, while the bottom rows are bar charts of expected calibration error.}
\label{fig:results_eps01}
\end{figure}

In order to verify that the conclusions drawn from our empirical experiments hold more generally, we conduct additional experiments varying the amount of label smoothing $\epsilon$. Additional smoothing parameters of $\epsilon = 0.1$ and $\epsilon = 0.3$ are used. As a fair comparison, given the label smoothing parameter $\epsilon$, hyper-parameters for Beta smoothing and self-distillation are adjusted so that the amount of label smoothing for samples on average is the same as that of label smoothing. Experimental results are summarized in Figure ~\ref{fig:results_eps01}. Observe that the general trend in terms of both the accuracy and calibration holds across different values of $\epsilon$. 

\newpage
\subsection{Additional Experiments with Self-Training Using EMA-Predictions}
\label{appendix:ema}
\begin{figure}[!htb]
\centering
\begin{subfigure}{.32\textwidth}
\centering
\includegraphics[width=1.05\linewidth]{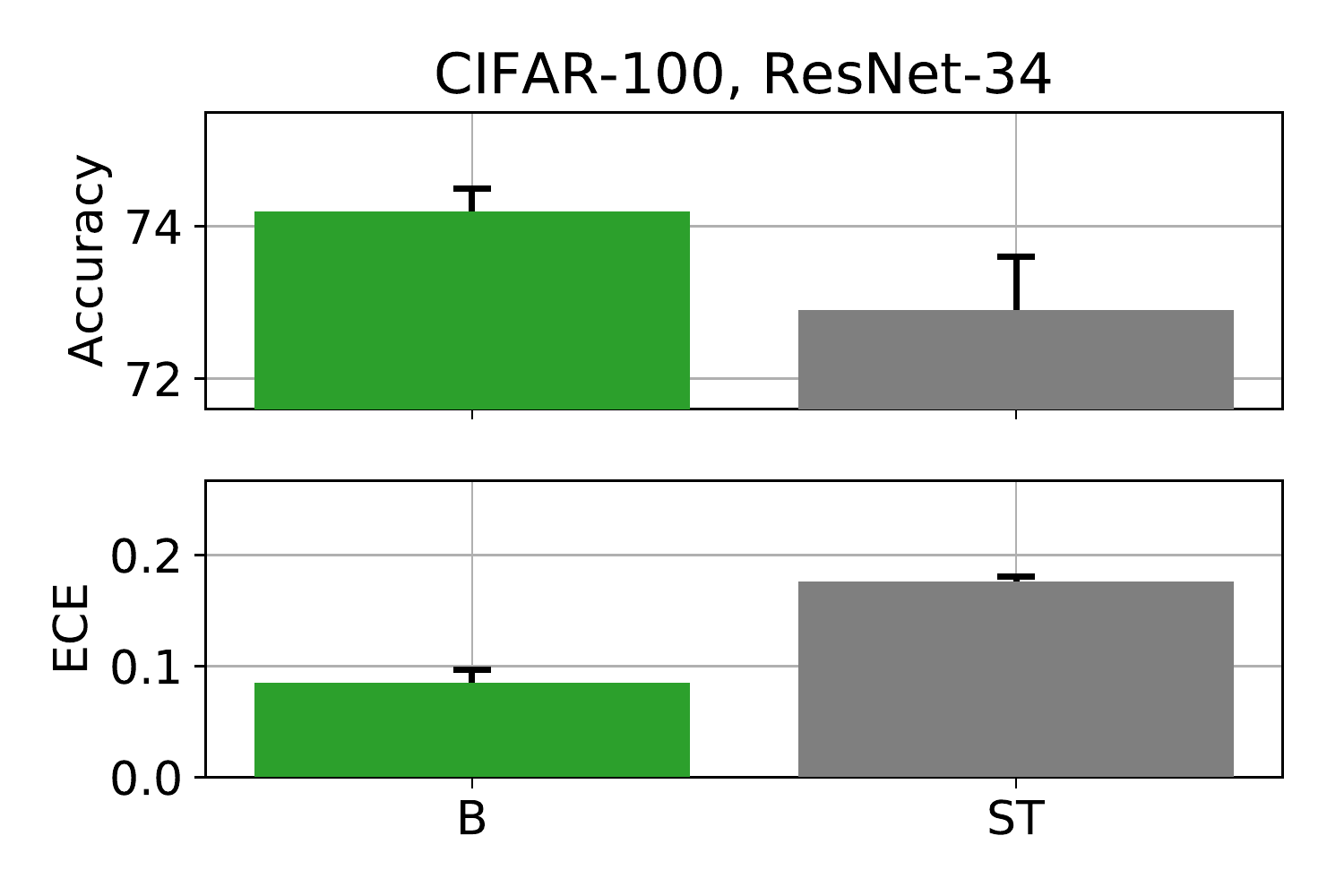}
\end{subfigure}
\begin{subfigure}{.32\textwidth}
\centering
\includegraphics[width=1.05\linewidth]{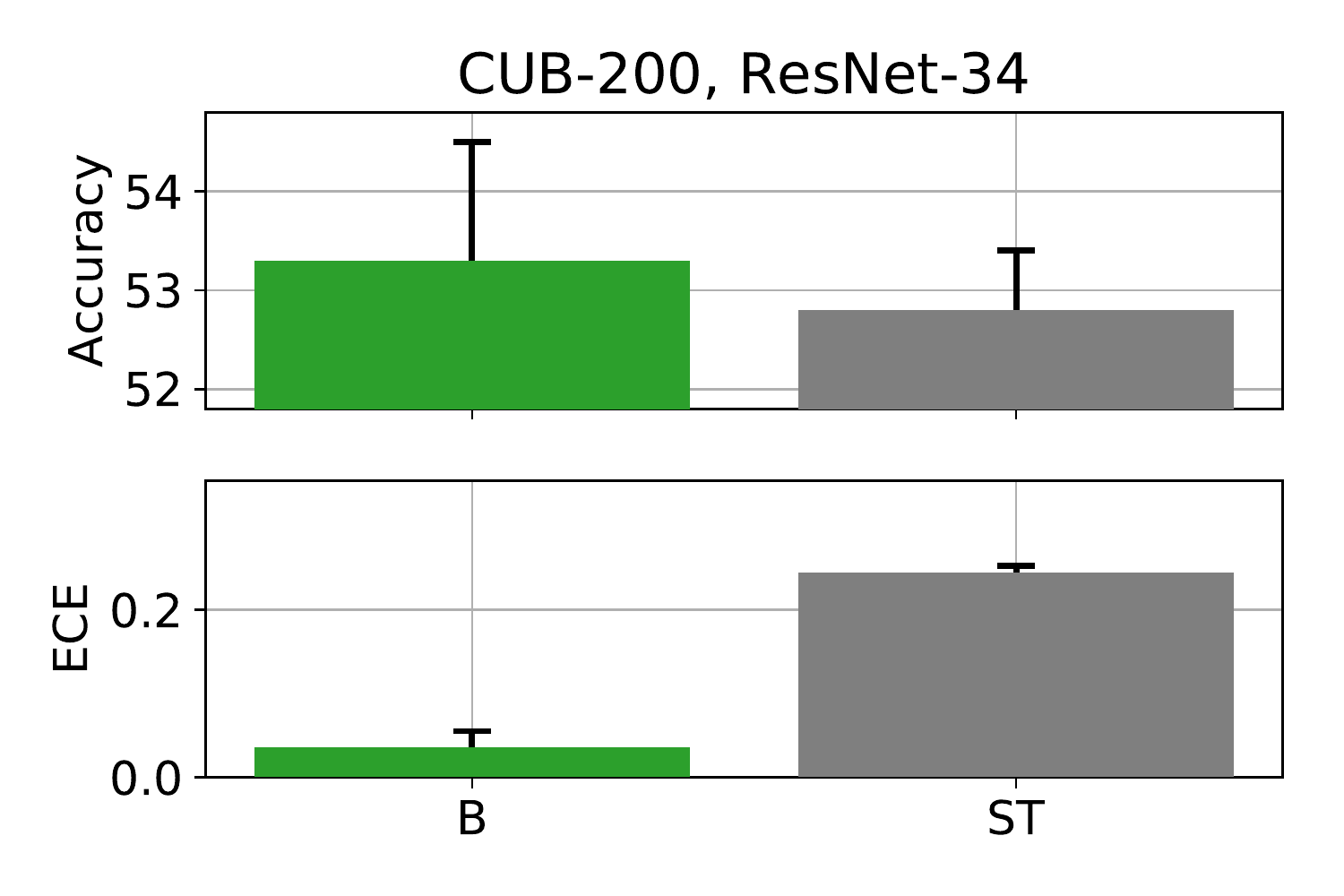}
\end{subfigure}
\begin{subfigure}{.32\textwidth}
\centering
\includegraphics[width=1.05\linewidth]{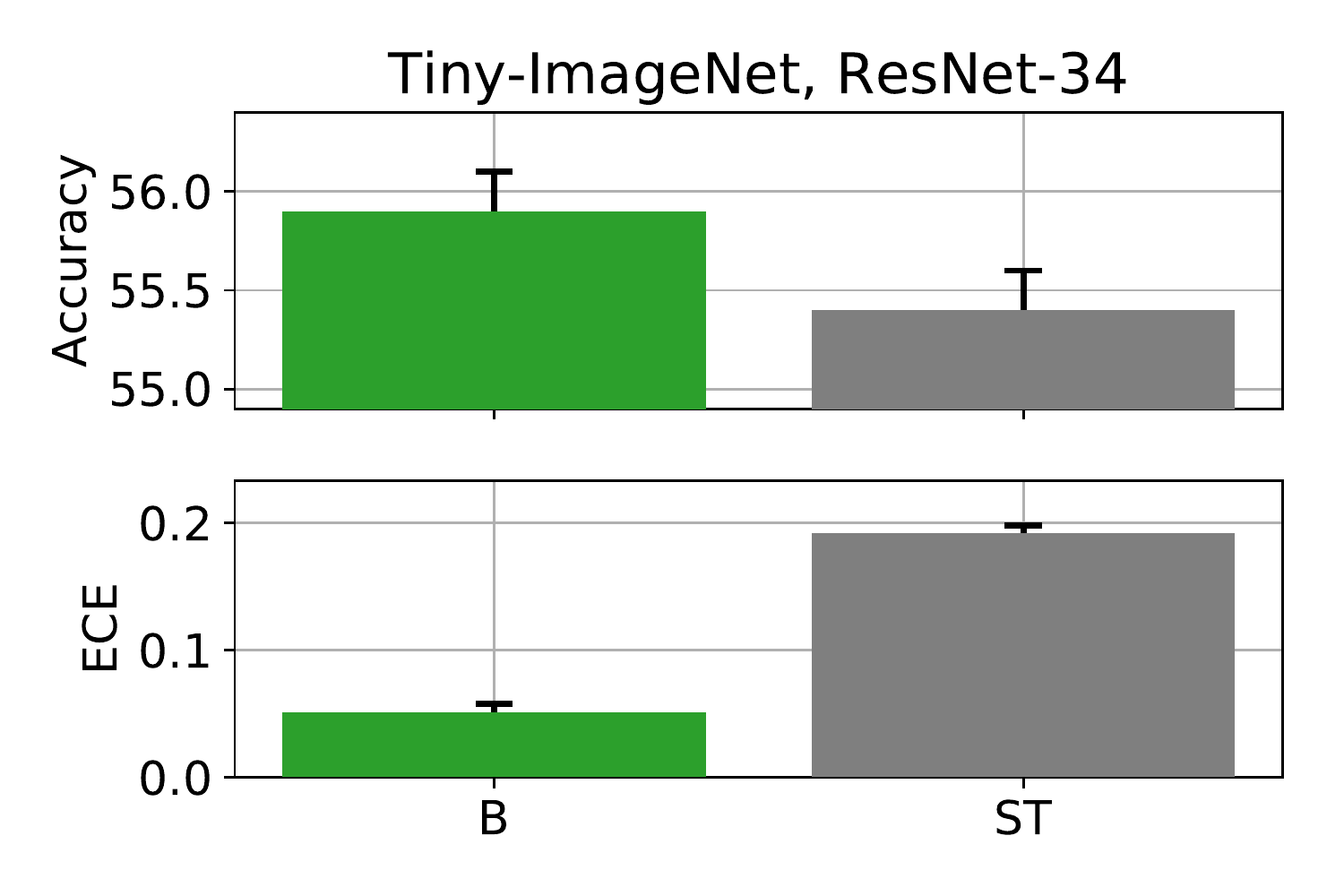}
\end{subfigure}
\caption{Additional results to compare Beta smoothing against self-training explicitly with the EMA predictions. "B" and "ST" refer to "beta smoothing" and "self-training" respectively. The top rows of each experiment show bar charts of accuracy on the test set for each experiment conducted, while the bottom rows are bar charts of expected calibration error.}
\label{fig:st_results}
\end{figure}

The proposed beta smoothing involves the use of EMA predictions to rank the confidence of samples within each minibatch during training in order to achieve instance-specific regularization. To further demonstrate that the gain in accuracy and calibration obtained through beta smoothing mainly comes from instance-specific regularization, we compare Beta smoothing against explicit self-training using the EMA predictions in which the EMA predictions are directly used as soft labels to compute cross-entropy loss. We follow the training procedure as described in ~\cite{tarvainen2017mean} for self-training with EMA predictions. Results using ResNet for all the datasets considered in this paper are summarized in Figure~\ref{fig:st_results}. Beta smoothing outperforms self-training using EMA predictions on all of the experiments conducted in terms of both accuracy and calibration. As such, while EMA predictions can be used as a reliable proxy to rank the relative confidence of samples, the predictions themselves are sub-optimal when used as teachers directly.

\newpage
\subsection{Additional Experiments with CIFAR-10 When Varying Trainset Size}
\label{appendix:train_size}
\begin{figure}[!htb]
\centering
\begin{subfigure}{.45\textwidth}
\centering
\includegraphics[width=1.0\linewidth]{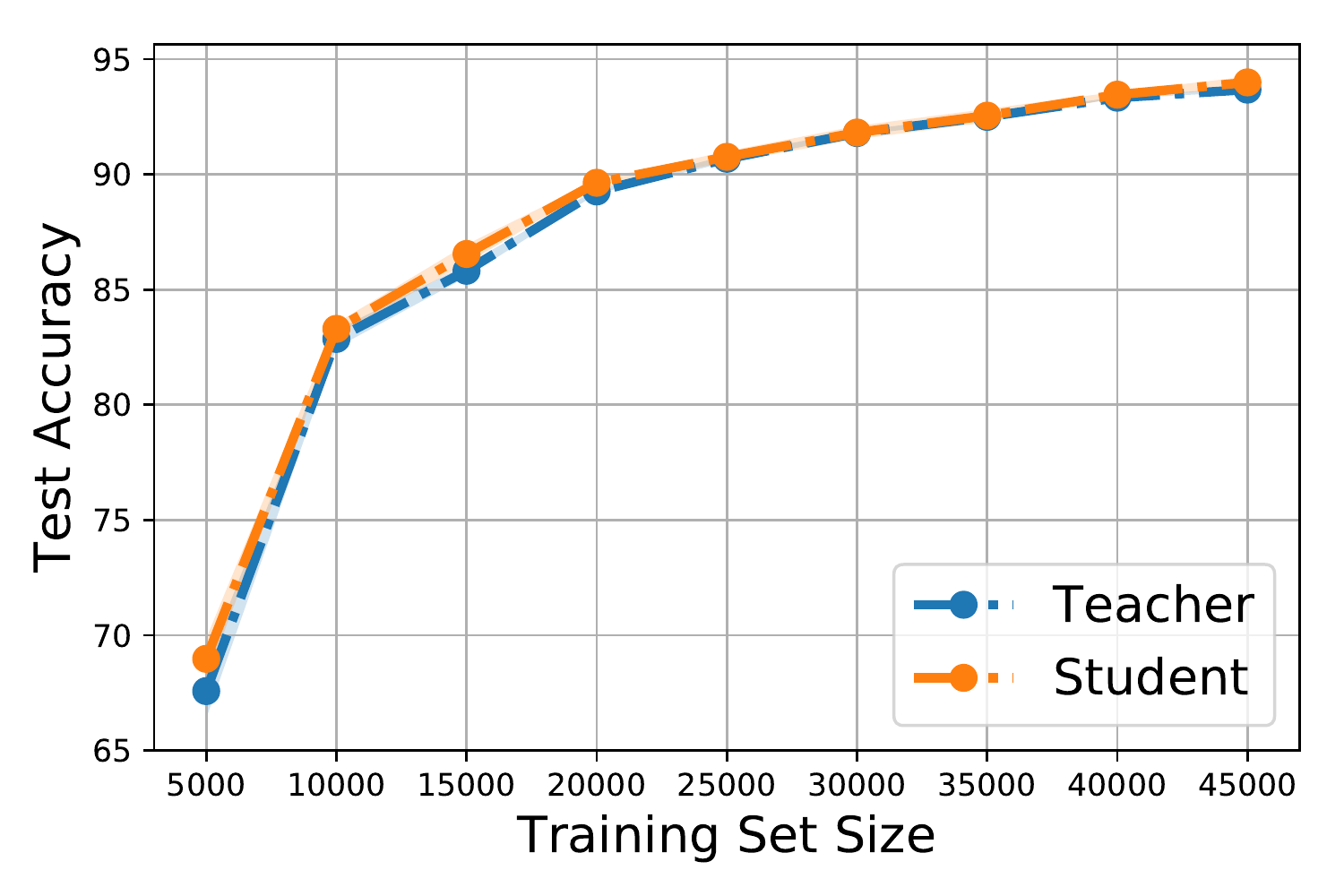}
\end{subfigure}
\begin{subfigure}{.45\textwidth}
\centering
\includegraphics[width=1.0\linewidth]{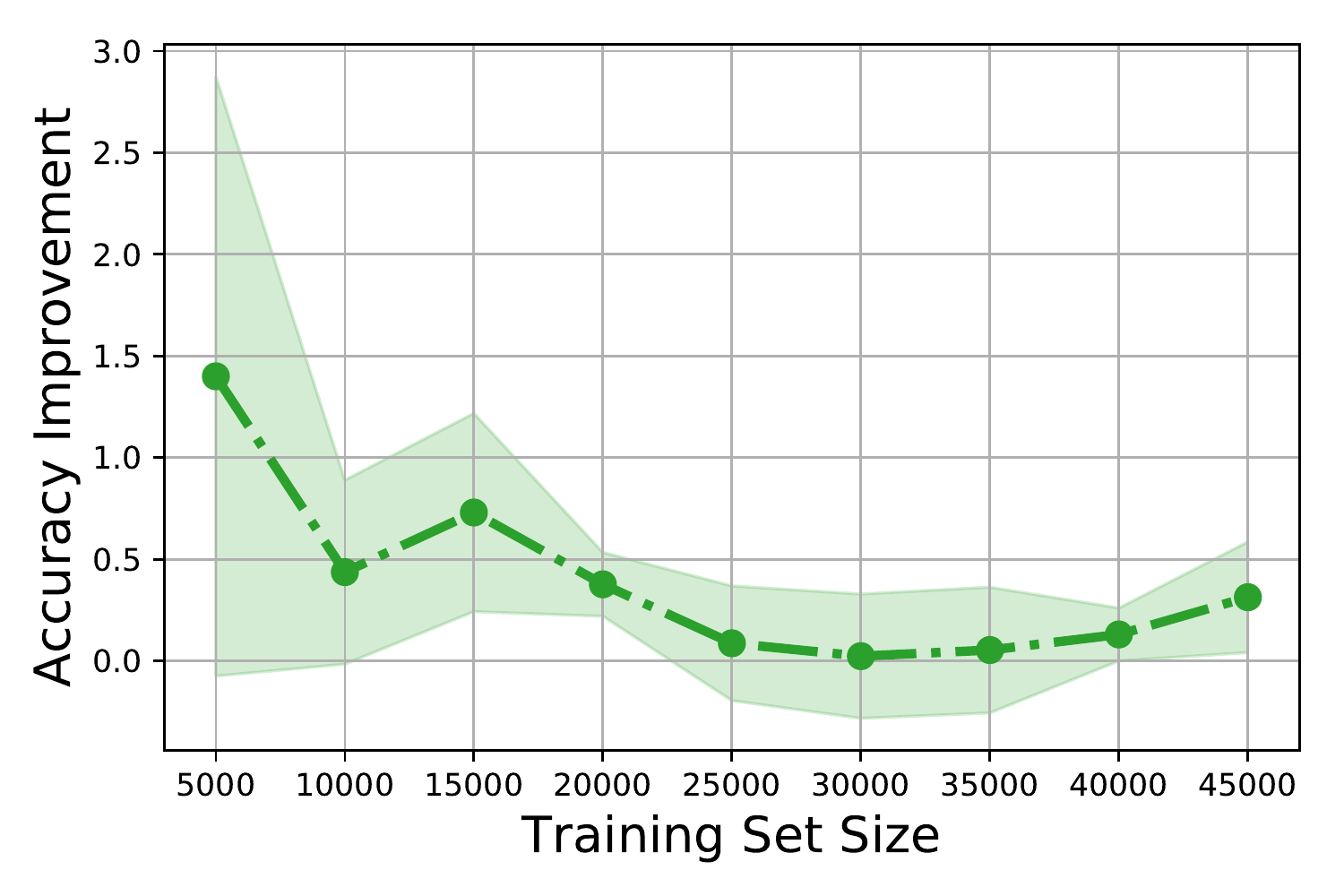}
\end{subfigure}
\caption{\textit{Left:} Test accuracies of ResNet-34 models on the CIFAR-10 dataset for the teacher and student models when the training set size is varied. \textit{Right:} The relative improvements in accuracy when the training set size is varied.}
\label{fig:cifar_10_results}
\end{figure}

Recent results show relatively small gain when performing knowledge distillation on the CIFAR-10 dataset~\cite{cho2019efficacy,furlanello2018born}. Our perspective of distillation as regularization provides a plausible explanation for this observation. Like all other forms of regularization, its effect diminishes with increasing the size of training data. We experimentally verify the claim by training ResNet-34 models with a varying number of training samples. The experiment are repeated 3 times. Fig.~\ref{fig:cifar_10_results} summarizes the results. As expected, increasing sample size leads to an increase in test accuracy for both of the models. Nevertheless, the relative improvement in the accuracy of the student model compared to the teacher decreases as the size of the training set increases, indicating that distillation is a form of regularization.

\subsection{Additional Experiments with CIFAR-100 When Varying Weight Decay}
\label{appendix:weight_decay}
\begin{figure}[!htb]
\centering
\begin{subfigure}{.45\textwidth}
\centering
\includegraphics[width=1.0\linewidth]{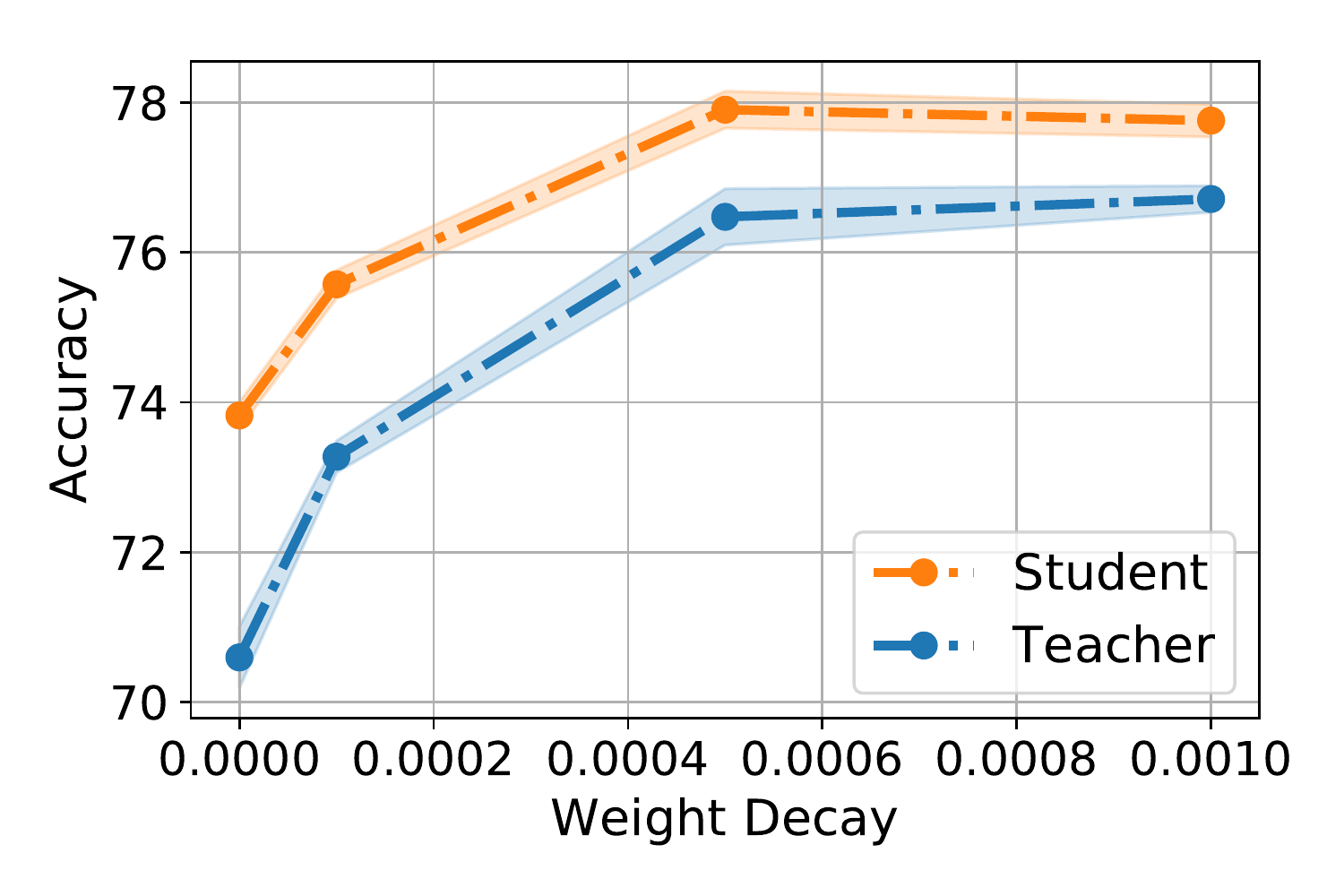}
\end{subfigure}
\begin{subfigure}{.45\textwidth}
\centering
\includegraphics[width=1.0\linewidth]{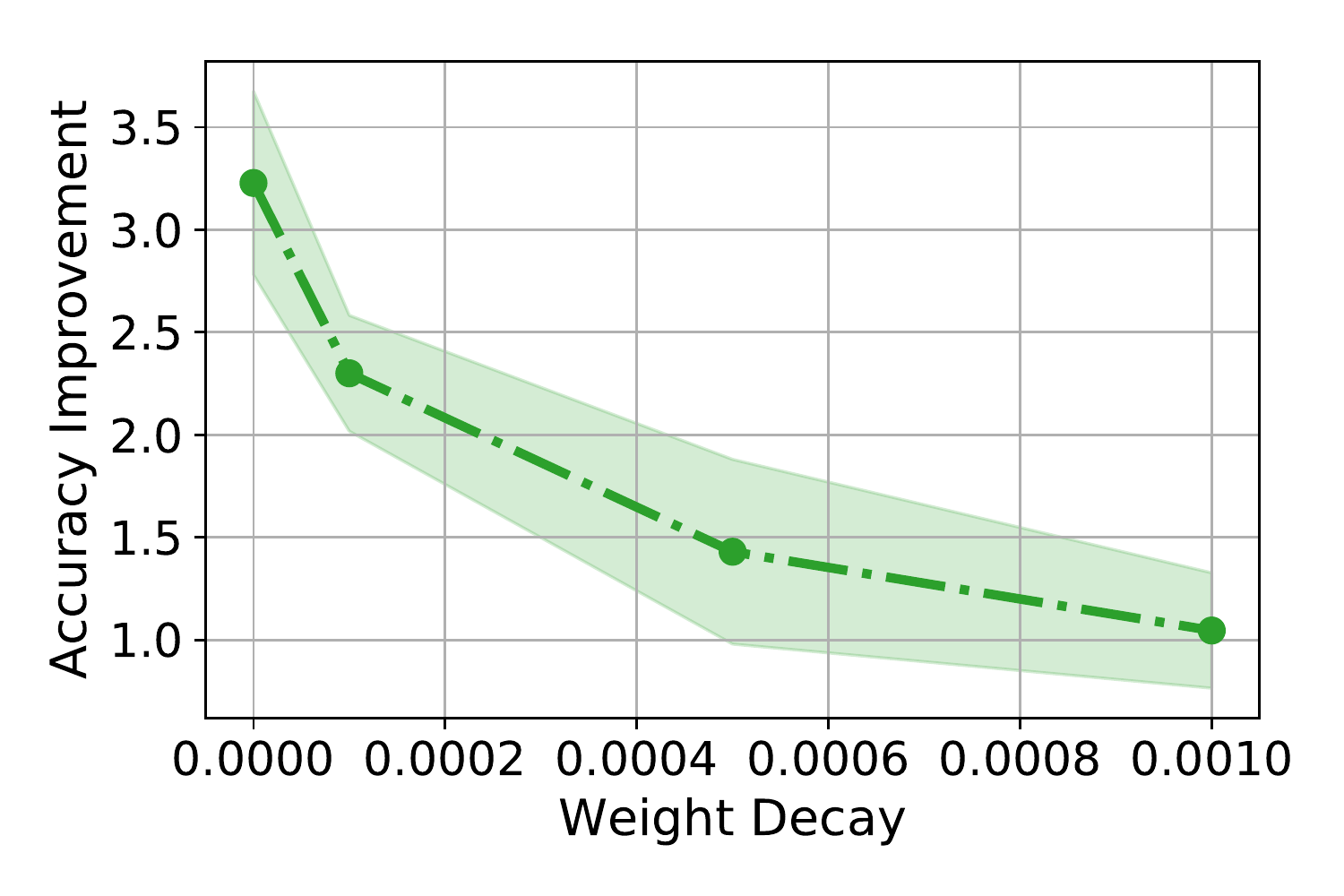}
\end{subfigure}
\caption{\textit{Left:} Test accuracies of ResNet-34 models on the CIFAR-100 dataset for the teacher and student models when the weight decay hyper-parameter is varied. \textit{Right:} The relative improvements in accuracy when the weight decay hyper-parameter is varied.}
\label{fig:wd_results}
\end{figure}

To further demonstrate that distillation is a regularization process, we also conduct an additional experiment on the CIFAR-100 dataset using ResNet-34, varying only the weight decay hyper-parameter. Intuitively, larger weight decay regularization makes NNs less prone to overfitting, which should, in turn, reduced the additional benefits obtainable from self-distillation, if it is indeed a form of regularization. 
To keep the quality of priors identical across all student models, we use the same teacher model obtained from using a weight decay of $10^{-4}$ for all distillation. Our results are summarized in Fig.~\ref{fig:wd_results}. It is evident that increasing the weight decay hyper-parameter leads to much smaller improvement in test accuracy. Interestingly, we see a noticeable gain in accuracy for baselines models trained with cross-entropy when adjusting the weight decay term, contradicting some of the recent findings that weight decay is ineffective for neural networks.

\newpage
\subsection{Additional Experiments on Beta Smoothing}
\label{appendix:beta}

\begin{figure}[!htb]
\centering
\begin{subfigure}{.32\textwidth}
\centering
\includegraphics[width=1.05\linewidth]{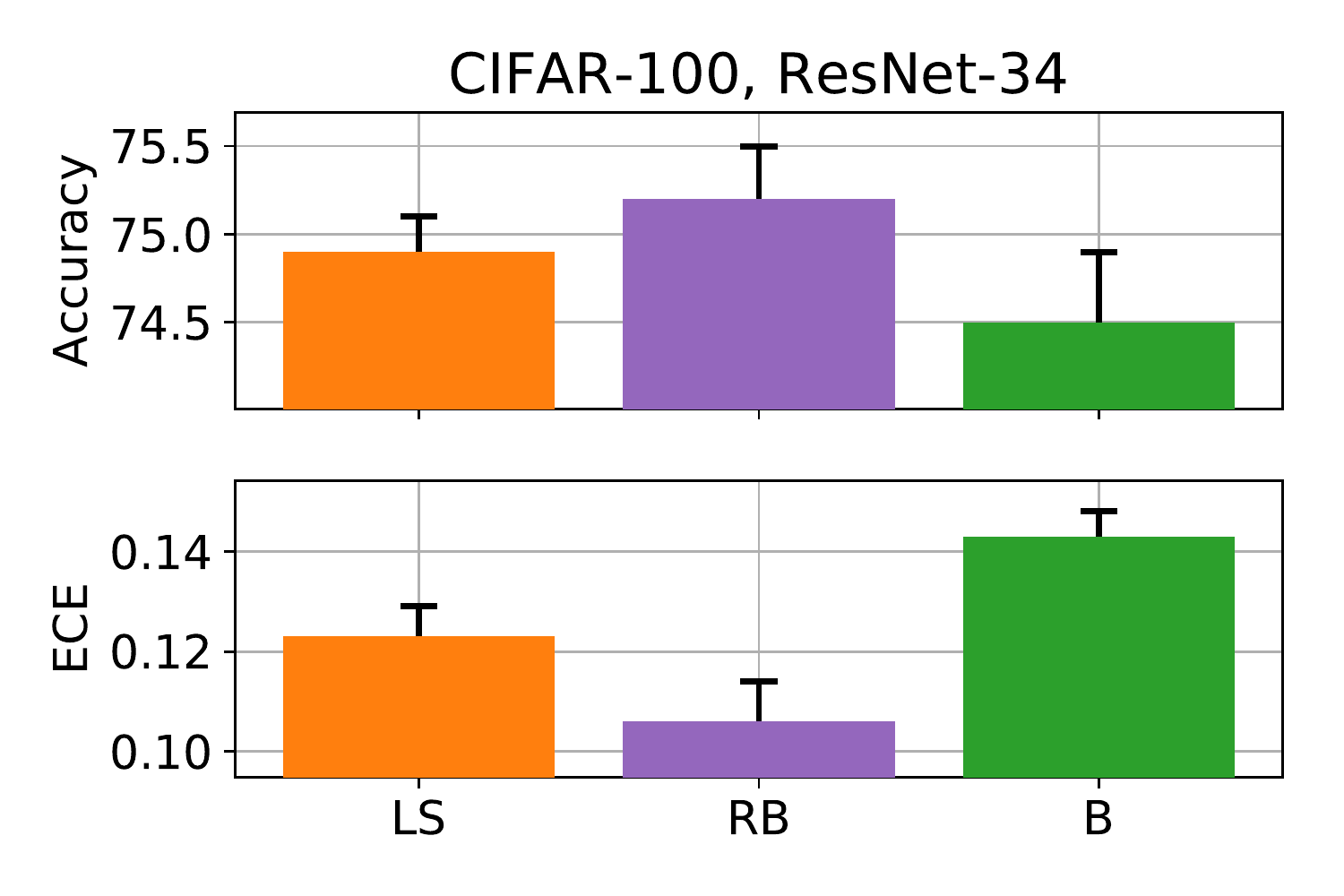}
\end{subfigure}
\begin{subfigure}{.32\textwidth}
\centering
\includegraphics[width=1.05\linewidth]{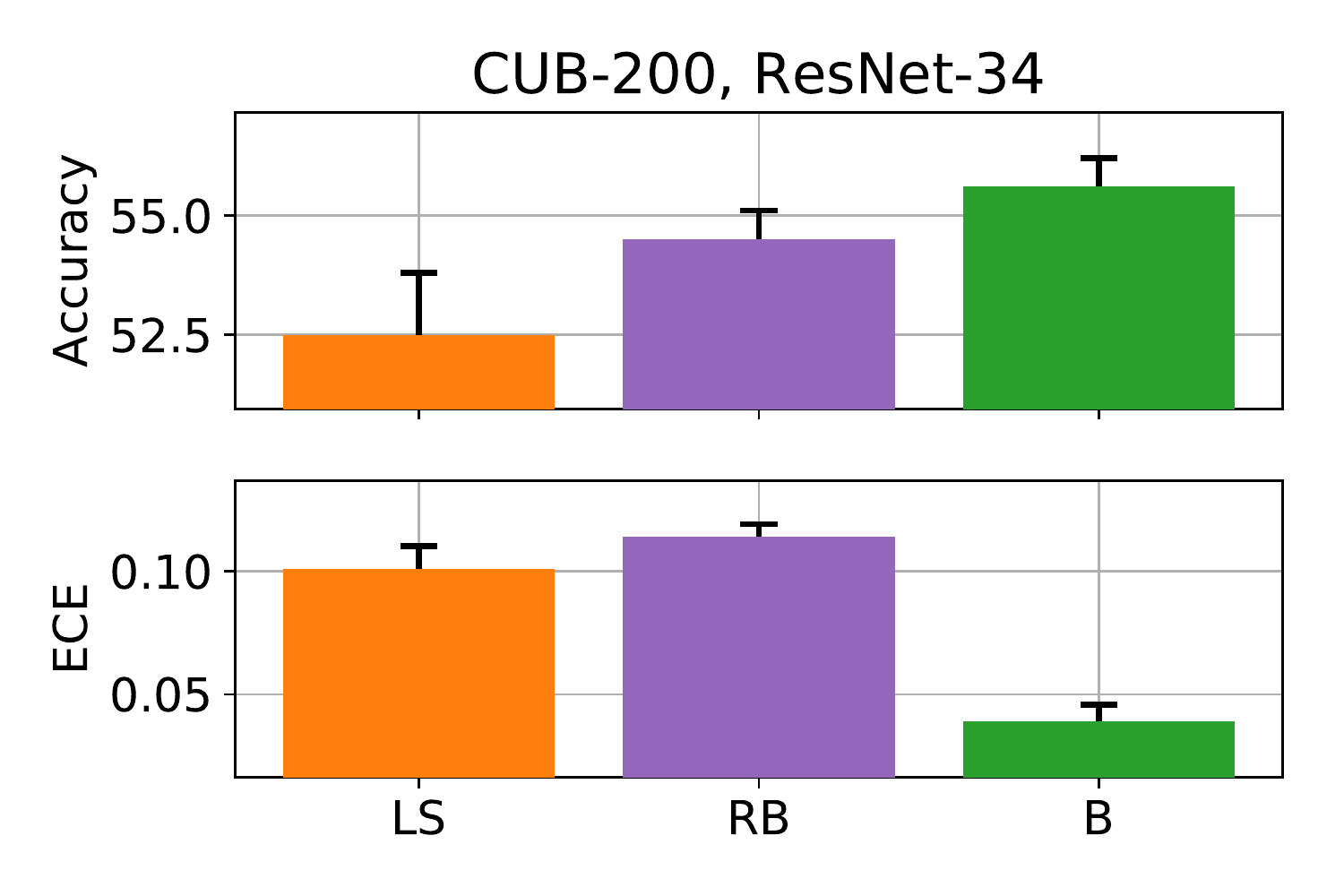}
\end{subfigure}
\begin{subfigure}{.32\textwidth}
\centering
\includegraphics[width=1.05\linewidth]{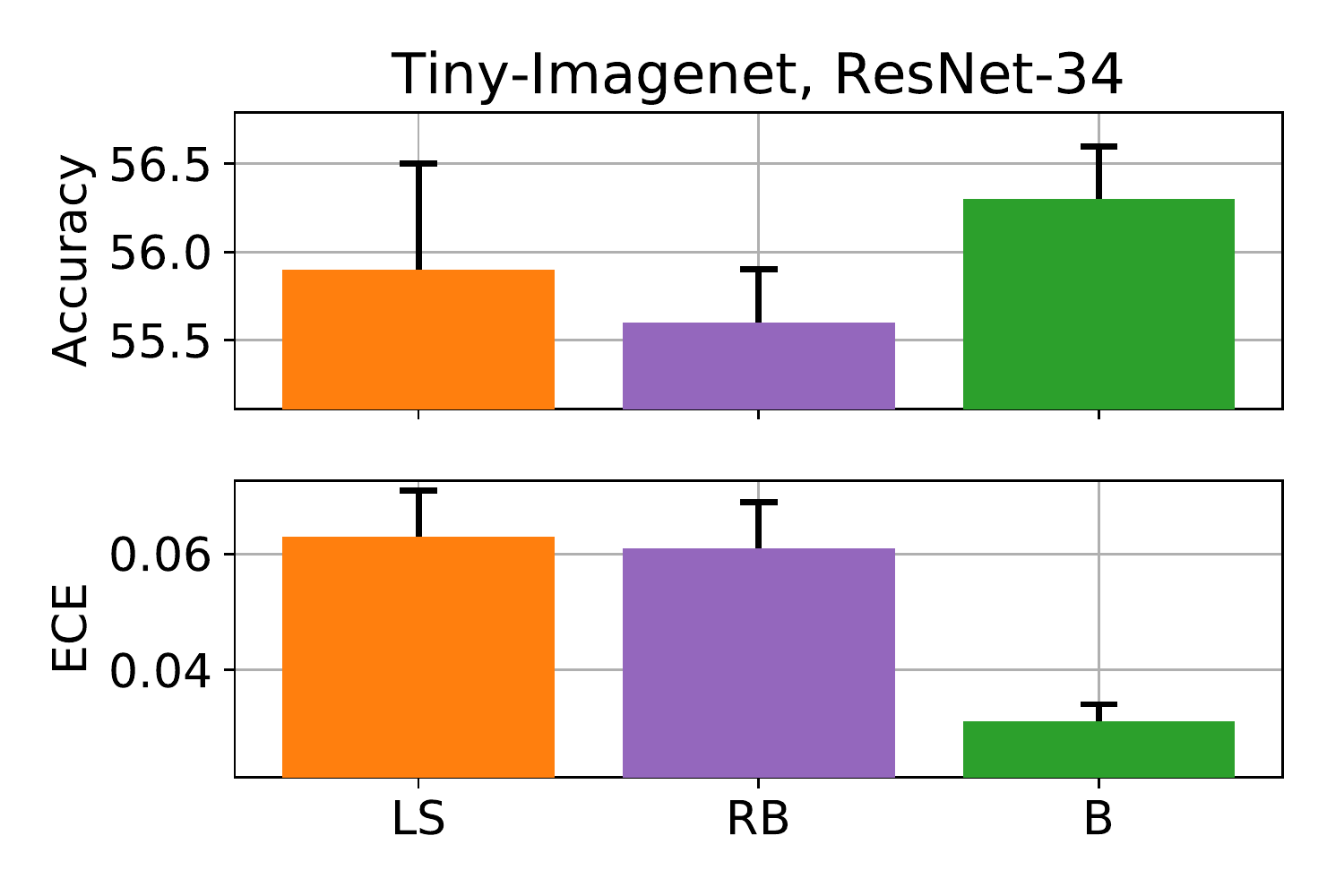}
\end{subfigure}
\begin{subfigure}{.32\textwidth}
\centering
\includegraphics[width=1.05\linewidth]{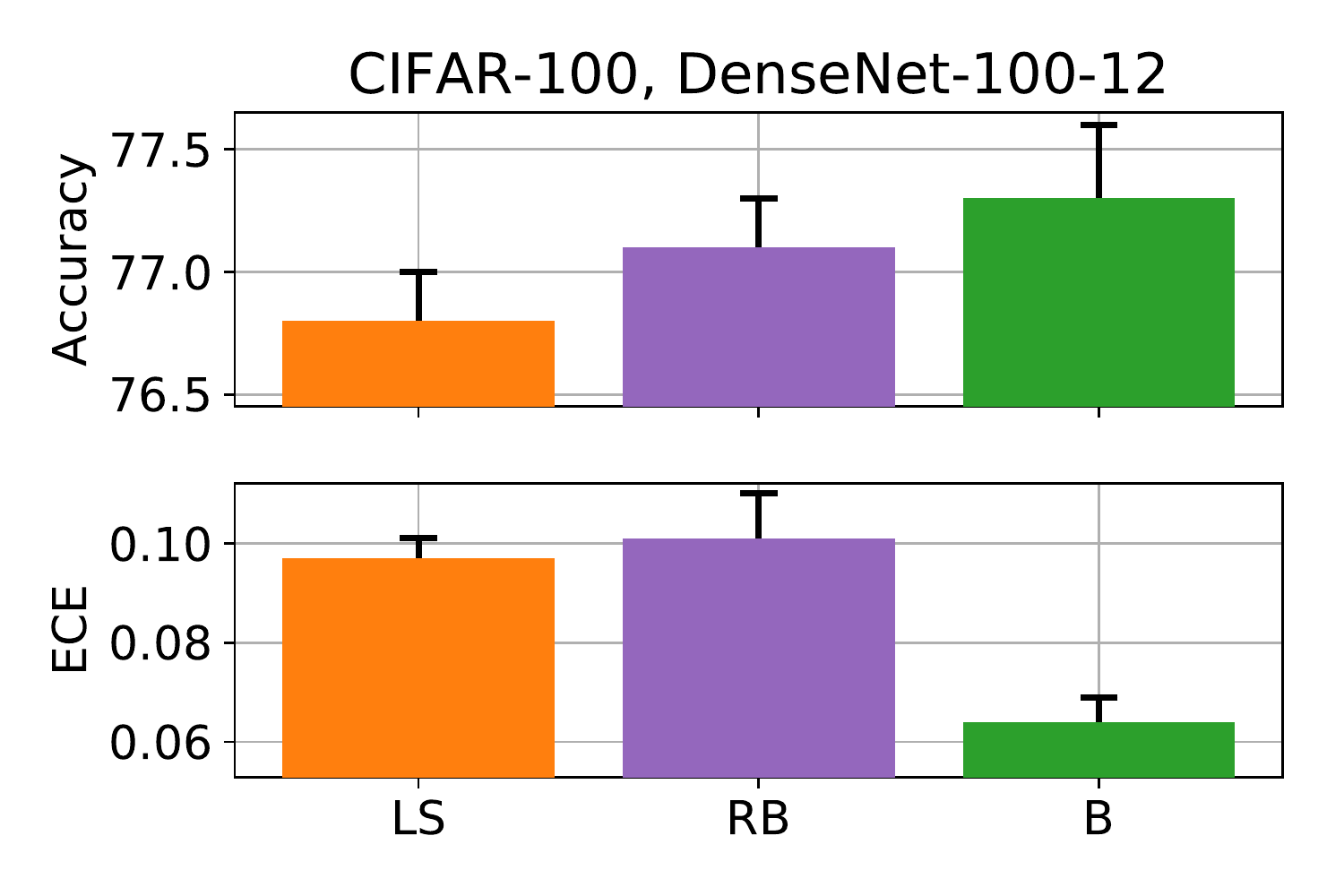}
\end{subfigure}
\begin{subfigure}{.32\textwidth}
\centering
\includegraphics[width=1.05\linewidth]{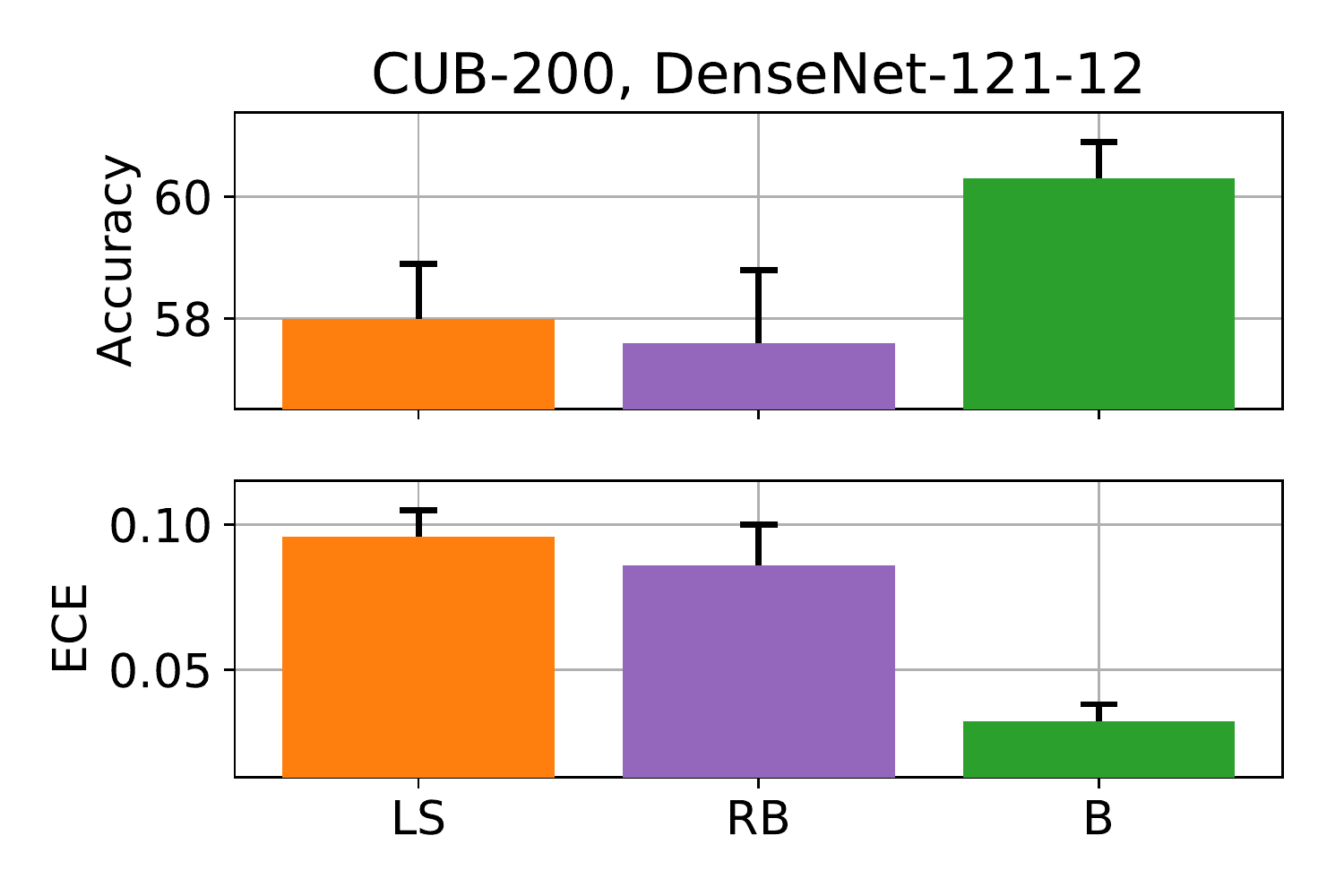}
\end{subfigure}
\begin{subfigure}{.32\textwidth}
\centering
\includegraphics[width=1.05\linewidth]{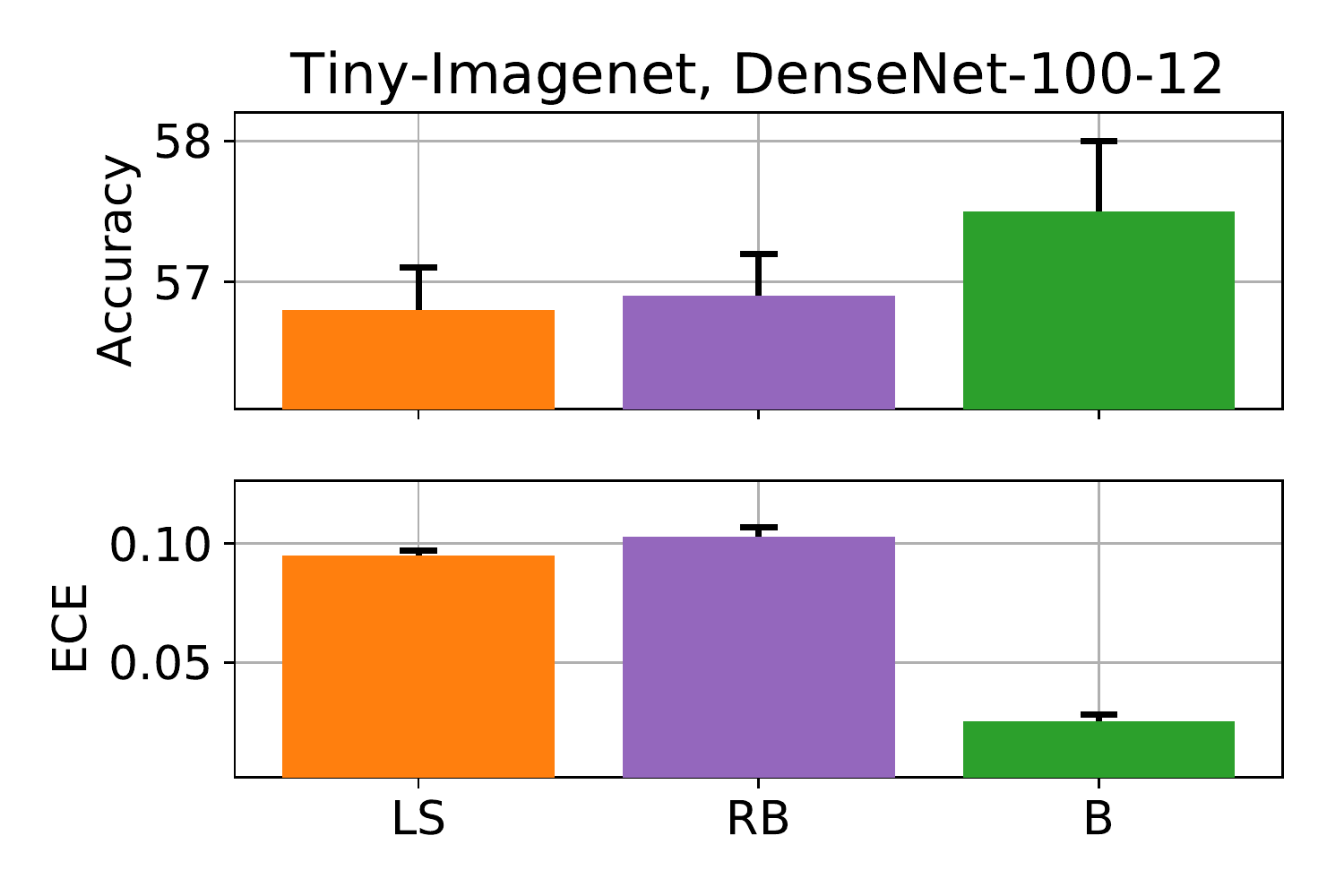}
\end{subfigure}
\caption{Ablation study on Beta smoothing. "LS", "RB" and "B" refers to "Label Smoothing", "Random Beta Smoothing" and "Beta Smoothing" respectively. The top rows of each experiment show bar charts of accuracy on the test set for each experiment conducted, while the bottom rows are bar charts of expected calibration error.}
\label{fig:beta_results}
\end{figure}
We conduct an ablation study on the proposed Beta smoothing regularization in order to demonstrate the importance of relative ranking. To do so, we run experiments with the identical setup as described in Section~\ref{section:experiment} for Beta smoothing with completely randomly assigned soft label noise from Beta distribution instead. We term this the ``random Beta smoothing''. Results are shown in Fig.~\ref{fig:beta_results}. For convenience, we also include results obtained with regular label smoothing as a benchmark comparison. As seen clearly, the proposed Beta smoothing with ranking obtained from EMA predictions leads to much better results in general in terms of both accuracy and ECE, suggesting that naively encouraging confidence diversity does not lead to significant improvements, and the relative confidence among different samples is also an important aspect in order to obtain better student models. This ablation study also serves as indirect evidence for why self-distillation still outperforms Beta smoothing - with a pre-trained model, much more reliable relative confidence among training samples can be obtained. 

\newpage
\subsection{Additional Experiments on the Effect of Quality of Teachers}
\label{appendix:cross-distillation}
\begin{figure}[!htb]
\centering
\begin{subfigure}{.32\textwidth}
\centering
\includegraphics[width=1.05\linewidth]{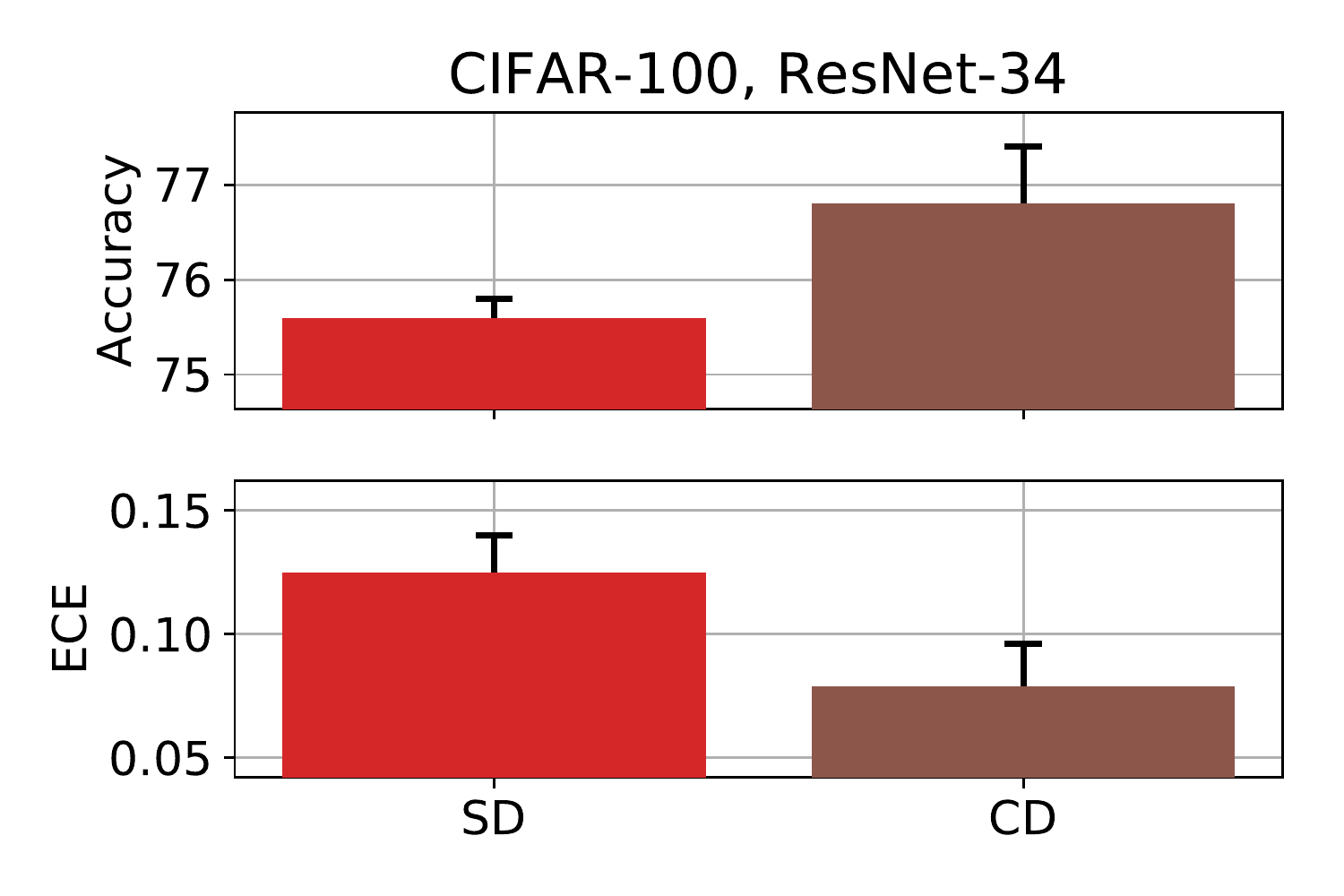}
\end{subfigure}
\begin{subfigure}{.32\textwidth}
\centering
\includegraphics[width=1.05\linewidth]{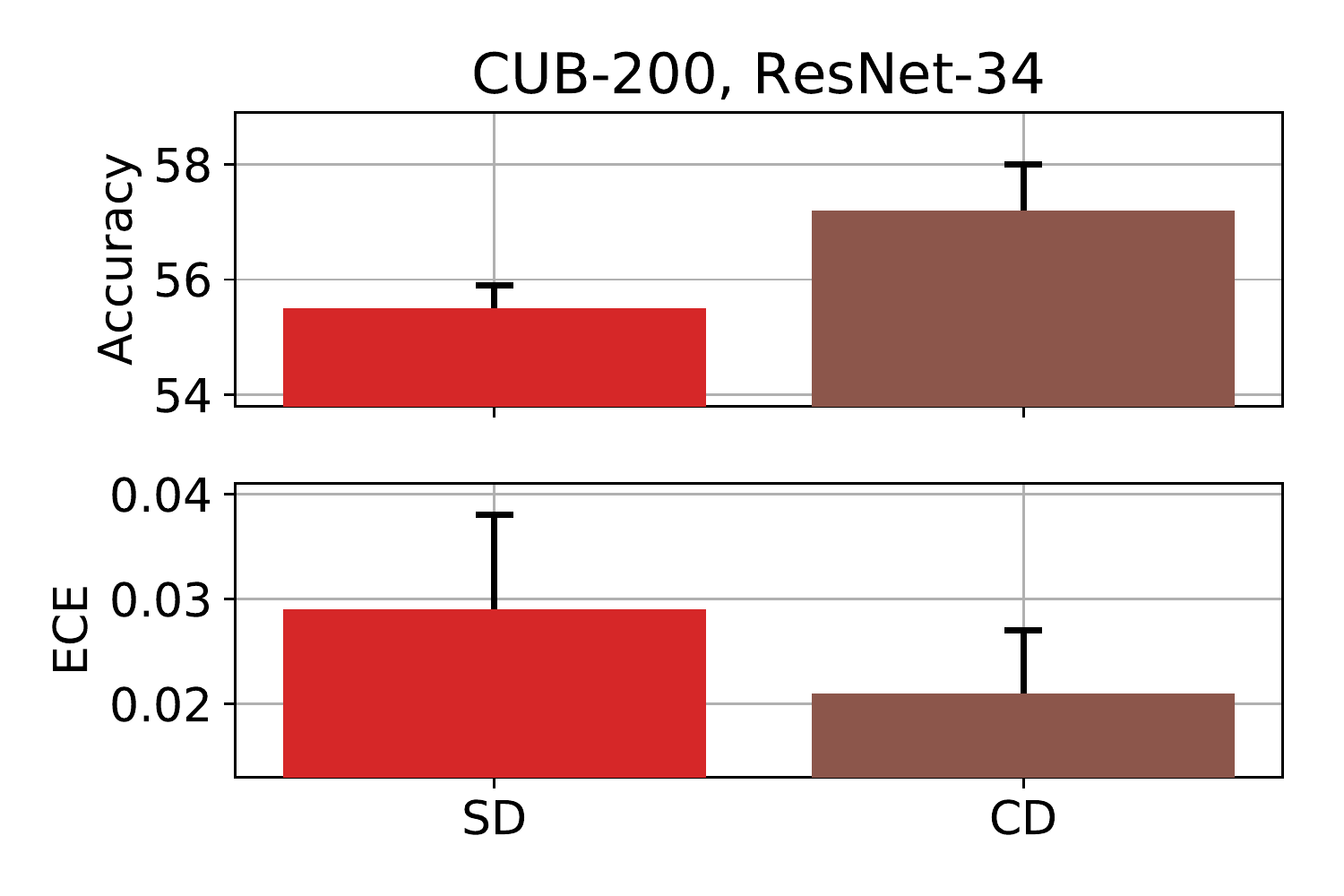}
\end{subfigure}
\begin{subfigure}{.32\textwidth}
\centering
\includegraphics[width=1.05\linewidth]{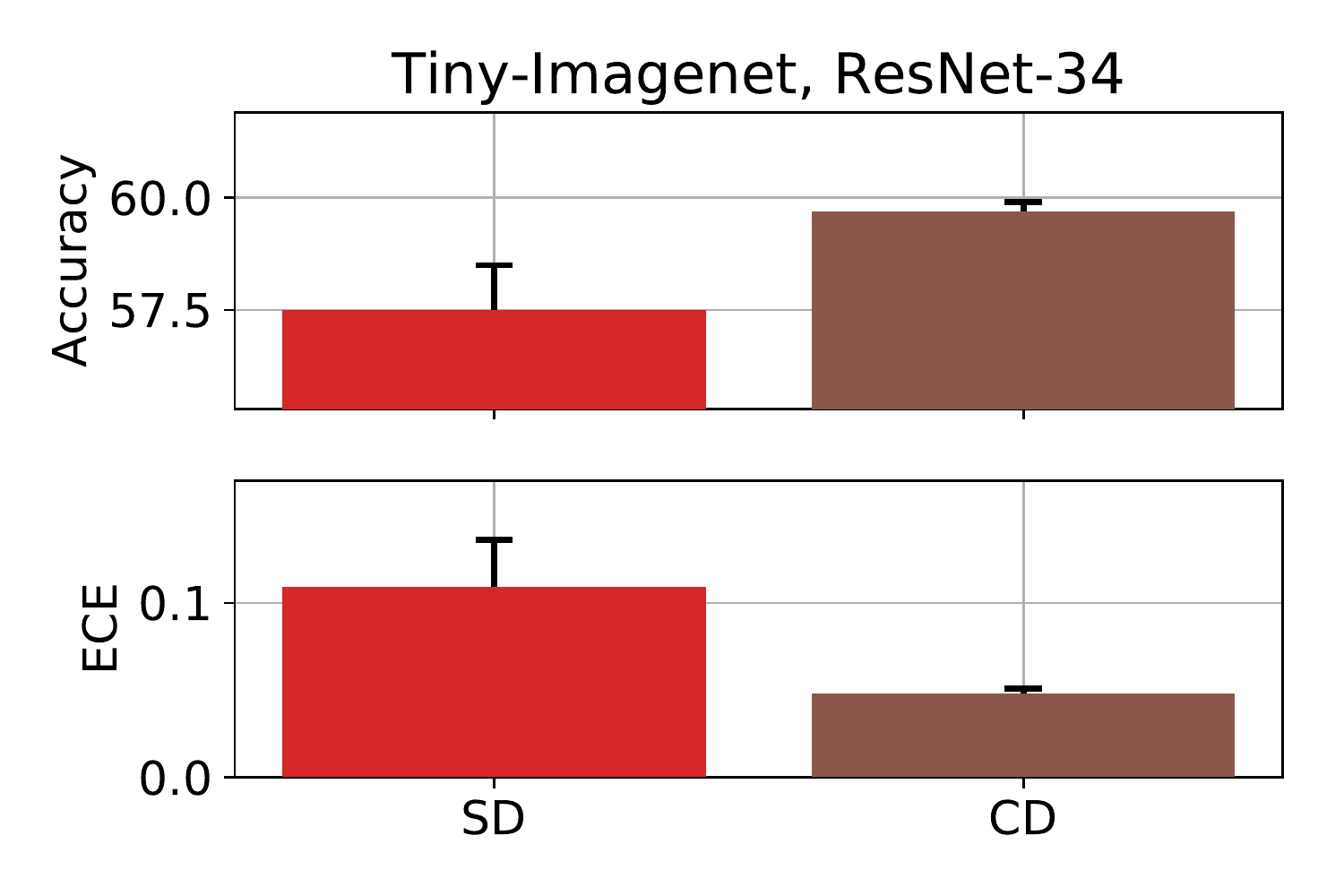}
\end{subfigure}
\begin{subfigure}{.32\textwidth}
\centering
\includegraphics[width=1.05\linewidth]{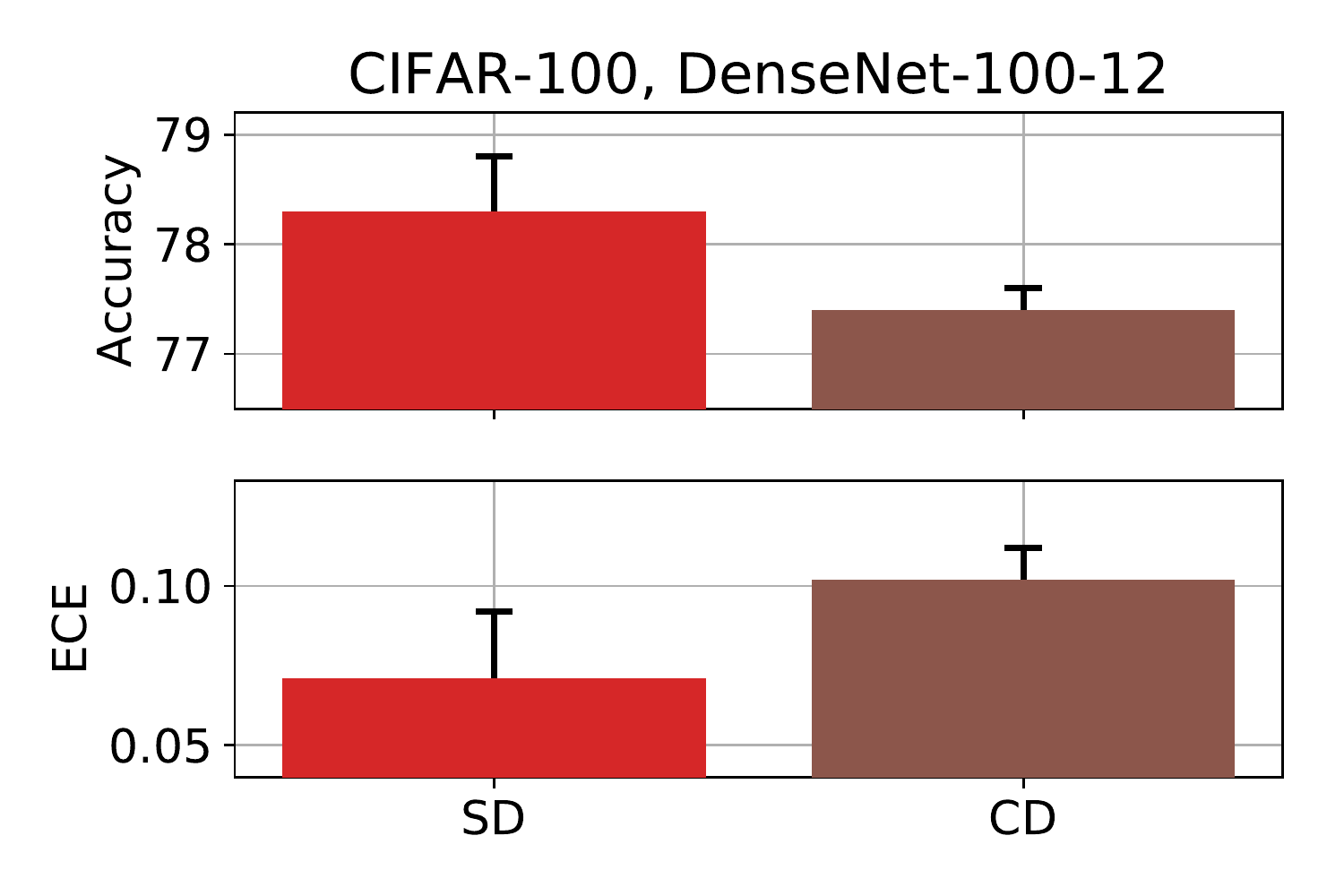}
\end{subfigure}
\begin{subfigure}{.32\textwidth}
\centering
\includegraphics[width=1.05\linewidth]{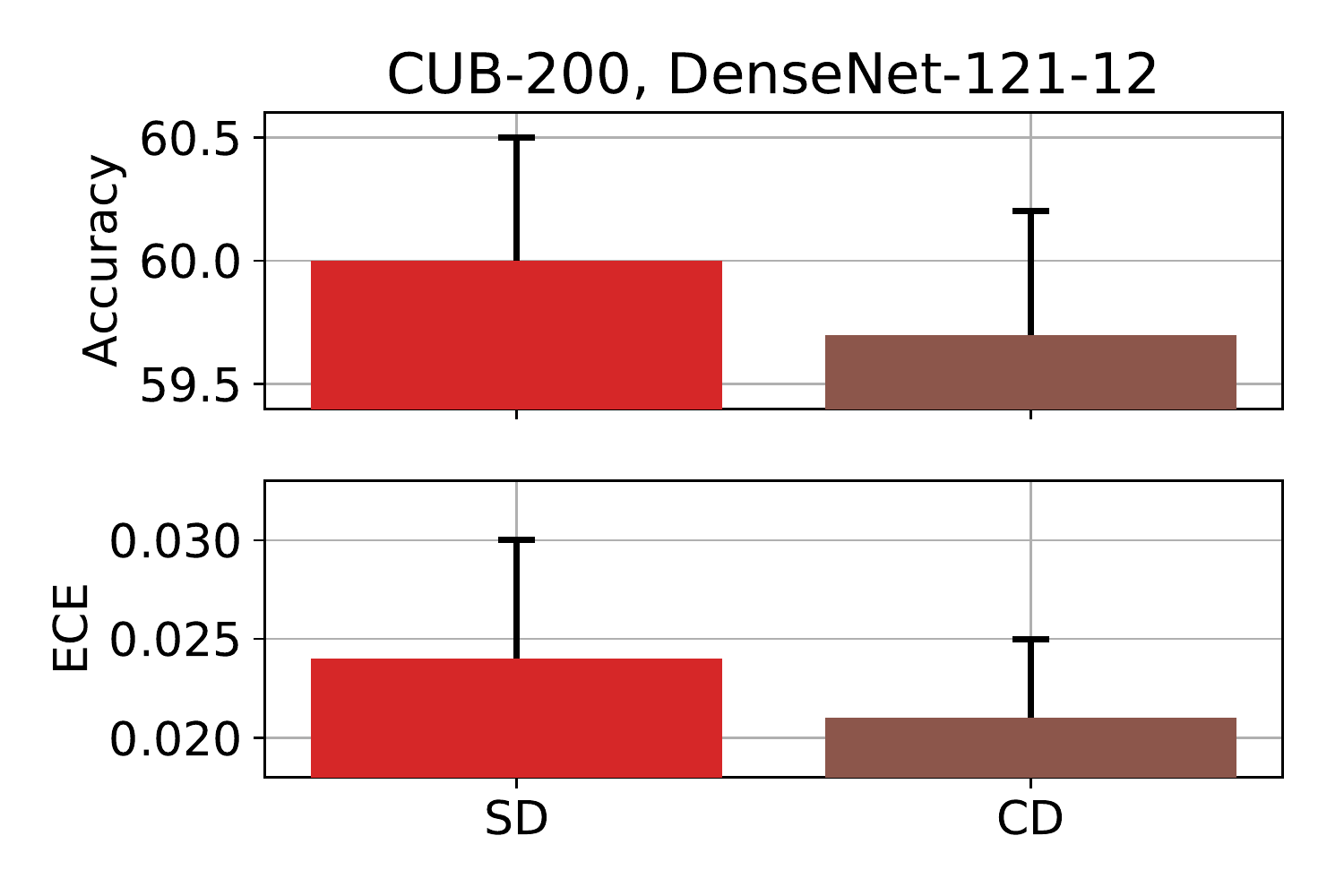}
\end{subfigure}
\begin{subfigure}{.32\textwidth}
\centering
\includegraphics[width=1.05\linewidth]{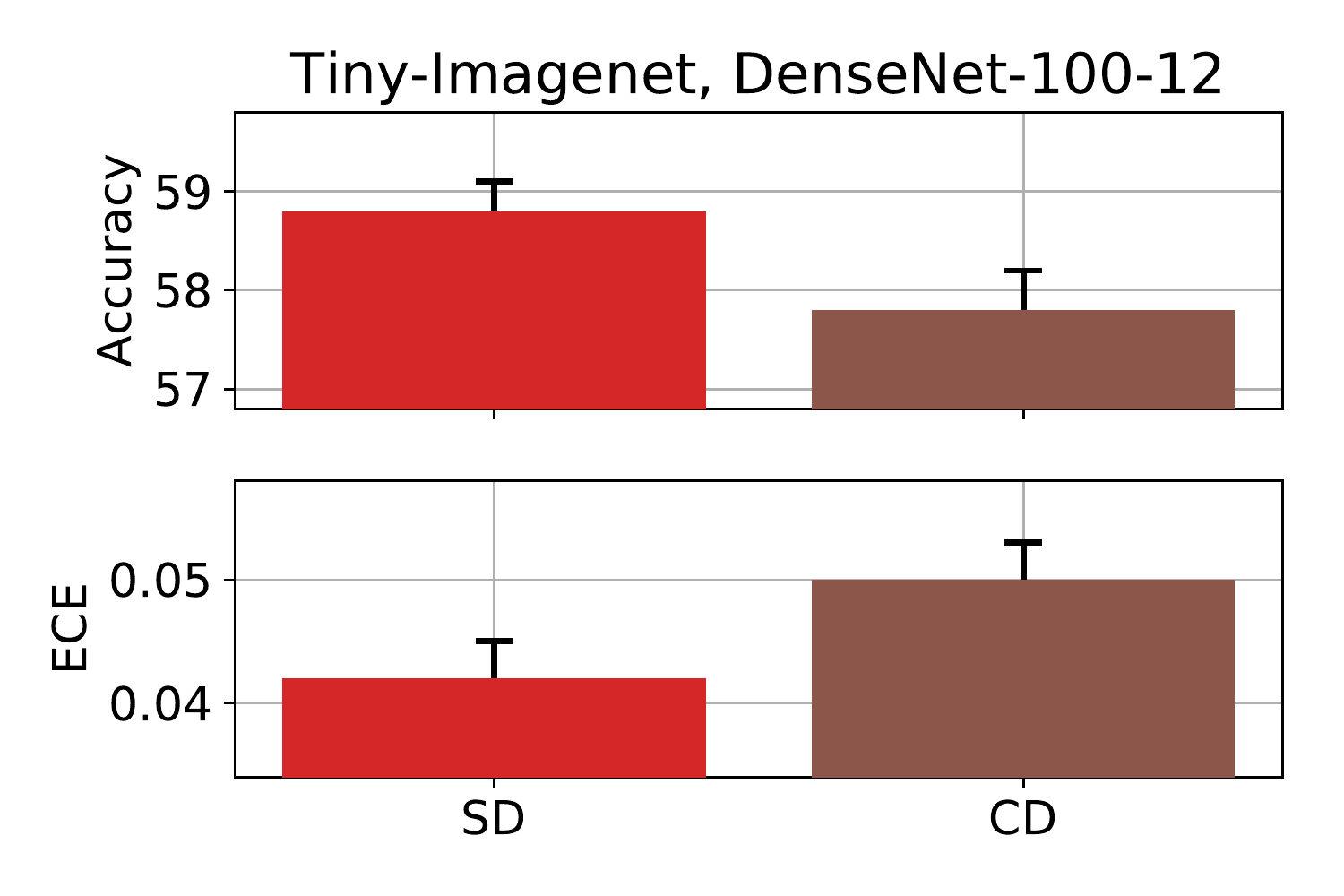}
\end{subfigure}
\caption{Additional results on cross-distillation. "SD" and "CD" refers to "self-distillation" and "cross-distillation" respectively. The top rows of each experiment show bar charts of accuracy on the test set for each experiment conducted, while the bottom rows are bar charts of expected calibration error.}
\label{fig:cd_results}
\end{figure}
We also perform an additional experiment with the identical setup as described in Section~\ref{section:experiment} on \textit{cross distillation} of the ResNet and the DenseNet models, in which a ResNet-34 teacher is used to train the DenseNet-100 student and vice versa in an attempt to examine the effect of better/worse priors in self-distillation. 
Hyper-parameters are fixed in this case such that the predictive uncertainty and diversity associated with the label predictions remain the same as that for self-distillation. 
Results are summarized in Fig.~\ref{fig:cd_results}.
As seen clearly from consistently better/worse performance of cross distillation for ResNet/DenseNet, better teachers lead to better performance. 
Thus, in addition to diversity among teacher predictions, the quality of the instance-specific prior used is also important for better generalization performance. 
Lastly, we also see an apparent benefit in terms of model calibration when a better teacher model is used. 

The interpretation of distillation as sample-specific regularization provides us with a reasonable explanation of why deeper NNs are potentially better teachers. 
With greater capacity, deeper networks can learn better representations that capture more closely the true underlining relative confidence among samples, thereby generating better priors and hence better performance. When too expressive models are used, however, there can be so much overfitting to the ground truth labels that the meaningful rankings are destroyed, despite better accuracy. Recent findings experimentally corroborate our argument~\cite{cho2019efficacy}. Similar observations were also made when label smoothing is applied~\cite{muller2019does}. From the regularization perspective, distillation can be also applied to very deep networks for potential improvements, and shallower teacher models can also serve as teacher models for deeper student networks.

\newpage
\subsection{Additional Experiments on Varying $\gamma$}
\label{appendix:gamma}
\begin{figure}[!htb]
\centering
\begin{subfigure}{.32\textwidth}
\centering
\includegraphics[width=1.05\linewidth]{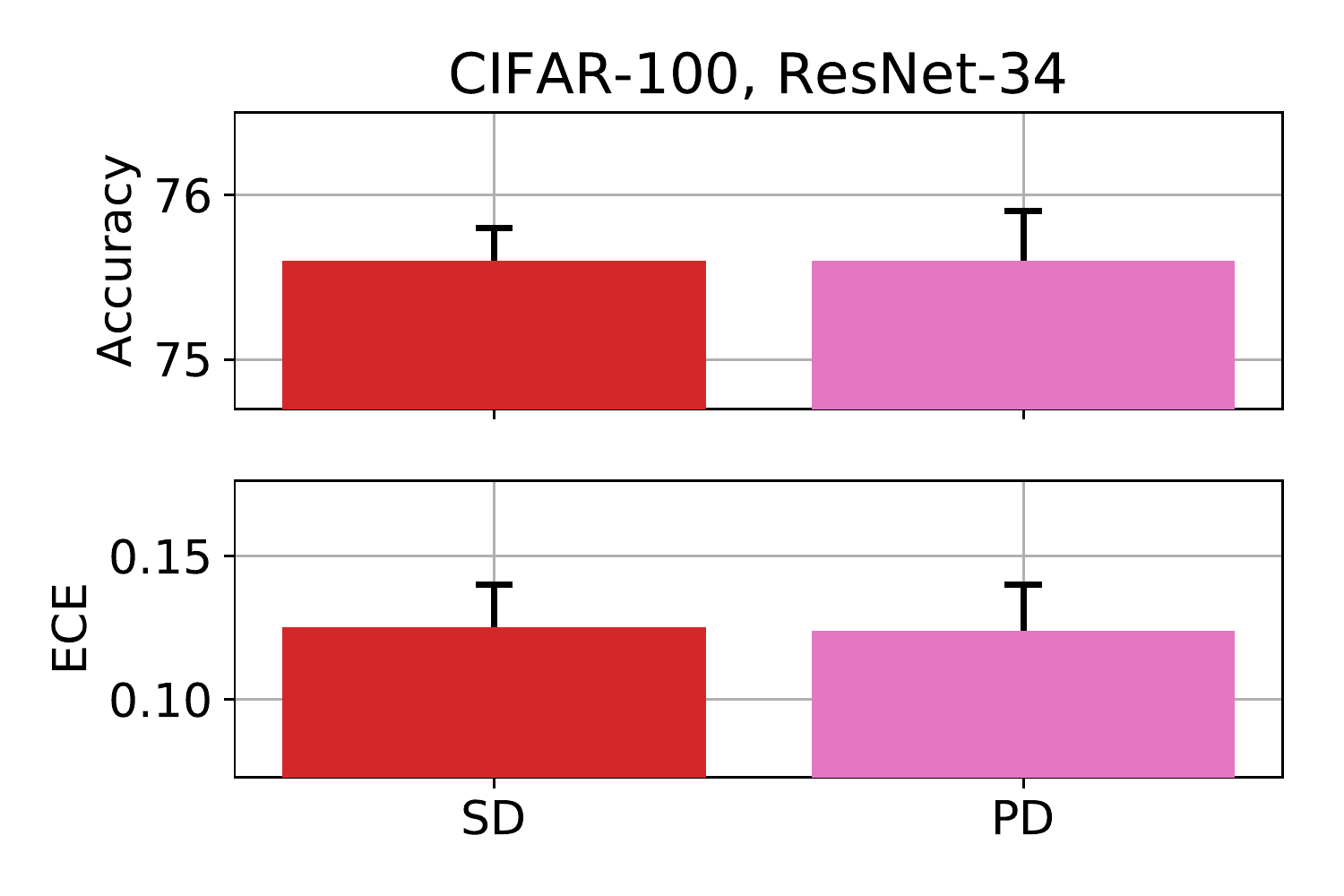}
\end{subfigure}
\begin{subfigure}{.32\textwidth}
\centering
\includegraphics[width=1.05\linewidth]{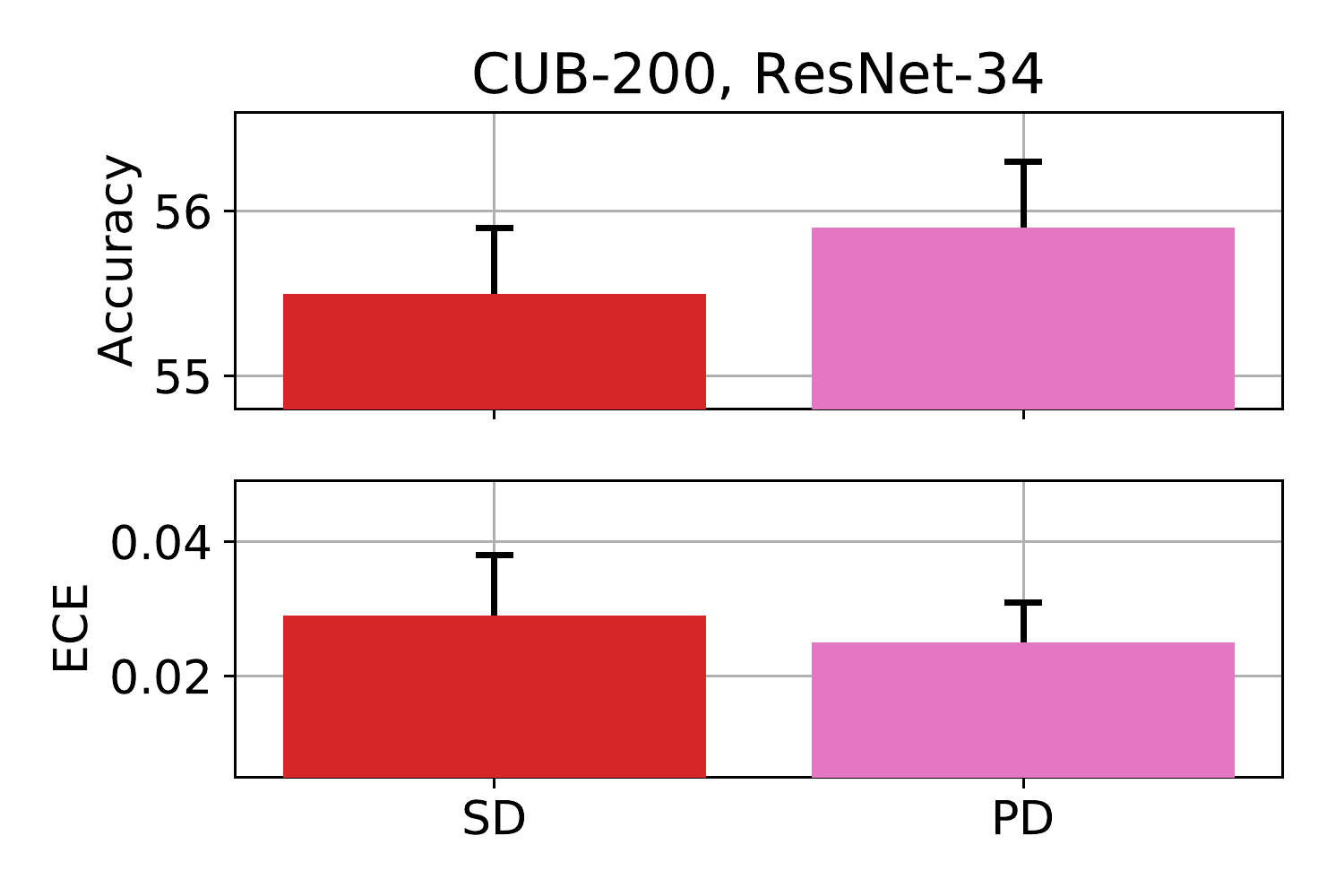}
\end{subfigure}
\begin{subfigure}{.32\textwidth}
\centering
\includegraphics[width=1.05\linewidth]{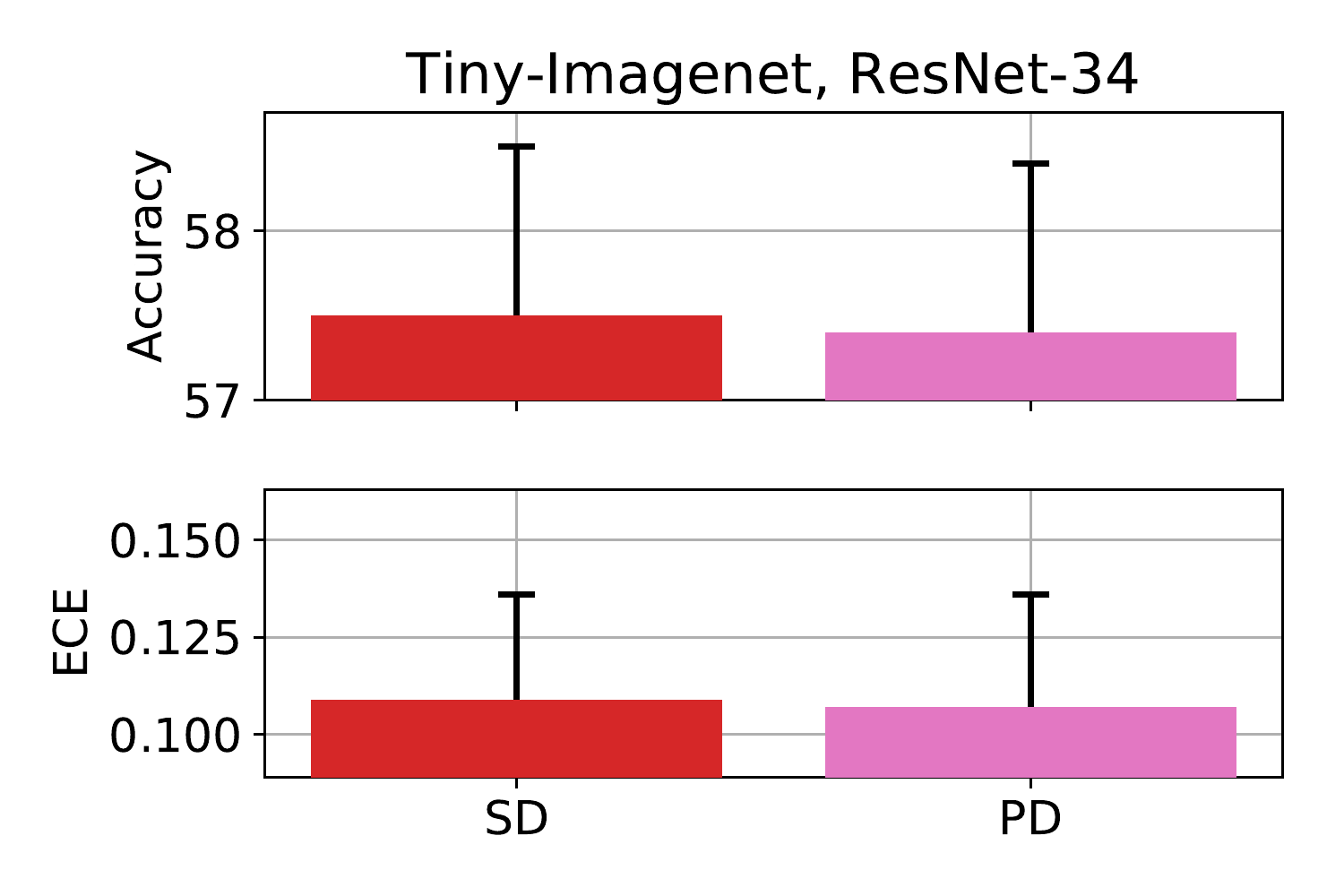}
\end{subfigure}
\begin{subfigure}{.32\textwidth}
\centering
\includegraphics[width=1.05\linewidth]{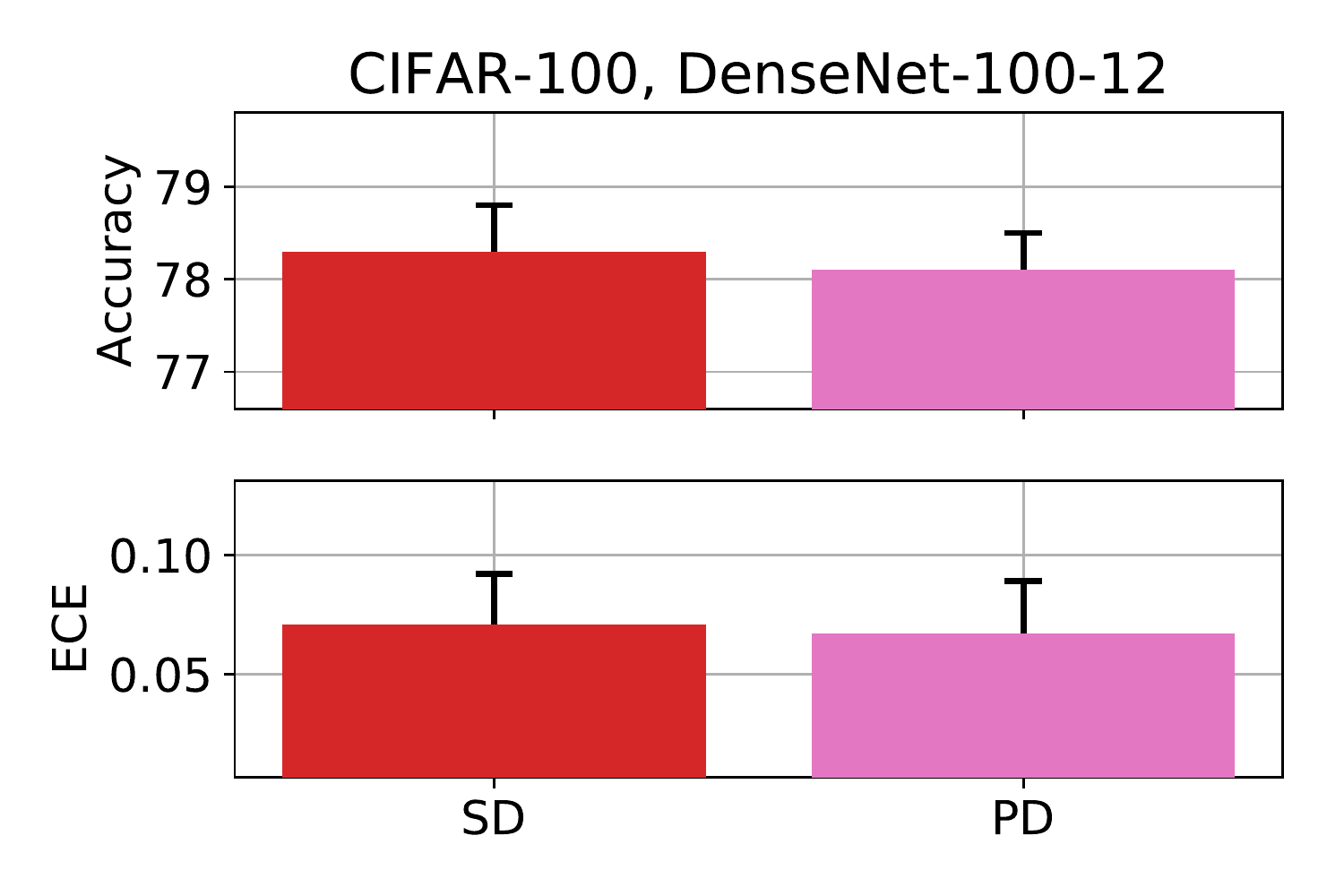}
\end{subfigure}
\begin{subfigure}{.32\textwidth}
\centering
\includegraphics[width=1.05\linewidth]{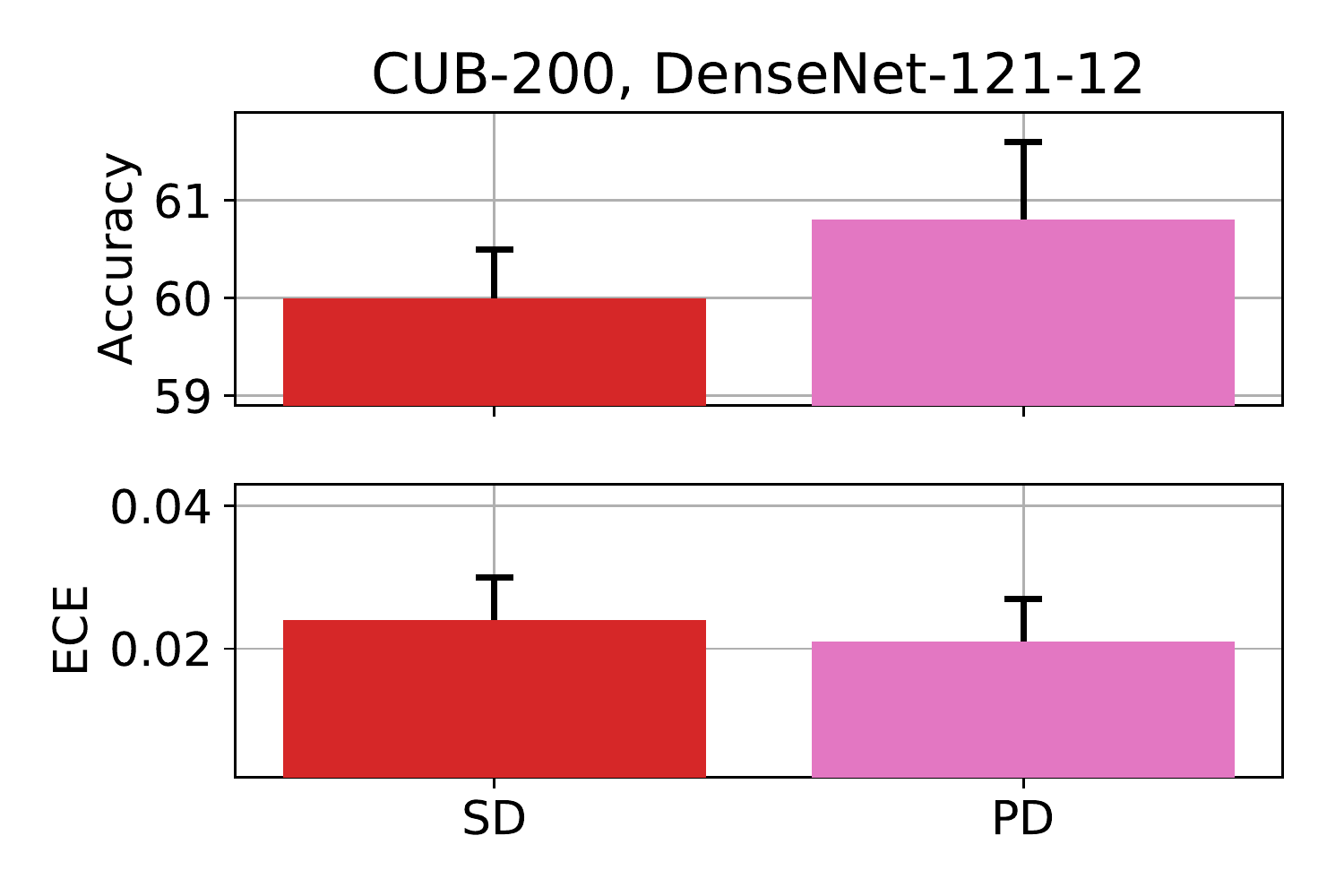}
\end{subfigure}
\begin{subfigure}{.32\textwidth}
\centering
\includegraphics[width=1.05\linewidth]{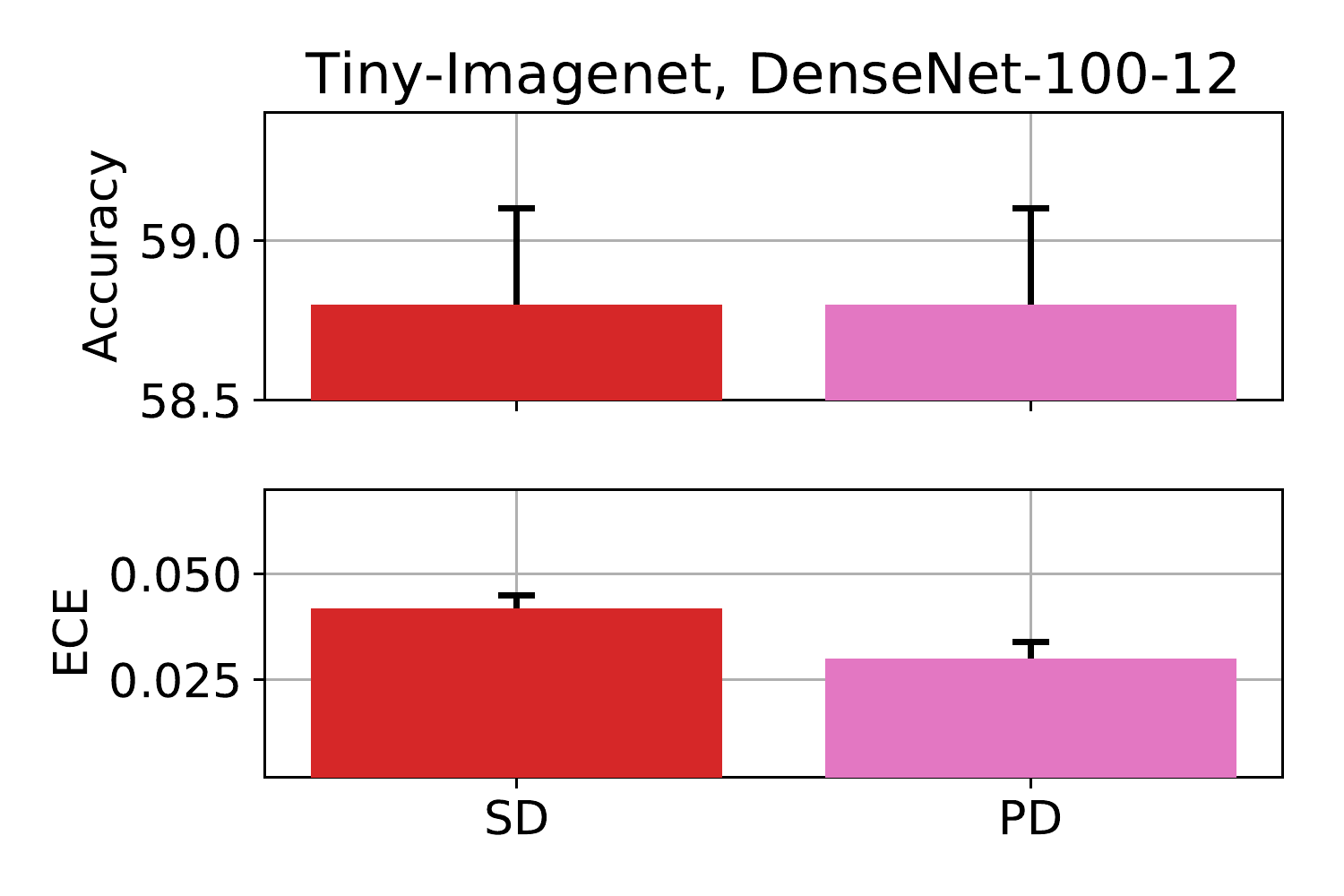}
\end{subfigure}
\caption{Additional results on pruned distillation. "SD" and "PD" refer to "self-distillation" and "pruned-distillation" respectively. The top rows of each experiment show bar charts of accuracy on the test set for each experiment conducted, while the bottom rows are bar charts of expected calibration error.}
\label{fig:pd_results}
\end{figure}
In addition, we consider a simple variation to distillation loss by varying $\gamma$. However, directly adjusting $\gamma$ can be problematic in practice. To understand the effect of changing $\gamma$, suppose we have some $\gamma$ such that $[\boldsymbol{\alpha}_{\boldsymbol{x}}]_{c} - 1 < 0$ for some $c \in \{ 1, ..., k \}$. Since the minimization objective with respect to this class is $-([\boldsymbol{\alpha}_{\boldsymbol{x}}]_{c} - 1)\log([\boldsymbol{z}]_{c})$, the closer the $[\boldsymbol{z}]_{c}$ to $0$, the smaller the loss function. This leads to numerical issues as the overall loss function can be pushed to negative infinity by forcing $[\boldsymbol{z}]_{c}$ arbitrarily close to zero.

To circumvent the numerical problem during optimization, we make the observation that the above objective is essentially equivalent to setting the particular element with $[\boldsymbol{\alpha}_{\boldsymbol{x}}]_{c} - 1 < 0$ to zero. As such, adjusting the threshold $\gamma$ enables us to prune out the smallest elements of the teacher predictions. To further force the pruned elements to zero, a new softmax probability vector is computed with the remaining elements. In practice, setting the optimal $\gamma$ can be challenging. We instead choose to prune out a fixed percentage of classes for all samples. 
For instance, pruning $50\%$ of the classes for a $100$-class classification amounts to using only the top $50$ most confident samples to compute softmax and setting the remaining to zero. We term this method the \textit{pruned-distillation}.

We show some preliminary results with pruned-distillation with $50\%$ of the classes pruned during distillation in Fig.~\ref{fig:pd_results}. While the performance overall remains the same for the CIFAR-100 and Tiny-Imagenet datasets, a slight improvement can be seen for CUB-200 in terms of both the accuracy and ECE, suggesting the method as an easy-to-implement adjustment with no harm. 

\end{document}